\documentclass{article}

\usepackage{microtype}
\usepackage{graphicx}
\usepackage{caption}
\usepackage{subcaption}
\usepackage{booktabs} %
\usepackage{enumitem}
\usepackage{xcolor}

\usepackage{hyperref}

\usepackage[accepted]{icml2026}

\defcitealias{drake}{Tedrake et~al., 2019}

\usepackage{amsmath}
\usepackage{amssymb}
\usepackage{mathtools}
\usepackage{amsthm}

\usepackage[capitalize,noabbrev]{cleveref}
\Crefname{appendix}{Appendix}{Appendices}

\usepackage[section]{placeins}

\theoremstyle{plain}

\theoremstyle{definition}

\theoremstyle{remark}

\icmltitlerunning{SceneSmith: Agentic Generation of Simulation-Ready Indoor Scenes}

\begin{document}

\twocolumn[
  \icmltitle{
  \raisebox{-0.25\height}{%
  \includegraphics[height=1.2em]{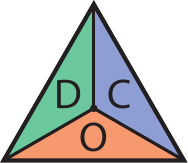}%
  }\hspace{0.5em}%
  SceneSmith: Agentic Generation of Simulation-Ready Indoor Scenes
  }

  \begin{icmlauthorlist}
    \icmlauthor{Nicholas Pfaff}{mit}
    \icmlauthor{Thomas Cohn}{mit}
    \icmlauthor{Sergey Zakharov}{tri}
    \icmlauthor{Rick Cory}{tri}
    \icmlauthor{Russ Tedrake}{mit}

  \end{icmlauthorlist}

  \icmlaffiliation{mit}{Massachusetts Institute of Technology}
  \icmlaffiliation{tri}{Toyota Research Institute}

  \icmlcorrespondingauthor{Nicholas Pfaff}{nepfaff@mit.edu}

  \icmlkeywords{Indoor Scene Synthesis, LLM Agents, Robotics Simulation}

  \vspace{0.5em}
  \centering
  \url{https://scenesmith.github.io/}
  \vspace{-0.75em}

  \begin{center}
    \centering
    \vspace{1em}
    \captionsetup{type=figure}
    \includegraphics{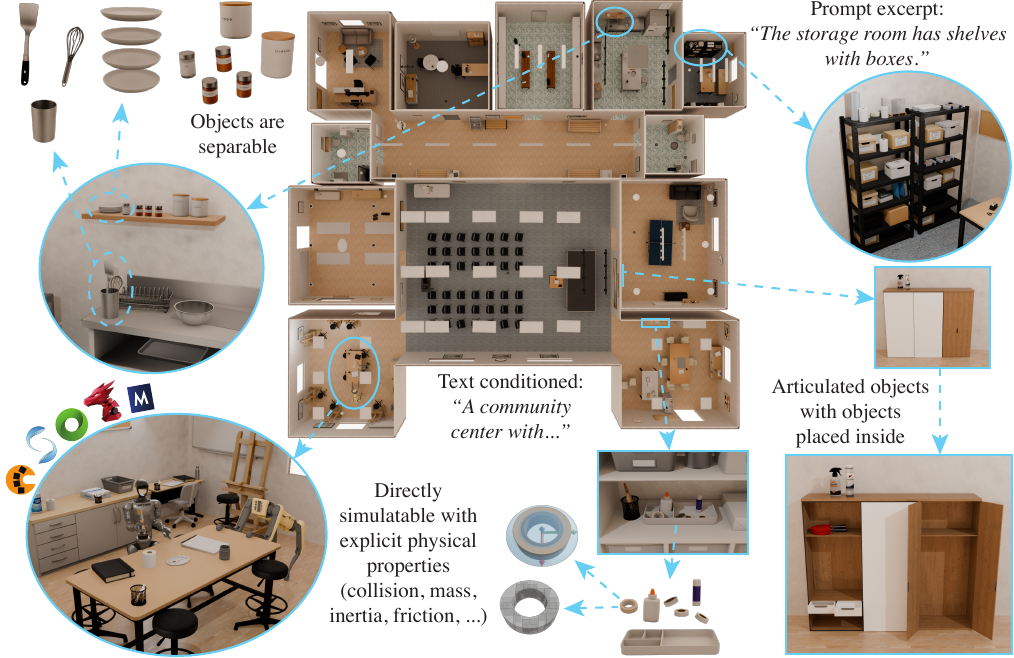}
    \captionof{figure}{
        \textbf{Fully automated text-to-scene generation.} This entire community center was generated by SceneSmith without any human intervention, from a single 151-word text prompt (full prompt in Appendix~\ref{app:prompts_house}). Beyond explicitly specified elements, SceneSmith places additional objects from inferred contextual information, such as ping pong paddles and balls placed near a ping pong table. Objects are generated on-demand, are fully separable (non-composite), and include estimated physical properties, enabling direct interaction within a simulation. The resulting scenes are immediately usable in arbitrary physics simulators (robots added for demonstration).
    }
    \label{fig:teaser}
\end{center}

  \vskip 0.3in

]

\printAffiliationsAndNotice{}  %

\begin{abstract}
  Simulation has become a key tool for training and evaluating home robots at scale, yet existing environments fail to capture the diversity and physical complexity of real indoor spaces. Current scene synthesis methods produce sparsely furnished rooms that lack the dense clutter, articulated furniture, and physical properties essential for robotic manipulation. We introduce SceneSmith, a hierarchical agentic framework that generates simulation-ready indoor environments from natural language prompts. SceneSmith constructs scenes through successive stages—from architectural layout to furniture placement to small object population—each implemented as an interaction among VLM agents: designer, critic, and orchestrator. The framework tightly integrates asset generation through text-to-3D synthesis for static objects, dataset retrieval for articulated objects, and physical property estimation. SceneSmith generates 3-6x more objects than prior methods, with $<$2\% inter-object collisions and 96\% of objects remaining stable under physics simulation. In a user study with 205 participants, it achieves 92\% average realism and 91\% average prompt faithfulness win rates against baselines. We further demonstrate that these environments can be used in an end-to-end pipeline for automatic robot policy evaluation.

\end{abstract}

\section{Introduction}

Recent progress in general-purpose robotics has been driven by large-scale foundation models that promise broad generalization across tasks, embodiments, and environments~\cite{intelligence2025pi05visionlanguageactionmodelopenworld, kim2024openvlaopensourcevisionlanguageactionmodel, trilbmteam2025carefulexaminationlargebehavior}. Companies such as 1X and Sunday are now explicitly targeting the deployment of robots into arbitrary human homes. Achieving this vision requires robots that can robustly perceive, reason, and act across the long tail of real-world indoor environments—spaces that vary widely in layout, object composition, clutter, articulation, and physical interaction affordances.

A central challenge in developing such robots is how to train and evaluate them at scale before deployment. While real-world data collection is essential, it is costly, slow, and difficult to scale across the diversity of homes robots are expected to operate in. As a result, simulation has emerged as a key tool for scalable robot training and evaluation, enabling rapid iteration, controlled experimentation, and safe testing of failure modes~\cite{adam_cotraining, maddukuri2025simandreal, trilbmteam2025carefulexaminationlargebehavior}.

However, most existing simulation environments remain overly simplistic and poorly matched to real human indoor spaces. Typical simulated environments consist of sparsely furnished rooms, limited object diversity, and largely static scenes \cite{maddukuri2025simandreal, trilbmteam2025carefulexaminationlargebehavior}. In contrast, real homes exhibit dense object arrangements, articulated furniture, and fine-grained clutter. For example, even a modest one-bedroom apartment may contain a kitchen cabinet densely packed with plates, bowls, and glasses—all individually manipulable and physically plausible. Such clutter is a central challenge in robotic manipulation~\cite{zeng2022robotic,jia2024cluttergen}, yet sparse simulated environments may fail to prepare robots for these conditions.
This gap between simulated environments and real-world indoor scene distributions limits the effectiveness of simulation-based training and evaluation for general-purpose home robots.
While manual environment design can produce high-quality scenes, this approach is costly, time-consuming, and does not scale easily.

We aim to close this gap by enabling the scalable generation of realistic, simulation-ready indoor environments that reflect the diversity and physical complexity of real homes. Specifically, we seek a framework that takes a natural-language description of an environment or task, and constructs a room- or house-level indoor environment through a sequence of grounded decisions, resulting in scenes that are immediately simulatable and suitable for robotic interaction. Samples from this framework should reflect real-world indoor scene distributions while matching the prompt and being physically feasible.

This problem spans both \emph{asset generation}, producing individual objects with geometry and physical properties, and \emph{scene generation}, where those assets are assembled into multi-room indoor environments with realistic layouts and object arrangements. Prior work has largely addressed these aspects in isolation. Asset-centric approaches focus on reconstructing or synthesizing individual objects with realistic geometrical and physical properties~\cite{scalable_real2sim}. Scene-centric approaches generate layouts or object arrangements assuming a fixed library of assets~\cite{yang2024holodeck, pfaff2025_steerable_scene_generation}. In contrast, we aim to jointly generate simulation-ready assets and assemble them into complete house-level scenes, enabling end-to-end generation of environments suitable for robotics.
Prior scene synthesis methods--whether procedural, data-driven, or LLM-based--primarily target furniture-level layout and visual realism, treating small objects, articulated assets, and physical properties as secondary~\cite{procthor, tang2024diffuscene, yang2025sceneweaver}. This is misaligned with robotics requirements, where dense arrangements of manipulable objects, hierarchical support relationships, and physically valid configurations are essential.

We propose \textbf{SceneSmith}, a hierarchical, agentic framework for generating simulation-ready indoor environments from natural language. SceneSmith constructs scenes through a sequence of stages--from architectural layout to furniture placement to small object population--organized as a tree where rooms and supporting surfaces form independent branches. Building upon recent advances in agentic AI, each stage is implemented as an interaction between three VLM agents: a designer that proposes scene modifications, a critic that evaluates feasibility and alignment, and an orchestrator that manages iterative refinement. Asset generation is tightly integrated through a routing mechanism that combines modern text-to-3D synthesis for static objects, dataset retrieval for articulated furniture, and physical property estimation, ensuring generated scenes are immediately usable for robotics simulation.

We evaluate SceneSmith across 210 diverse room- and house-level prompts, demonstrating its ability to generate dense, articulated, and physically feasible indoor environments. In a user study with 205 participants, SceneSmith achieves 92.2\% average realism win rate and 91.5\% average prompt faithfulness win rate against baselines. SceneSmith generates 3-6x more objects than baselines (71.1 vs 11-23 objects per room on average) while maintaining $<$2\% inter-object collisions and 95.6\% of objects remaining stable under physics simulation, compared to 3-29\% collisions and 8-61\% stability for baselines.
In addition, we demonstrate an end-to-end robot policy evaluation pipeline in which natural-language task descriptions are converted into diverse scene prompts, simulated robots execute policies in the generated environments, and an evaluator agent verifies task completion by jointly reasoning over symbolic scene state and visual observations.
We also include qualitative demonstrations with teleoperation of a humanoid robot (RB-Y1) and zero-shot policy rollouts, further illustrating the suitability of SceneSmith scenes for robot simulation.

In summary, our key contributions are:
\begin{itemize}[nosep]
    \item We introduce \textbf{SceneSmith}, a hierarchical, agentic framework for constructing simulation-ready indoor environments from natural language, designed to support scalable robot training and evaluation.
    \item We develop an \textbf{integrated asset generation and routing pipeline} combining text-to-3D synthesis with retrieval for articulated objects, augmenting all assets with collision geometry and physical properties.
    \item We show that SceneSmith \textbf{outperforms all baselines} in user studies and automated metrics, generating denser, collision-free, and physically stable scenes.
    \item We demonstrate SceneSmith in an \textbf{end-to-end robotics evaluation pipeline}, from natural-language task descriptions to automatic success verification.
\end{itemize}

\section{Related Work}

\textbf{Indoor Scene Synthesis.}
Indoor scene synthesis has been approached through procedural, data-driven, and language-guided paradigms.
Procedural methods like ProcTHOR~\cite{procthor} and Infinigen Indoors~\cite{raistrick2024infinigen} encode object relationships via hand-crafted rules. While scalable, they have limited semantic expressivity and extensibility across scene types.
Data-driven generative models learn spatial patterns from 3D scene datasets~\cite{tang2024diffuscene, yang2024physcene, hu2024mixed}. However, they typically operate under restrictive assumptions such as floor-aligned SE(2) layouts, producing sparse scenes with limited object diversity.
Recent work leverages LLMs and VLMs to provide open-vocabulary semantic priors and natural-language controllability~\cite{yang2024holodeck, IDesign_2024, sun2024layoutvlm, zhou2025roomcraftcontrollablecomplete3d, sun20253dgeneralistselfimprovingvisionlanguageactionmodels}.
While these approaches capture high-level semantics, they often struggle with fine-grained spatial reasoning and physical feasibility, resulting in implausible placements or object interpenetrations.
More broadly, none of the above methods produce scenes with the collision geometry and physical properties required for robotics simulation.
HSM~\cite{pun2026hsm} decomposes scene generation hierarchically—placing furniture before small objects—and identifies support surfaces for manipuland placement. We use their support surface detection and adopt a similar hierarchical structure. Our extensions include hierarchical prompt refinement, where placement prompts derive constraints from the global scene description, and joint surface population, where related surfaces are populated together for consistency. This enables coordinated placement across surfaces (e.g., ``books on one shelf, plants on the other'').

\textbf{Agentic Scene Synthesis.}
Recent work has moved from single-shot generation to iterative refinement through agentic systems.
SceneWeaver~\cite{yang2025sceneweaver} employs a reason-act-reflect paradigm where a single LLM planner selects one tool per turn from an extensible toolkit, guided by physical and semantic evaluations.
LL3M~\cite{lu2025ll3m} introduces iterative refinement between a designer (coding agent) and critic for 3D asset generation; we adapt this pattern for scene-level generation.
We build on these agentic approaches but allow unlimited tool invocations per agent turn, including visual observations and state feedback, enabling agents to self-verify before handoff.

\section{Agentic Scene Construction for Simulation-Ready Environments}

\begin{figure*}[t]
    \centering
    \includegraphics[width=1\linewidth, keepaspectratio]{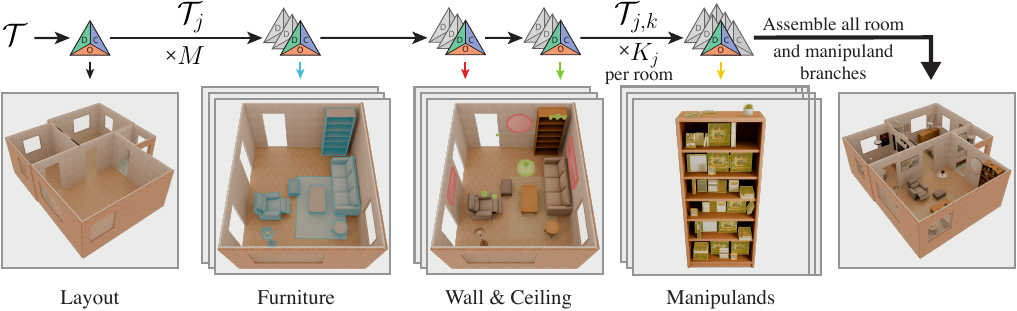}
    \caption{
        \textbf{SceneSmith's hierarchical scene construction pipeline.} A scene prompt $\mathcal{T}$ is processed by a layout agent to generate architectural geometry for $M$ rooms. Each room is then independently populated through furniture, wall-mounted, and ceiling-mounted stages using room-specific prompts $\mathcal{T}_j$. In each room, $K_j$ supporting entities subsequently form additional branches populated with manipulable objects using entity-specific prompts $\mathcal{T}_{j,k}$. Colored highlights indicate objects added at each stage. Each stage (colored triangle) is implemented as an agentic interaction between a \textcolor[HTML]{68c0a4}{Designer}, \textcolor[HTML]{8ca0cc}{Critic}, and \textcolor[HTML]{f9a87e}{Orchestrator}. Stacked frames indicate parallel branches.
    }
    \label{fig:system_diagram}
\end{figure*}

We present SceneSmith, a hierarchical agentic system for constructing simulation-ready indoor environments from natural language prompts. SceneSmith decomposes scene creation into a sequence of decisions over layout, furnishing, object population, and refinement. Each stage is implemented as an agentic interaction between a designer, a critic, and an orchestrator, each equipped with specialized tools.

We begin by describing the scene representation and hierarchical construction process (Section~\ref{sec:scene_rep}). We then detail the agentic interactions and tool abstractions (Section~\ref{sec:agents}). Next, we introduce the asset generation and routing pipeline used to produce simulation-ready objects (Section~\ref{sec:assets}), followed by physical feasibility post-processing (Section~\ref{sec:physics}). Finally, we describe an application to automatic robot policy evaluation using our generated scenes (Section~\ref{sec:task_eval}).

\subsection{Scene Hierarchy}
\label{sec:scene_rep}

We represent a \emph{scene} as a set of rooms, $\mathcal{S} = \{ \mathcal{R}_j \mid j \in \{1, \dots, M\} \}$, constructed from a natural-language prompt $\mathcal{T}$. Each \emph{room} $\mathcal{R}_j = (\mathcal{G}_j, \mathcal{O}_j)$ consists of \emph{architectural geometry} $\mathcal{G}_j$ (walls, floor, doors, windows) and \emph{objects} $\mathcal{O}_j = \{ (\mathcal{A}_i, \mathcal{X}_i) \}$. Each object pairs a simulation-ready \emph{asset} $\mathcal{A}_i$, comprising visual geometry, collision geometry, physical properties, and joint definitions if it is articulated, with a pose $\mathcal{X}_i \in SE(3)$.

SceneSmith constructs scenes $\mathcal{S}$ through a tree of stages (Figure~\ref{fig:system_diagram}). The root stage generates architectural layout, determining the number of rooms $M$ and producing geometry $\mathcal{G}_j$ for each. This geometry specifies room extents and structural elements. Each room $\mathcal{R}_j$ is then populated independently by adding objects $\mathcal{O}_j$ through successive stages: furniture, wall-mounted objects, and ceiling fixtures. Each stage is guided by a room-specific prompt $\mathcal{T}_j$ derived from $\mathcal{T}$. Finally, selected supporting entities (furniture surfaces, wall shelves, floor regions) spawn additional branches for adding small manipulable objects to $\mathcal{O}_j$, each guided by an entity-specific prompt $\mathcal{T}_{j,k}$. All stages are implemented as agentic interactions (Section~\ref{sec:agents}). This hierarchical prompt refinement enables local decisions to be made independently while remaining coherent with scene intent. Finally, all room and manipuland branches are assembled into the flat scene representation $\mathcal{S}$, suitable for direct export to robot simulators.

\subsection{Agentic Trio: Designer, Critic, and Orchestrator}
\label{sec:agents}

Each stage of SceneSmith’s hierarchical construction process is implemented as an interaction between three agents with complementary roles: a \emph{designer}, a \emph{critic}, and an \emph{orchestrator}. This decomposition separates scene proposal, evaluation, and control flow, enabling structured refinement while keeping individual agent responsibilities simple. This separation reduces self-evaluation bias: an independent critic is better positioned to identify errors or omissions that a proposal-focused designer may overlook.

The \textbf{designer} proposes modifications to the scene state at the current stage using structured tools. The \textbf{critic} evaluates the resulting scene with respect to factors such as semantic plausibility, physical feasibility, and alignment with the stage objective, providing scalar scores and natural-language feedback. The \textbf{orchestrator} coordinates this interaction, tracking scores and determining when to accept a proposal, request further refinement, or terminate the stage. To prevent degradation during iterative refinement, the orchestrator maintains checkpoints of prior scene states and can revert changes when critic scores decrease.

\textbf{Agent Tools.}  
Agents interact with the scene exclusively through tools that provide structured observation and modification operations. We organize tools into functional categories that are shared across stages, including \emph{state observation tools} (e.g., querying object metadata and poses), \emph{visual observation tools} (e.g., rendering scene views), \emph{scene modification tools} (e.g., placing or adjusting assets), \emph{asset acquisition tools} for generating or retrieving simulation assets, and \emph{feasibility verification tools} (e.g., collision and reachability checks).
Object placement is performed relative to supporting surfaces. Agents specify $SE(2)$ poses within a surface coordinate frame (e.g., on floors, walls, or shelves). Full $SE(3)$ object poses arise by lifting these placements through the known pose of the supporting surface.
In addition, certain stages expose \emph{specialized tools} tailored to their construction context. For example, furniture placement stages include snapping tools that translate objects into contact-aligned relative configurations (e.g., chairs snapped toward tables or cabinets snapped against walls), as well as relational facing queries that evaluate whether an object is oriented toward or away from another object or architectural element. Manipuland population stages additionally include tools for assembling multiple assets into composite object groups that are placed jointly (e.g., constructing a fruit bowl by placing individual fruit relative to a bowl and then placing the assembled group on a supporting surface).
Agents may invoke arbitrarily many tools within a single turn, making complex edits via multiple atomic operations. A complete description of available tools is provided in Appendix~\ref{app:architecture}. Tool access is role-dependent: the designer has access to scene modification tools, the critic is restricted to observation and verification tools, and the orchestrator invokes designer and critic agents as tools while managing global control operations such as state rollback.

\textbf{Agent Memory.}
SceneSmith maintains a persistent, agent-specific session memory within each construction stage, allowing agents to retain context across turns during iterative refinement. To bound context growth, each agent uses a turn-based memory in which the current and immediately preceding turn are retained in full, while earlier agent turns are replaced by LLM-generated summaries. Visual observations are stored in a bounded sliding window and discarded once no longer needed. Memory is reset between stages.

\subsection{Asset Generation and Routing}
\label{sec:assets}

\begin{figure*}
    \centering
    \includegraphics[width=1\linewidth, keepaspectratio]{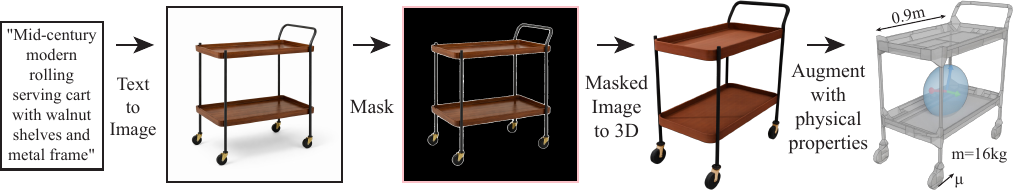}
    \caption{
        \textbf{Text-to-3D asset generation pipeline.} Given an object description, we generate an image, segment the foreground, and reconstruct a textured 3D mesh. The mesh is augmented with collision geometry (gray convex pieces) and physical properties estimated by a VLM, including mass, center of mass, friction, and inertia (blue ellipsoid). The mesh is also scaled to target dimensions.
    }
    \label{fig:text_to_3d}
\end{figure*}

SceneSmith integrates asset generation directly into the scene construction process, enabling open-set object vocabularies while ensuring simulation readiness. Given an asset request from a designer agent, an \emph{asset router} resolves the request into one or more atomic assets and selects an appropriate acquisition strategy, returning validated simulation-ready objects.

\textbf{Generative Asset Synthesis.}
For static objects, SceneSmith primarily relies on generative text-to-3D synthesis rather than retrieval from existing asset libraries. Generating assets on demand also avoids training data contamination, enabling fair evaluation of robot policies on truly unseen objects. Given a textual object description, we generate a reference image using a text-to-image model (GPT Image 1.5), segment the foreground object using SAM3~\cite{sam3}, and reconstruct a textured 3D mesh from the segmented image using SAM3D~\cite{sam3d}. The resulting mesh is canonicalized to a standard orientation, scaled to target dimensions specified in the asset request, and augmented with collision geometry and estimated physical properties (Figure~\ref{fig:text_to_3d}).

\textbf{Articulated Object Library.}
For objects with movable parts, such as cabinets, drawers, or appliances, SceneSmith retrieves assets from ArtVIP~\cite{artvip}, an articulated object library containing pre-authored multi-link models with joint definitions. We augment these with estimated physical properties (Appendix~\ref{app:physics_estimation}). While generative methods are effective for static objects, in our experience, current text-to-3D approaches do not yet reliably produce articulated structure and kinematics suitable for robotics simulation \cite{chen2024urdformerpipelineconstructingarticulated, jiayi2024singapo}.

\textbf{Thin Coverings.}
To represent flat decorative elements such as rugs, posters, or tablecloths, we introduce \emph{thin coverings}: lightweight geometric surfaces paired with physically based materials. Thin coverings capture visual detail and clutter while avoiding unnecessary rigid-body complexity. We retrieve materials from ambientCG\footnote{\url{https://ambientcg.com/}} and fall back to image-generated materials when library assets cannot satisfy the request (Appendix~\ref{app:thin_covering}).

\textbf{Asset Routing and Validation.}
The asset router decomposes composite requests into individually manipulable assets when needed (e.g., representing a fruit bowl as a bowl plus multiple fruit objects) and selects among generation, articulated retrieval, or thin covering strategies based on object type and placement context. All candidate assets undergo validation, including mesh integrity checks and VLM-based semantic verification. Assets that fail validation are regenerated or rerouted up to a fixed retry budget, after which failure feedback is returned to the agent. This routing and validation loop enables robust, scalable asset acquisition without manual curation (Appendices~\ref{app:asset_router},~\ref{app:asset_validation}).

\subsection{Simulation Readiness}
\label{sec:physics}

Architectural elements use volumetric geometry (e.g., walls with finite thickness) rather than planes, improving robustness to penetration under discrete time-stepping physics simulation. All objects are augmented with estimated physical properties (mass, center of mass, inertia, friction) during asset generation, enabling realistic dynamics (Appendix \ref{app:physics_estimation}). While SceneSmith's agentic construction process encourages semantically plausible and physically reasonable placements through iterative feedback, agents are not required to satisfy physical constraints exactly during generation. As a result, generated scenes may contain inter-object penetrations or objects placed in configurations that are not statically stable. To ensure that environments are directly simulatable, we apply a lightweight post-processing step after both furniture and manipuland placement stages that enforces physical feasibility.

We first resolve inter-object penetrations by projecting object positions to the nearest collision-free configuration using nonlinear optimization, while preserving orientations, as in~\cite{pfaff2025_steerable_scene_generation}. We then simulate the scene under gravity in Drake \citepalias{drake} to allow unstable objects to settle into statically stable configurations. This minimizes penetrations and ensures static equilibrium without requiring manual intervention or object welding.

\subsection{Application: Robot Policy Evaluation}
\label{sec:task_eval}

As one application of SceneSmith, we present an end-to-end evaluation pipeline that connects natural-language robot tasks to generated environments, robot execution, and automated task verification. This enables scalable evaluation of robot policies in diverse, simulation-ready scenes without manual environment or evaluation predicate design.

Given a natural-language task description (e.g., ``find a fruit and place it on the table''), we first use a language model to generate a set of diverse scene prompts consistent with the task requirements. These prompts are passed to SceneSmith to produce multiple task-relevant indoor environments, allowing evaluation across varied layouts, object configurations, and clutter conditions.

Robots then execute policies directly in the generated environments. As an example instantiation, we demonstrate this pipeline using a model-based pick-and-place policy that operates on the generated simulator scenes. While we use Drake for demonstration, our scenes can be exported to other major robotics simulators (e.g., MuJoCo~\cite{mujoco}, Isaac Sim, Genesis; see Appendix~\ref{app:export}).

Finally, task completion is verified by an evaluator agent that jointly reasons over symbolic scene state and visual observations rendered from the simulator. The evaluator does not rely on fixed success predicates; instead, it uses structured tools to gather evidence for task completion, including querying object poses and rendering selected objects for visual inspection. This avoids hand-crafted success metrics and supports open-ended tasks, though at the cost of determinism. See Appendix~\ref{app:robot_eval} for details.

\section{Evaluation}

\begin{figure*}[t]
    \centering
    \includegraphics[width=\textwidth]{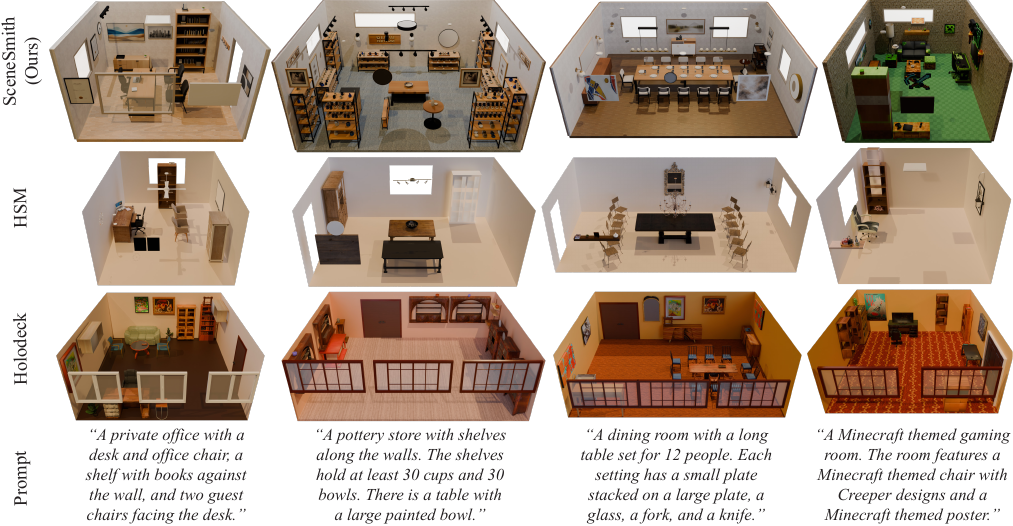}
    \caption{Qualitative comparison with HSM and Holodeck, the two strongest baselines in our user study. SceneSmith produces denser scenes that better satisfy prompt requirements. See Appendix~\ref{app:qualitative_results} for additional qualitative visual results. Many SceneSmith-generated scenes can be inspected in the interactive 3D visualizer on the \href{https://scenesmith.github.io/}{project page}.}
    \label{fig:room_qualitative}
\end{figure*}

\subsection{Experimental Setup}

\textbf{Baselines.}
We compare against five external baselines: HSM~\cite{pun2026hsm}, a hierarchical framework using learned scene motifs; Holodeck~\cite{yang2024holodeck}, which uses constraint satisfaction for layout optimization; I-Design~\cite{IDesign_2024}, a multi-role LLM pipeline; LayoutVLM~\cite{sun2024layoutvlm}, combining visual prompting with differentiable optimization; and SceneWeaver~\cite{yang2025sceneweaver}, a single-agent framework using iterative refinement. We evaluate LayoutVLM with both its original curated asset library and the larger Objaverse~\cite{deitke2022objaverse} library used by Holodeck for fair comparison.
A feature-level comparison with prior scene-generation systems is provided in Appendix~\ref{app:feature_comparison}.

\textbf{Ablations.}
We evaluate six ablations: NoCritic (initial design only), NotGenerated (HSSD~\cite{khanna2023hssd} assets instead of generated), NoAssetValidation (no asset validation), NoSpecializedTools (no specialized furniture or manipuland tools), NoObserveScene (no visual observations), and NoAgentMemory (no session memory).

\textbf{Input Text Descriptions.}
We evaluate on 210 prompts across five categories: \emph{SceneEval-100}~\cite{tam2025sceneeval} room prompts, \emph{Type Diversity} prompts covering underrepresented room types like pet stores and yoga studios, \emph{Object Density} prompts for high object count scenarios, \emph{Themed Scenes} with stylistic constraints, and \emph{House-Level} multi-room prompts. The corpus spans both detailed instructions and sparse prompts, such as ``A bedroom with a bed and a wardrobe.'' House-level prompts are limited to SceneSmith and Holodeck, the only multi-room baseline. See Appendix~\ref{app:prompts} for the complete prompt corpus.

\textbf{Human Study.}
We conducted a pairwise comparison study with 205 crowdsourced participants, collecting 3,051 valid responses. Each comparison presents two scenes side-by-side with an interactive 3D viewer and asks two questions: (1) which scene looks more realistic (forced choice), and (2) which scene better follows the prompt requirements (with ``Equal'' option). See Appendix~\ref{app:user_study_methodology} for details.

\textbf{Automatic Evaluation.}
We use SceneEval~\cite{tam2025sceneeval} with the following metrics: CNT (object count), ATR (object attributes), OOR (object-object relationships), OAR (object-architecture relationships), ACC (accessibility), NAV (navigability), and OOB (out-of-bounds). We note that these VLM-based metrics have limitations including false positives and negatives (Appendix~\ref{app:sceneeval_metrics}). We add two physics metrics using Drake~\citepalias{drake} to evaluate simulation-readiness: COL (collision rate) and STB (static equilibrium). Since baseline methods do not produce simulation-ready scenes, we augment their outputs with collision geometry and physical properties to enable fair comparison (Appendix~\ref{app:sceneeval_metrics}).

\subsection{Scene Generation Results}

\begin{table*}[t]
\centering
\caption{User study results (179 room-level, 31 house-level prompts): SceneSmith vs baselines and ablations. \#Obj shows the comparison method's average object count with 95\% CI. Win rates show percentage preferred (excluding ties for faithfulness), with 95\% Wilson score CIs. Cohen's $h$ measures effect size. Significance: ** $p<0.001$, * $p<0.01$, -- $p \geq 0.05$ (FDR-corrected).}
\label{tab:user_study}
\small
\setlength{\tabcolsep}{4pt}
\begin{tabular}{lccccccccc}
\toprule
& & \multicolumn{4}{c}{\textbf{Realism}} & \multicolumn{4}{c}{\textbf{Faithfulness}} \\
\cmidrule(lr){3-6} \cmidrule(lr){7-10}
\textbf{Comparison} & \textbf{\#Obj} & \textbf{Win\%} & \textbf{95\% CI} & \textbf{$h$} & \textbf{Sig.} & \textbf{Win\%} & \textbf{95\% CI} & \textbf{$h$} & \textbf{Sig.} \\
\midrule
\multicolumn{10}{l}{\textit{External Baselines (SceneSmith averages $71.1 \pm 13.0$ objects)}} \\
vs HSM & $22.7 \pm 2.6$ & 88.5 & [83.5, 92.1] & 0.88 & ** & 85.2 & [79.7, 89.3] & 0.78 & ** \\
vs Holodeck & $23.0 \pm 1.4$ & 88.6 & [83.8, 92.2] & 0.88 & ** & 90.6 & [85.9, 93.8] & 0.95 & ** \\
vs SceneWeaver & $13.5 \pm 1.0$ & 91.7 & [87.3, 94.7] & 0.99 & ** & 92.9 & [88.6, 95.6] & 1.03 & ** \\
vs I-Design & $13.0 \pm 1.0$ & 94.1 & [90.1, 96.5] & 1.08 & ** & 90.6 & [85.9, 93.8] & 0.95 & ** \\
vs LayoutVLM (Curated) & $11.2 \pm 1.6$ & 94.9 & [91.1, 97.1] & 1.12 & ** & 95.8 & [92.3, 97.8] & 1.16 & ** \\
vs LayoutVLM (Objaverse) & $14.2 \pm 1.7$ & 95.4 & [91.7, 97.5] & 1.14 & ** & 93.9 & [89.9, 96.4] & 1.07 & ** \\
\midrule
\multicolumn{10}{l}{\textit{Ablations}} \\
vs NotGenerated & $57.7 \pm 9.9$ & 63.8 & [57.2, 69.9] & 0.28 & ** & 67.0 & [60.0, 73.3] & 0.35 & ** \\
vs NoAssetValidation & $72.7 \pm 16.3$ & 63.0 & [56.3, 69.1] & 0.26 & ** & 62.2 & [54.8, 69.1] & 0.25 & * \\
vs NoObserveScene & $69.7 \pm 15.4$ & 61.5 & [54.9, 67.7] & 0.23 & * & 53.2 & [45.7, 60.5] & 0.06 & -- \\
vs NoSpecializedTools & $61.5 \pm 14.1$ & 54.8 & [48.2, 61.2] & 0.10 & -- & 53.2 & [45.7, 60.5] & 0.06 & -- \\
vs NoAgentMemory & $78.9 \pm 19.7$ & 53.4 & [46.8, 59.9] & 0.07 & -- & 55.1 & [47.5, 62.4] & 0.10 & -- \\
vs NoCritic & $54.0 \pm 8.3$ & 51.8 & [45.2, 58.4] & 0.04 & -- & 47.5 & [40.4, 54.8] & $-$0.05 & -- \\
\midrule
\multicolumn{10}{l}{\textit{House-Level (SceneSmith averages $214.1 \pm 60.9$ objects)}} \\
vs Holodeck & $81.3 \pm 18.3$ & 80.3 & [63.5, 91.0] & 0.65 & ** & 84.7 & [67.5, 93.8] & 0.76 & ** \\
\bottomrule
\end{tabular}
\end{table*}

\begin{table*}[t]
\centering
\caption{Automated evaluation (179 room-level, 31 house-level scenes). Values are mean $\pm$ 95\% CI. $\uparrow$ = higher is better, $\downarrow$ = lower is better. \textbf{Bold} = best within section, \underline{underline} = second best. See Appendix~\ref{app:sceneeval_results} for full results with ablations.}
\label{tab:sceneeval}
\small
\setlength{\tabcolsep}{4pt}
\resizebox{\textwidth}{!}{%
\begin{tabular}{l c ccccccccc}
\toprule
\textbf{Method} & \textbf{\#Obj} & \textbf{CNT}$\uparrow$ & \textbf{ATR}$\uparrow$ & \textbf{OOR}$\uparrow$ & \textbf{OAR}$\uparrow$ & \textbf{ACC}$\uparrow$ & \textbf{NAV}$\uparrow$ & \textbf{COL}$\downarrow$ & \textbf{STB}$\uparrow$ & \textbf{OOB}$\downarrow$ \\
\midrule
HSM & 22.7$\pm$2.6 & 60.6$\pm$4.2 & \underline{61.5$\pm$7.3} & \underline{30.6$\pm$5.7} & 66.2$\pm$6.6 & 88.9$\pm$1.7 & 99.4$\pm$0.4 & 20.6$\pm$3.0 & 45.2$\pm$3.9 & 5.5$\pm$1.1 \\
Holodeck & \underline{23.0$\pm$1.4} & 44.2$\pm$4.3 & 44.4$\pm$7.2 & 16.8$\pm$4.6 & 38.0$\pm$7.1 & 84.9$\pm$1.4 & \underline{99.6$\pm$0.2} & 12.3$\pm$2.3 & 31.9$\pm$2.7 & 0.8$\pm$0.4 \\
SceneWeaver & 13.5$\pm$1.0 & 41.8$\pm$3.9 & 29.1$\pm$6.7 & 15.3$\pm$4.2 & 31.6$\pm$6.5 & 77.5$\pm$2.7 & 98.1$\pm$1.2 & 12.5$\pm$2.5 & 37.3$\pm$4.4 & \textbf{0.0$\pm$0.0} \\
I-Design & 13.0$\pm$1.0 & \underline{69.3$\pm$4.0} & 50.3$\pm$7.2 & 28.6$\pm$5.5 & \underline{66.6$\pm$6.7} & 70.1$\pm$2.8 & 95.9$\pm$1.4 & \underline{3.0$\pm$1.5} & \underline{60.8$\pm$4.6} & 4.3$\pm$2.0 \\
LayoutVLM (Cur.) & 11.2$\pm$1.6 & 41.0$\pm$4.0 & 25.6$\pm$6.6 & 14.1$\pm$4.0 & 22.4$\pm$6.1 & \textbf{93.8$\pm$1.9} & \textbf{99.7$\pm$0.2} & 25.9$\pm$4.4 & 19.4$\pm$3.6 & 6.2$\pm$2.1 \\
LayoutVLM (Obj.) & 14.2$\pm$1.7 & 55.5$\pm$4.7 & 34.3$\pm$7.0 & 20.7$\pm$5.0 & 17.4$\pm$5.6 & \underline{91.5$\pm$1.6} & 98.7$\pm$0.8 & 28.9$\pm$3.5 & 8.1$\pm$1.7 & 5.1$\pm$1.4 \\
SceneSmith (Ours) & \textbf{71.1$\pm$13.0} & \textbf{82.9$\pm$3.4} & \textbf{74.4$\pm$6.2} & \textbf{67.6$\pm$5.8} & \textbf{80.6$\pm$5.5} & 83.4$\pm$1.5 & 97.6$\pm$1.1 & \textbf{1.2$\pm$0.6} & \textbf{95.6$\pm$1.7} & \underline{0.2$\pm$0.4} \\
\midrule
\multicolumn{11}{l}{\textit{House-Level}} \\
Holodeck & 81.3$\pm$18.3 & 64.6$\pm$6.8 & 43.2$\pm$27.6 & 30.1$\pm$10.1 & 58.4$\pm$6.3 & 79.4$\pm$3.0 & \textbf{99.6$\pm$0.2} & 3.8$\pm$3.1 & 17.9$\pm$5.6 & \textbf{0.3$\pm$0.2} \\
SceneSmith (Ours) & \textbf{214.1$\pm$60.9} & \textbf{83.3$\pm$6.8} & \textbf{66.8$\pm$24.6} & \textbf{76.6$\pm$9.2} & \textbf{82.6$\pm$5.8} & \textbf{79.9$\pm$2.0} & 96.5$\pm$2.1 & \textbf{0.9$\pm$0.5} & \textbf{79.8$\pm$9.1} & 0.6$\pm$0.3 \\
\bottomrule
\end{tabular}%
}
\end{table*}

\Cref{fig:room_qualitative} shows qualitative comparisons with baselines.
\Cref{tab:user_study} presents human study results. SceneSmith achieves 92.2\% average realism win rate and 91.5\% average prompt faithfulness win rate against all room-level baselines (all $p < 0.001$). The largest margins are against the LayoutVLM variants, while the smallest is against HSM (88.5\% realism, 85.2\% faithfulness). For house-level scenes, SceneSmith achieves 80.3\% realism and 84.7\% faithfulness win rates against Holodeck, the only multi-room baseline.

Among ablations, the most impactful are NotGenerated (63.8\% realism, 67.0\% faithfulness), NoAssetValidation (63.0\%, 62.2\%), and NoObserveScene (61.5\% realism), all showing significant effects. These results demonstrate that generative asset acquisition, asset validation, and visual feedback each contribute meaningfully. NoCritic, NoSpecializedTools, and NoAgentMemory show smaller effects (51--55\% realism) that did not reach significance with our study power; detecting these would require 6-18x more comparisons. Notably, NoCritic achieves similar preference scores while being 70\% cheaper (Appendix~\ref{app:generation_costs}), though it produces 24\% fewer objects. While higher object density enables richer manipulation scenarios for robotics, NoCritic offers a cost-efficient alternative for applications where this trade-off is acceptable.

Table~\ref{tab:sceneeval} presents automated metrics. SceneSmith achieves the best performance on CNT, ATR, OOR, OAR, COL, and STB. Notably, we achieve 2.2x improvement on OOR over the best baseline. The lower ACC and NAV scores are expected given our 3-6x higher object density (71.1 vs 11-23 objects), which inherently reduces free space. The most striking difference is physics quality: SceneSmith achieves 1.2\% collision rate versus 3-29\% for baselines, and 95.6\% stability versus 8-61\%. The remaining collisions are slight penetrations (3.8mm mean depth) that are 3-12x shallower than baselines (Appendix~\ref{app:sceneeval_results}). This is a critical differentiator for robotics as our scenes are simulation-ready without post-hoc correction. House-level results are similar: SceneSmith generates 2.6x more objects (214 vs 81) while maintaining 0.9\% collisions and 79.8\% stability versus Holodeck's 3.8\% and 17.9\%.

Beyond quantitative metrics, SceneSmith produces houses with realistic room connectivity. For example, a generated hotel scene has the entrance leading through the reception, en suite bathrooms accessible only through their associated bedrooms, and all rooms connected via a central hallway. In contrast, Holodeck often generates implausible layouts where an entire hotel might only be reachable through a guest room, or a second bedroom is accessible only through the first bedroom's bathroom (Figure~\ref{fig:hotel_connectivity}). See \Cref{app:human_evaluation,app:automated_evaluation} for additional quantitative and \Cref{app:qualitative_results} for additional qualitative results.

\subsection{Robot Policy Evaluation}

We demonstrate the agentic robot policy evaluation pipeline from \Cref{sec:task_eval}. We evaluate across 100 generated scenes spanning four pick-and-place tasks (three room-level, one house-level). To validate that the pipeline can discriminate between policies of varying quality, we compare a simple model-based policy against a deliberately degraded variant with relaxed motion constraints. The standard policy achieves 16\% success versus 12\% for the degraded variant, demonstrating that the automatic evaluation system can detect meaningful differences in policy quality. We manually verified all 300 evaluator judgments (100 scenes $\times$ 3 states: initial, standard policy, degraded policy) and found 99.7\% agreement with human labels. The single disagreement was an ambiguous case where a fruit landed on the edge of a plate. See Appendix~\ref{app:robot_eval} for details and the \href{https://scenesmith.github.io/}{project page} for representative rollout videos.

\subsection{Robot Simulation Demonstrations}

\begin{figure*}[t]
    \centering
    \newlength{\robotdemoimgwidth}
    \setlength{\robotdemoimgwidth}{\dimexpr\textwidth/3-1.3333pt\relax}
    \begin{tabular}{@{}c@{\hspace{2pt}}c@{\hspace{2pt}}c@{}}
        \multicolumn{3}{@{}c@{}}{\footnotesize RB-Y1 teleoperation}\\[0.05em]
        \includegraphics[width=\robotdemoimgwidth]{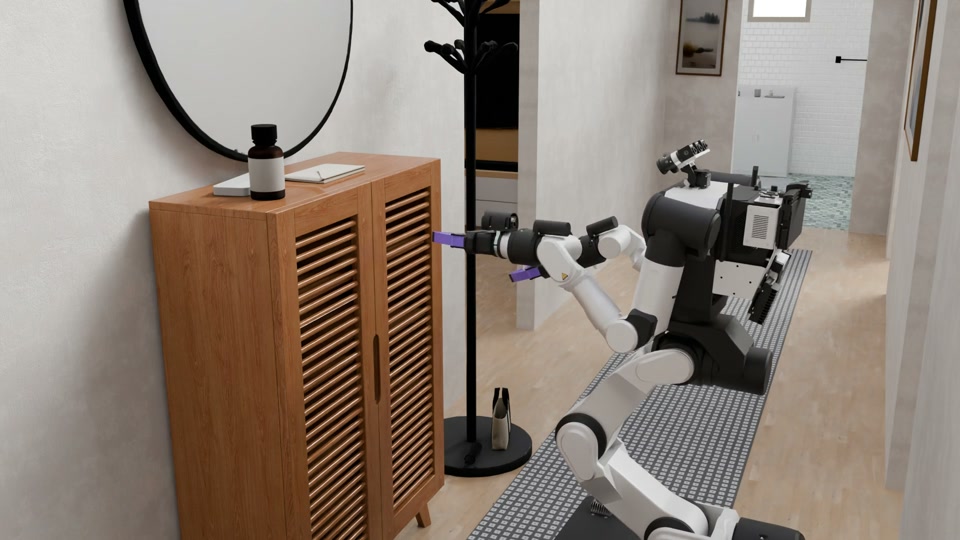} &
        \includegraphics[width=\robotdemoimgwidth]{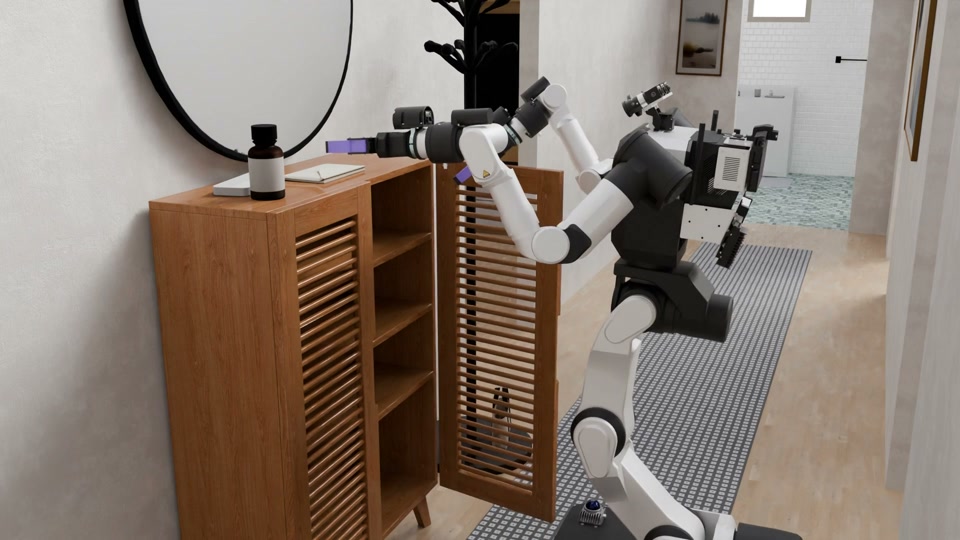} &
        \includegraphics[width=\robotdemoimgwidth]{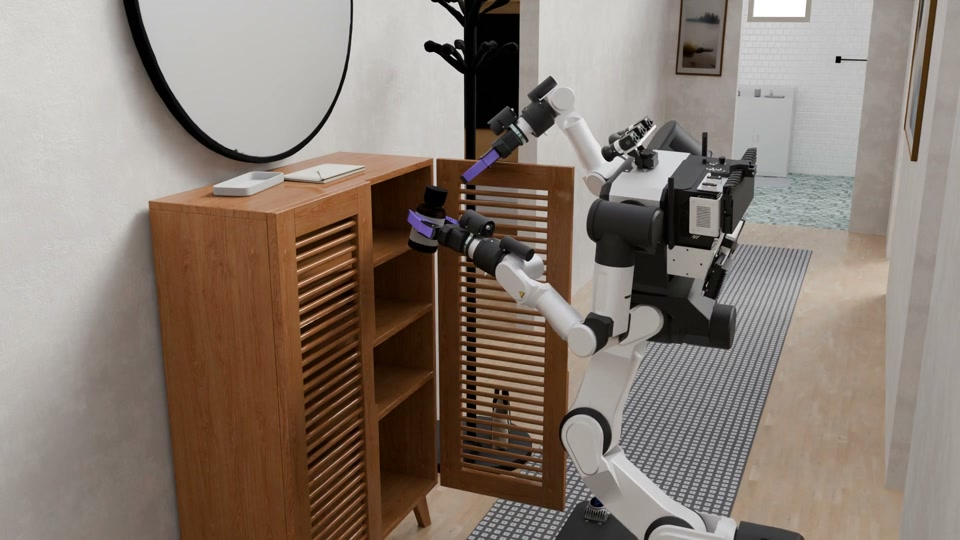}\\[0.25em]
        \multicolumn{3}{@{}c@{}}{\footnotesize Zero-shot policy rollout conditioned on: \textit{``Take the apple from the bowl and place it onto the cutting board.''}}\\[0.05em]
        \includegraphics[width=\robotdemoimgwidth]{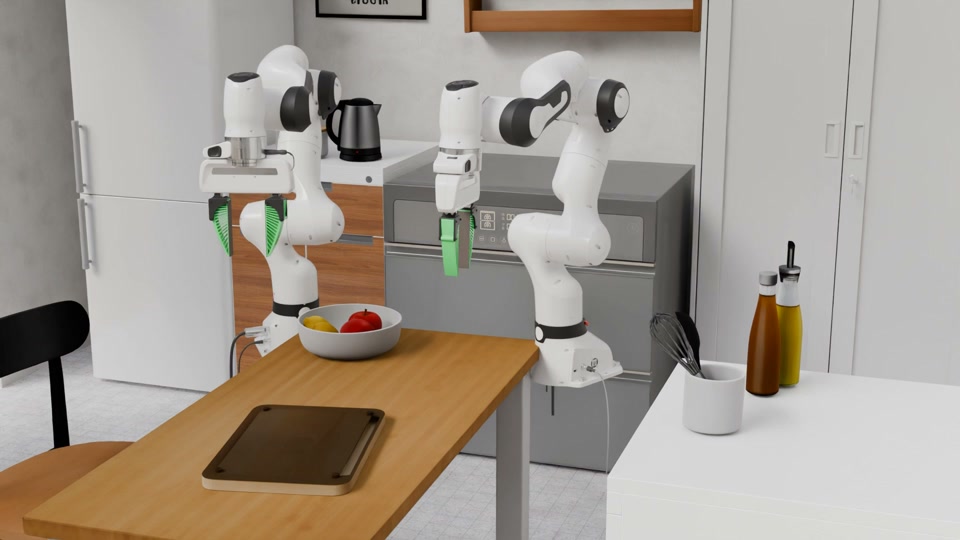} &
        \includegraphics[width=\robotdemoimgwidth]{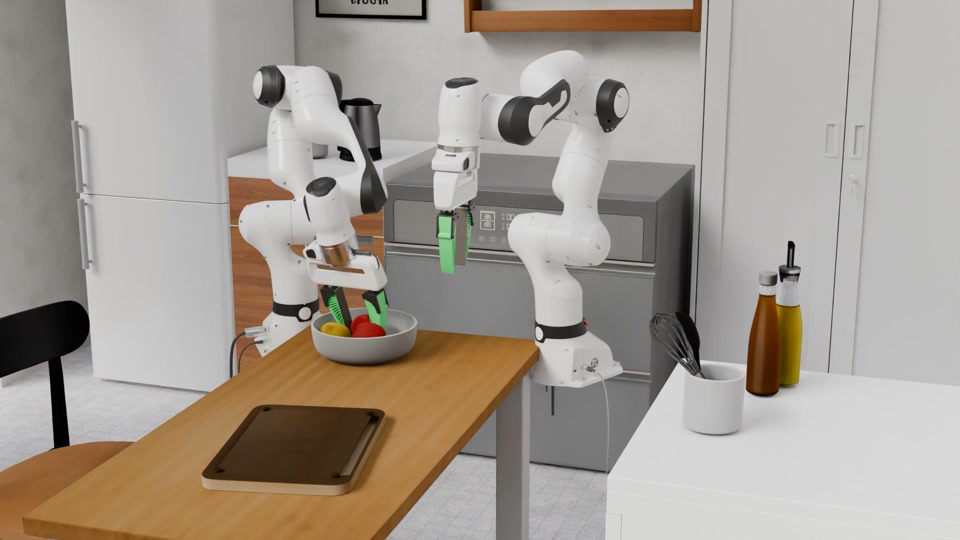} &
        \includegraphics[width=\robotdemoimgwidth]{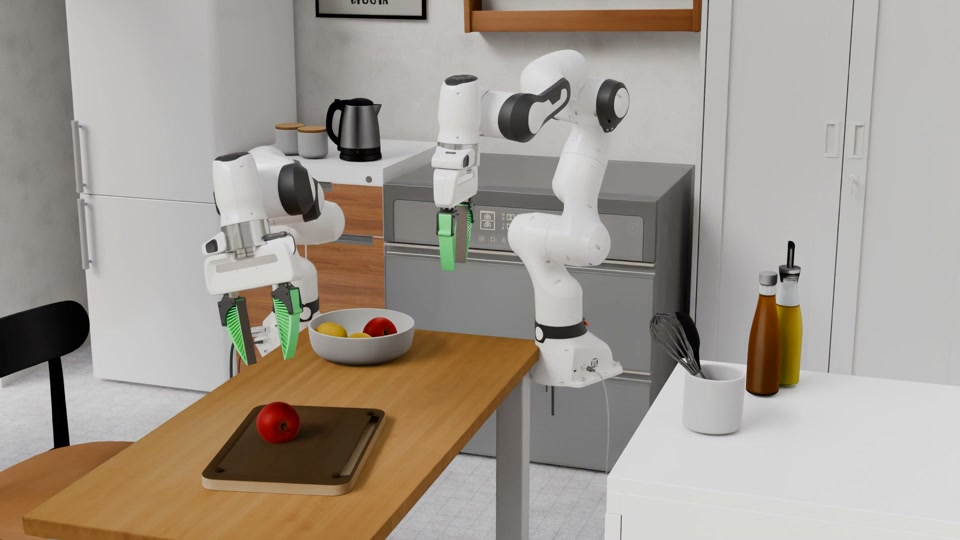}
    \end{tabular}
    \caption{\textbf{Qualitative robot simulation demonstrations.} Each row shows frames sampled from a single trajectory, ordered left to right. \textit{Top:} a teleoperated RB-Y1 opens an articulated cabinet and places a bottle inside. \textit{Bottom:} a policy from prior work~\citep{lin2026systematiccotrain} is executed zero-shot in a SceneSmith scene; conditioned on the shown text instruction, it moves the apple from the bowl to the cutting board. Additional teleoperation and zero-shot videos are provided on the \href{https://scenesmith.github.io/}{project page}.}
    \label{fig:robot_interaction_demos}
\end{figure*}

To demonstrate SceneSmith scenes in complementary robotic interaction settings, we run two qualitative robot simulation workflows. Teleoperation tests whether generated scenes support interactive manipulation and potential data collection in simulation. Zero-shot policy rollouts test whether the scenes are realistic enough for an externally trained generalist robot policy, trained primarily on real-world robot data, to operate without SceneSmith-specific tuning. For teleoperation, \Cref{fig:robot_interaction_demos} (top) shows an RB-Y1 manipulating a cabinet, with additional videos of navigation, articulated manipulation, and mobile pick-and-place on the \href{https://scenesmith.github.io/}{project page}. For zero-shot rollouts, we use the simulation setup from LBM Eval~\citep{lbm_eval2025}, replacing its environments with SceneSmith scenes, and run the text-conditioned robot policy from \citet{lin2026systematiccotrain}. \Cref{fig:robot_interaction_demos} (bottom) shows one representative rollout conditioned on \textit{``Take the apple from the bowl and place it onto the cutting board.''} In this example, the policy recognizes the apple in the bowl and the cutting board, then picks and places the apple as requested. The policy predates SceneSmith and has no SceneSmith rendering, physics, scene, or asset training data.

\section{Conclusion}

We present SceneSmith, a hierarchical agentic framework for generating simulation-ready indoor environments from natural language prompts. SceneSmith decomposes scene construction into hierarchical stages, each driven by designer-critic-orchestrator interactions, producing dense, physically valid scenes that capture the complexity of real homes. Our integrated asset pipeline combines generative text-to-3D synthesis with retrieval for articulated objects, enabling open-vocabulary generation while ensuring simulation readiness. Experiments demonstrate dramatic improvements over baselines: SceneSmith generates 3-6x more objects while achieving $<$2\% inter-object collisions and 95.6\% static stability, compared to 3-29\% collisions and 8-61\% stability for baselines. Human evaluators preferred SceneSmith with a 92\% average win rate for realism and a 91\% average for prompt faithfulness. Qualitative teleoperation and zero-shot policy rollouts further indicate that these generated scenes are suitable for interactive robot simulation beyond static scene-quality metrics. We believe these results mark a point where environment generation is no longer the primary bottleneck for scalable robot training and evaluation in simulation. We hope SceneSmith proves useful for robotics and beyond.

\section*{Acknowledgements}
This work was partially supported by Amazon.com (PO No.~2D-19158853), the Office of Naval Research (ONR) under Grant No.~N00014-22-1-2121, the Toyota Research Institute (TRI), and the National Science Foundation Graduate Research Fellowship Program under Grant No.~2141064. This article reflects only the views of the authors and not those of TRI or any other Toyota entity. Any opinions, findings, and conclusions or recommendations expressed in this material are those of the authors and do not necessarily reflect the views of the National Science Foundation or other funding organizations.
We thank Hongkai Dai, Jeremy Binagia, Peter Werner, and Tom Erez for helpful discussions. We also thank Patrick Farrell for assistance with the videos.

\section*{Impact Statement}
This work aims to advance robot development through improved simulation environments, enabling safer evaluation before real-world deployment. We do not foresee direct negative societal impacts from this research.

\bibliography{ref}
\bibliographystyle{icml2026}

\newpage
\onecolumn
\appendix
\crefalias{section}{appendix}
\crefalias{subsection}{appendix}
\section*{Appendix}
\label{app:start}

\vspace{0.5em}
\noindent\textbf{Contents}
\vspace{0.3em}

\noindent
\begin{tabular}{@{}p{0.85\textwidth}@{}r@{}}
A. System Architecture & \pageref{app:architecture} \\
B. Asset Acquisition & \pageref{app:assets} \\
C. Layout Generation & \pageref{app:layout} \\
D. Furniture Placement & \pageref{app:furniture} \\
E. Wall Object Placement & \pageref{app:wall} \\
F. Ceiling Object Placement & \pageref{app:ceiling} \\
G. Manipuland Placement & \pageref{app:manipuland} \\
H. Physical Feasibility Post-Processing & \pageref{app:physics} \\
I. Stochastic Placement & \pageref{app:stochastic} \\
J. Export and Simulator Compatibility & \pageref{app:export} \\
K. Comparison to Prior Scene Generation Systems & \pageref{app:feature_comparison} \\
L. Task-Driven Robot Evaluation & \pageref{app:robot_eval} \\
M. Evaluation Methodology & \pageref{app:eval_methodology} \\
N. Human Evaluation & \pageref{app:human_evaluation} \\
O. Automated Evaluation & \pageref{app:automated_evaluation} \\
P. Evaluation Prompts & \pageref{app:prompts} \\
Q. User Study Interface Screenshots & \pageref{app:interface_screenshots} \\
R. Additional Qualitative Results & \pageref{app:qualitative_results} \\
S. Limitations and Failure Analysis & \pageref{app:limitations} \\
\end{tabular}

\vspace{1em}

\section{System Architecture}
\label{app:architecture}

SceneSmith employs a hierarchical multi-agent architecture to construct scenes $\mathcal{S} = \{\mathcal{R}_j\}$ through five sequential stages: layout generation, furniture placement, wall-mounted object placement, ceiling-mounted object placement, and manipuland placement. Each stage follows a unified \textbf{Designer-Critic-Orchestrator} pattern.

\subsection{Agent Roles}
\label{app:agent_roles}

\textbf{Designer Agent.} Executes scene modifications through tool calls when invoked by the orchestrator. The designer can observe the scene (via visual renders or structured state queries) and uses placement, modification, and validation tools. Each stage provides a specific tool set; see Tables~\ref{tab:layout_tools}--\ref{tab:manipuland_tools} for complete tool descriptions.

\textbf{Critic Agent.} Evaluates scene quality across multiple dimensions, providing categorical scores (0--10) with written justifications. The critic identifies specific issues and suggests improvements without directly modifying the scene.

\textbf{Orchestrator.} Manages the critique-and-improve loop, coordinating between designer and critic using the tools in Table~\ref{tab:orchestrator_tools}. It handles checkpoint management, enabling rollback when quality degrades, and determines when to terminate based on score thresholds or iteration limits.

\begin{figure}[htbp]
\centering
\includegraphics[width=\textwidth]{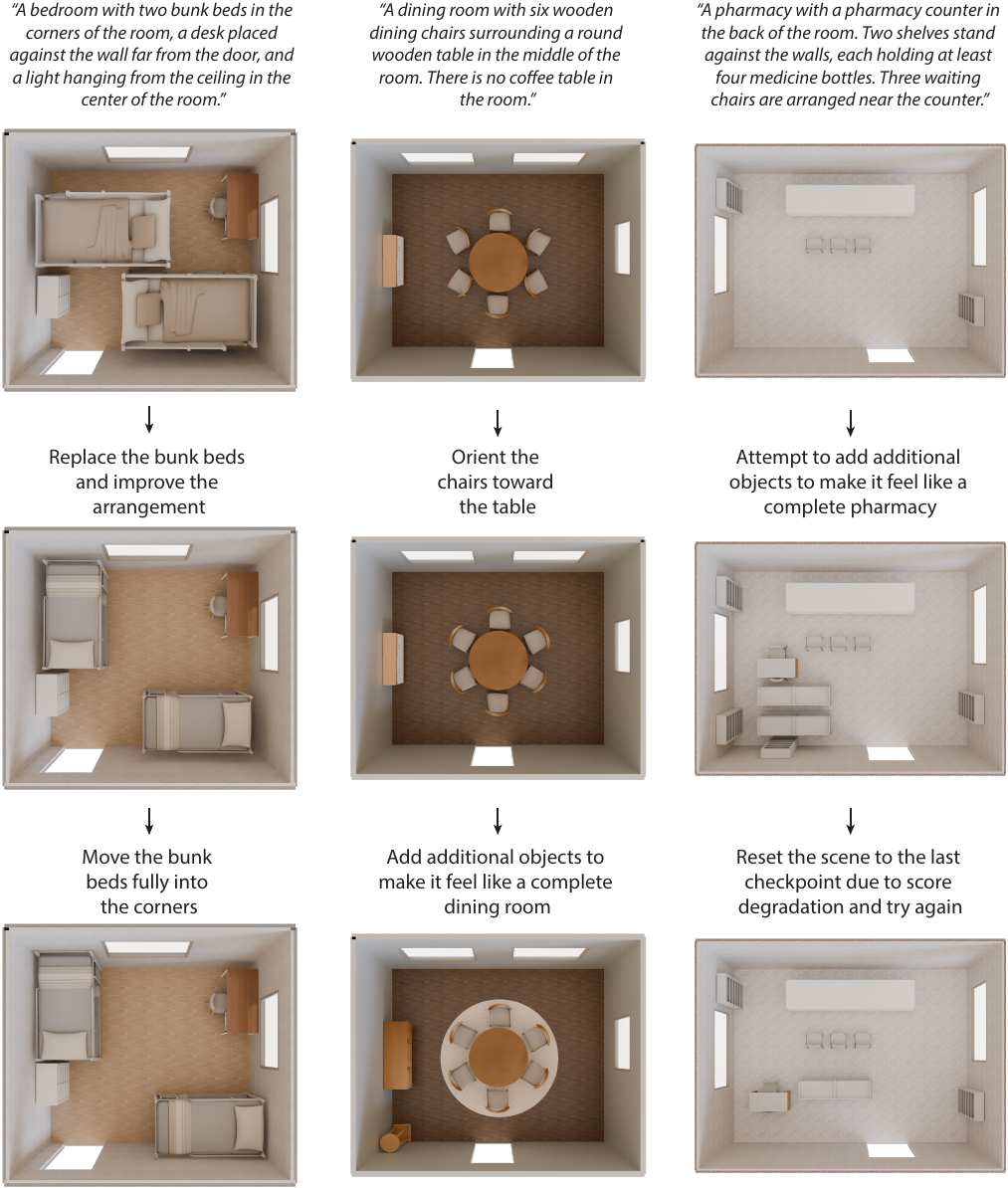}
\caption{\textbf{Designer-critic iteration during furniture placement.} Each column shows a different room evolving through two rounds of critic feedback and designer refinement. \textbf{Top row:} Initial designs after the first designer pass, with scene prompts shown above. \textbf{Middle row:} Scenes after the first critique-and-improve cycle. \textbf{Bottom row:} Scenes after the second cycle. Text annotations describe the changes made at each step. The bedroom (left) progressively improves bunk bed placement. The dining room (center) refines chair orientation and adds furnishings. The pharmacy (right) illustrates checkpoint rollback: when the designer's additions degrade the critic score, the orchestrator resets to the previous checkpoint and prompts a different approach.}
\label{fig:critic_trajectory}
\end{figure}

Figure~\ref{fig:critic_trajectory} illustrates this iterative refinement process during furniture placement, showing how scenes evolve through multiple critique-and-improve cycles.

\begin{table}[htbp]
\centering
\small
\caption{Orchestrator tools (shared across all stages).}
\label{tab:orchestrator_tools}
\begin{tabular}{lp{5.5cm}}
\toprule
\textbf{Tool} & \textbf{Description} \\
\midrule
\texttt{request\_initial\_design} & Invoke designer for initial scene \\
\texttt{request\_design\_change} & Invoke designer for refinement \\
\texttt{request\_critique} & Invoke critic for evaluation \\
\texttt{reset\_scene\_to\_checkpoint} & Revert to previous checkpoint \\
\texttt{select\_placement\_style} & Set natural/perfect placement mode \\
\bottomrule
\end{tabular}
\end{table}

\subsection{Model Configuration}
\label{app:model_config}

All agents and VLM calls use GPT-5.2. Designers and critics use high reasoning effort for thorough analysis, while planners use low reasoning for efficient coordination. Auxiliary VLM tasks (asset routing, physics estimation, validation) use medium reasoning.

\subsection{Scene Observation}
\label{app:observation}

Both designer and critic agents observe scenes through two complementary channels: structured state queries and visual renders.

\textbf{Structured State.} State query tools return spatial metadata for precise reasoning without image interpretation. Object placement stages query the object set $\mathcal{O}_j$ via \texttt{get\_current\_scene\_state}, returning object IDs, surface-local poses, surface assignments, bounding boxes, dimensions, and object descriptions. The layout stage uses \texttt{render\_ascii} to query room layouts and wall segments within $\mathcal{G}_j$.

\textbf{Visual Renders.} The \texttt{observe\_scene} tool produces annotated multi-view renders tailored to each construction stage. Figures~\ref{fig:layout_observe}--\ref{fig:manipuland_observe_articulated} in subsequent sections show representative outputs with stage-specific annotations.

\subsection{Checkpoint and Rollback}
\label{app:checkpoint}

After each critique, the system saves a checkpoint containing complete scene state and scores, maintaining both current and previous checkpoints. If scores decrease beyond thresholds, the orchestrator rolls back to the previous checkpoint and instructs the designer to try an alternative approach (Figure~\ref{fig:critic_trajectory}, right column). Scene state is restored but agent context is preserved, allowing the designer to learn from the failed attempt and pursue a different strategy.

\subsection{Parallel Scene Generation}
\label{app:parallel}

Scene generation parallelizes across multiple GPUs to achieve practical throughput for large-scale dataset creation. A GPU worker pool spawns one text-to-3D worker per available GPU.

The key design enabling throughput is \textbf{batch request pipelining}: scene workers submit multiple asset generation requests simultaneously rather than sequentially. The geometry server processes these requests across available GPUs and streams results back, allowing the scene worker to perform mesh post-processing (collision geometry, physics estimation) on completed assets while the GPUs process remaining requests. This overlap prevents text-to-3D from becoming a bottleneck.

Multiple scenes generate in parallel. In our experiments, we generate 25 scenes concurrently across 8 NVIDIA L40S GPUs; per-scene generation time remains nearly constant as parallel scene count increases, since the shared geometry server efficiently distributes requests across all available GPUs.

\section{Asset Acquisition}
\label{app:assets}

Before placement, simulation-ready assets $\mathcal{A}_i$ are acquired through generation or retrieval, then processed to extract collision geometry and physical properties.

\subsection{Asset Router}
\label{app:asset_router}

An intelligent router uses a VLM to analyze each asset request and performs three tasks:

\textbf{Stage Filtering.} The router rejects items that belong to a different construction stage. For example, a ``coffee mug'' requested during furniture placement is flagged as a manipuland and rejected with feedback to the designer.

\textbf{Composite Decomposition.} Composite requests are split into individually manipulable assets. A ``fruit bowl'' becomes a bowl plus individual fruits; a ``desk with monitor'' separates into furniture (desk) and manipuland (monitor) items for their respective stages. This decomposition is essential for robotic manipulation: a fruit bowl generated as a single fused mesh would be neither realistic nor interactable.

\textbf{Strategy Selection.} For each valid item, the router selects an acquisition strategy:
\begin{itemize}[nosep]
    \item \textbf{Generated}: Text-to-3D generation for standard furniture and objects
    \item \textbf{Articulated}: Retrieval from curated datasets for objects with moving parts (doors, drawers)
    \item \textbf{Thin covering}: Procedural generation for flat textured surfaces (floor rugs, tablecloths, posters)
\end{itemize}

\subsection{Text-to-3D Generation}
\label{app:text_to_3d}

Standard assets are generated using a text-to-image-to-3D pipeline:
\begin{enumerate}[nosep]
    \item Generate reference image from text description using GPT Image 1.5 (1024$\times$1024, solid opaque background for clean segmentation)
    \item Segment foreground object using SAM3~\cite{sam3}
    \item Reconstruct textured 3D surface mesh from single image using SAM3D~\cite{sam3d}
\end{enumerate}

\subsection{Articulated Object Retrieval}
\label{app:articulated_retrieval}

For objects with movable parts (cabinets, drawers, appliances), we retrieve from the ArtVIP library~\cite{artvip} containing pre-authored multi-link models with joint definitions. We evaluated PartNet-Mobility~\cite{sapien} but found its visual and joint quality insufficient for realistic scene generation.

\textbf{Embedding Pre-computation.} To enable semantic similarity search, we pre-compute visual embeddings for each object in the library. Each object is rendered from 8 views (4 upper + 4 lower at 30° elevation) with joints at zero positions. We compute CLIP~\cite{clip} embeddings using ViT-H-14 and average across views, producing 1024-dimensional vectors stored for retrieval.

\textbf{Two-Stage Retrieval.} Following~\cite{pun2026hsm}, we use a two-stage retrieval process. Given a text query and optional target dimensions:
\begin{enumerate}[nosep]
    \item Filter by object type (e.g., floor-standing furniture, wall-mounted, ceiling-mounted, or manipuland)
    \item Rank candidates by CLIP similarity to the text query, selecting top-k
    \item If target dimensions specified, re-rank by L1 distance to desired bounding box
\end{enumerate}

\subsection{Thin Covering Generation}
\label{app:thin_covering}

Thin coverings represent flat textured surfaces such as floor rugs, tablecloths, and wall posters. Floor and manipuland thin coverings are purely decorative elements without collision geometry. The system supports both rectangular and circular shapes, automatically inferred from the object description (e.g., ``round rug'' $\rightarrow$ circular).

\textbf{Texture Modes.} Two texture modes address different use cases:
\begin{itemize}[nosep]
    \item \textbf{Tileable}: Repeating patterns (rugs, fabrics) where the texture tiles across the surface
    \item \textbf{Single image}: Artwork spanning the full surface (posters, paintings) with no repetition
\end{itemize}

\noindent The router's VLM selects the appropriate mode based on semantic understanding of the description (e.g., ``persian rug'' $\rightarrow$ tileable, ``poster depicting detailed dog washing instructions'' $\rightarrow$ single image).

\textbf{Texture Retrieval.} The system first attempts to retrieve a matching PBR texture from a material library (ambientCG\footnote{\url{https://ambientcg.com/}}) using CLIP~\cite{clip} similarity search. Pre-computed embeddings (ViT-H-14, 1024-dimensional) are generated from material preview images.

\textbf{AI-Generated Textures.} When retrieval fails, the system falls back to AI image generation (GPT Image 1.5). Tileable textures are prompted for seamless repeating patterns, while artwork is prompted for full edge coverage without background. Generated images are supplemented with flat normal maps and uniform roughness to form complete PBR materials. For tileable textures, an edge-blending algorithm ensures seamless repetition by linearly blending opposing edges (left-right and top-bottom) over a transition zone.

\subsection{Physical Property Estimation}
\label{app:physics_estimation}

A vision-language model (VLM) analyzes multi-view renders to estimate physical properties.

\subsubsection{Rigid Body Estimation}

The VLM receives 6 views (top and bottom + 4 side views at 30° elevation) and outputs:
\begin{itemize}[nosep]
    \item Dominant material category (from 19 predefined types), used to determine friction coefficients
    \item Mass estimate with confidence range $[m_{\min}, m_{\max}]$. The confidence range enables domain randomization over plausible masses
    \item Canonical orientation: The VLM identifies the up axis and selects which numbered image shows the functional front face (e.g., the side users face for furniture, the decorative face for wall art). This image-selection approach is more robust than directly predicting axis labels.
\end{itemize}

The moment of inertia $\boldsymbol{I}_B \in S_3$ (the set of $3 \times 3$ symmetric matrices) is computed assuming uniform density:
\begin{equation}
    \rho = \frac{m}{V_{\text{mesh}}}, \quad \boldsymbol{I}_B = \rho \cdot \boldsymbol{I}_{\text{unit}}
\end{equation}
where $V_{\text{mesh}}$ is the mesh volume and $\boldsymbol{I}_{\text{unit}}$ is the unit-density inertia tensor from mesh geometry. We found SAM3D meshes to be of sufficient quality that mesh repair operations are not required for these computations.

Based on the VLM-predicted axes, meshes are transformed to canonical Z-up, Y-forward orientation with object-type-specific positioning: floor and manipuland objects have their bottom at $z=0$, ceiling objects have their top at $z=0$, and wall objects have their back at $y=0$. This canonicalization enables $SE(2)$ placement on support surfaces.

\subsubsection{Articulated Body Estimation}

For objects with multiple parts (e.g., cabinets with doors), properties are estimated per-link. The VLM receives combined views of the complete object, isolated views for each articulated link, and overall and per-link bounding box dimensions. It first estimates the total object mass, then estimates per-link materials (for friction coefficients) and masses constrained to sum to this total. Estimating the total first grounds the part estimates in a quantity that is easier to reason about. Per-link moments of inertia $\boldsymbol{I}_B$ are computed independently using each link's mesh geometry and assigned mass.

\subsection{Collision Geometry}
\label{app:collision_geometry}

Visual meshes cannot be used directly for simulation of interactable objects, as most physics engines~\citepalias{drake}~\cite{mujoco} require convex geometry for efficient contact resolution. We decompose visual meshes into convex pieces using V-HACD~\cite{vhacd}. We limit the number of convex pieces per object: 128 for furniture (larger objects requiring more detail), 64 for wall objects and manipulands, and 16 for ceiling objects (rarely involved in collisions). We also evaluated CoACD~\cite{coacd}, which produces fewer convex pieces due to its tree search and thus can result in faster simulations. However, CoACD tends to inflate collision geometry, causing visual meshes to appear floating when objects are stacked. V-HACD does not exhibit this inflation effect. Additionally, we found CoACD to be one to two orders of magnitude slower to compute than V-HACD, which can become a bottleneck when generating scenes with hundreds or thousands of objects.

\subsection{Asset Validation}
\label{app:asset_validation}

\begin{figure}[htbp]
    \centering
    \includegraphics[width=\textwidth]{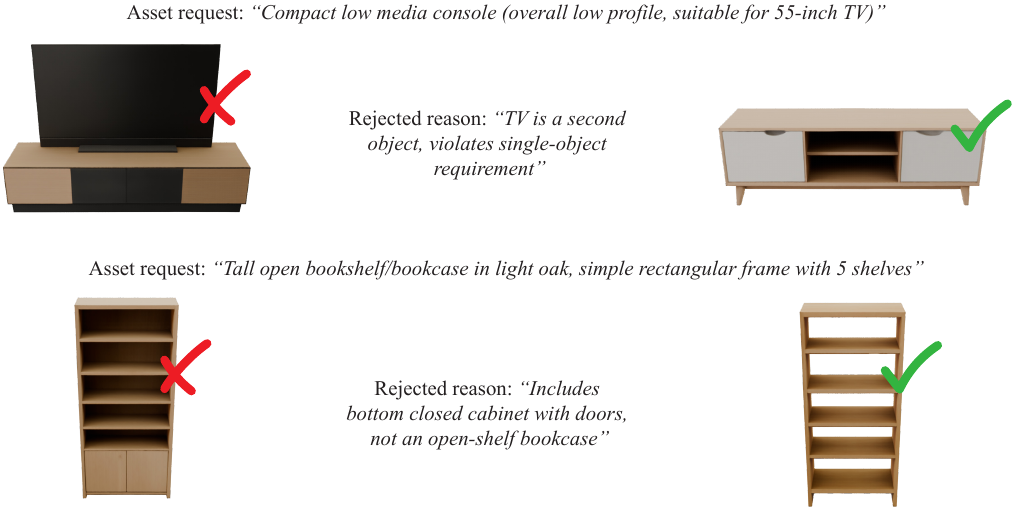}
    \caption{\textbf{Asset validation examples.} Each row shows a text-to-3D request with a rejected (left) and accepted (right) generation. \textbf{Top:} The validator rejects a media console generated with a TV attached, violating the single-object requirement. \textbf{Bottom:} The validator rejects a bookshelf with closed cabinet doors, which does not match the ``open bookshelf'' specification.}
    \label{fig:asset_validation}
\end{figure}

Generated and retrieved assets undergo VLM-based validation to verify suitability for the scene. The VLM checks:
\begin{itemize}[nosep]
    \item \textbf{Object type}: Mesh matches request (e.g., rejecting a stool when a chair was requested)
    \item \textbf{Style consistency}: Colors and materials match specifications (e.g., rejecting a brown chair when red was requested)
    \item \textbf{Single object}: No multiple objects or furniture with manipulands attached (e.g., rejecting a dining table generated with chairs)
    \item \textbf{Completeness}: No missing parts (e.g., rejecting a chair missing a leg)
    \item \textbf{Reasonable proportions}: Not severely distorted
    \item \textbf{Closed state}: Doors and drawers are closed, since generated assets cannot articulate and open doors would remain frozen in place
\end{itemize}
The validator is lenient toward text and labels (which current 3D generation cannot reliably produce) and metallic or transparent materials (which require PBR properties beyond SAM3D). For retrieved assets such as articulated objects or the HSSD asset ablation, we use a more lenient validator that focuses on object type and basic structure while ignoring cosmetic details (color, style, finish) and minor structural variations (e.g., accepting a 4-drawer dresser when 6 drawers were requested). Limited library sizes necessitate accepting close matches rather than exact specifications. Assets that fail validation are regenerated or re-retrieved up to a configurable retry budget. If all retries fail, failure feedback is returned to the designer agent, which can request an alternative object or adjust the scene. Figure~\ref{fig:asset_validation} shows examples of accepted and rejected assets with failure reasons.

\section{Layout Generation}
\label{app:layout}

The layout agent receives the global scene prompt $\mathcal{T}$ and generates architectural geometry $\{\mathcal{G}_j\}_{j=1}^{M}$ for $M$ rooms, including walls, doors, windows, and material assignments. It also derives room-specific prompts $\{\mathcal{T}_j\}_{j=1}^{M}$ that propagate to downstream agents. The designer has access to all tools in Table~\ref{tab:layout_tools} and can invoke them in any order, iterating freely based on its own judgment or critic feedback. This flexibility allows the designer to regenerate the entire layout, adjust individual rooms, or refine architectural details as needed.

\begin{table}[htbp]
\centering
\small
\caption{Layout agent tools. Access: D = Designer, C = Critic.}
\label{tab:layout_tools}
\begin{tabular}{llcp{4.0cm}}
\toprule
\textbf{Tool} & \textbf{Category} & \textbf{Access} & \textbf{Description} \\
\midrule
\texttt{observe\_scene} & Visual & D, C & Top-down render with room labels \\
\texttt{render\_ascii} & Visual & D, C & ASCII floor plan with wall labels \\
\texttt{generate\_room\_specs} & Modification & D & Create rooms with dimensions \\
\texttt{resize\_room} & Modification & D & Adjust room dimensions \\
\texttt{add\_adjacency} & Modification & D & Add room connection \\
\texttt{remove\_adjacency} & Modification & D & Remove room connection \\
\texttt{add\_open\_connection} & Modification & D & Create open floor plan \\
\texttt{remove\_open\_connection} & Modification & D & Remove opening \\
\texttt{set\_wall\_height} & Modification & D & Set wall height \\
\texttt{add\_door} / \texttt{remove\_door} & Modification & D & Manage doors \\
\texttt{add\_window} / \texttt{remove\_window} & Modification & D & Manage windows \\
\texttt{get\_material} & Asset & D & Semantic material search \\
\texttt{set\_room\_materials} & Modification & D & Assign floor/wall materials \\
\texttt{validate} & Feasibility & D, C & Check placement validity \\
\bottomrule
\end{tabular}
\end{table}

\subsection{Room Layout Creation}
\label{app:room_layout}

When creating the room layout via \texttt{generate\_room\_specs}, the designer provides room types, dimensions, adjacency constraints, and a room-specific prompt $\mathcal{T}_j$ for each room $j$. These prompts are derived from the global scene prompt $\mathcal{T}$ by extracting constraints relevant to each room:
\begin{itemize}[nosep]
    \item Room-specific furniture, colors, and features (e.g., ``living room with red sofa'')
    \item House-wide style applied to all rooms (modern, rustic, minimalist)
    \item Resolved ambiguous items: furniture mentioned without clear room assignment is allocated based on room types present (e.g., dining table $\rightarrow$ kitchen when no dining room exists)
\end{itemize}
The room prompts $\mathcal{T}_j$ propagate to all downstream agents (furniture, wall, ceiling, manipuland) to ensure placed objects remain consistent with the original scene description.

\subsection{Room Placement Algorithm}
\label{app:room_placement}

Given room specifications with adjacency constraints, \texttt{generate\_room\_specs} invokes a placement algorithm that finds valid spatial arrangements. The algorithm uses best-first backtracking search:
\begin{enumerate}[nosep]
    \item Sort rooms so adjacency constraints can be evaluated incrementally
    \item Place the first room at the origin
    \item For each subsequent room: generate candidate positions along edges of already-placed rooms, score candidates, and explore in best-first order
    \item On completing a layout, evaluate global score and track the best found
    \item Backtrack to try alternative placements; return best layout found within timeout
\end{enumerate}
The algorithm is complete and optimal within the fixed room ordering and discretized position space. Different orderings may yield different layouts; exploring all orderings is intractable. As an anytime algorithm, it can be run with a timeout and will return the best layout found so far. The following subsections detail each component.

\subsubsection{Room Ordering}

Rooms are sorted so that when placing each room, at least one of its required neighbors (if any) is already placed, allowing adjacency constraints to guide placement. Rooms with no adjacency requirements (\emph{anchors}) are placed first to establish the layout foundation. Rooms with requirements (\emph{connectors}) are placed once at least one required neighbor exists. Within each category, rooms are sorted by area (largest first) to place more constrained rooms earlier.

\subsubsection{Edge-Based Attachment}

New rooms are placed by attaching them to edges of already-placed rooms. The algorithm samples 11 candidate positions along each edge and tests both original and 90° rotated orientations for non-square rooms.

\subsubsection{Placement Scoring}

The algorithm uses two levels of scoring. Local scores rank candidate positions to guide the best-first search, while global scores evaluate complete layouts.

\textbf{Local scoring.} Each candidate is scored to prioritize placements that satisfy adjacency constraints while keeping the layout compact:
\begin{itemize}[nosep]
    \item Base score $S_{\text{base}}$ for all valid placements
    \item Adjacency bonus $w_{\text{adj}}$ per required neighbor that is already placed and adjacent
    \item Centroid penalty $w_{\text{dist}}$ per meter from the centroid of already-placed rooms
\end{itemize}
Candidates violating required adjacencies to already-placed rooms are rejected.

\textbf{Global scoring.} Complete layouts are evaluated on compactness and stability. Compactness measures space efficiency as the ratio of total room area to bounding box area:
\begin{equation}
    S_{\text{compact}} = \frac{A_{\text{rooms}}}{A_{\text{bbox}}}
\end{equation}
Stability rewards preserving room positions during iterative editing:
\begin{equation}
    S_{\text{stable}} = \sum_r \exp\left(-\frac{\|\mathbf{p}_r - \mathbf{p}_r^{\text{prev}}\|_2}{2}\right)
\end{equation}
where $\mathbf{p}_r, \mathbf{p}_r^{\text{prev}} \in \mathbb{R}^2$ are current and previous room centers. This enables incremental refinement. For example, when the designer narrows a central hallway, the modified room must be re-placed and surrounding rooms must shift to maintain adjacency. The stability score ensures all rooms remain as close as possible to their original positions rather than triggering a complete layout redesign.

\subsection{Architectural Elements}
\label{app:architectural_elements}

After establishing room layout, the designer adds architectural elements: wall height, doors (interior connections between rooms and exterior access), windows (on exterior walls for natural lighting), open connections (removing the entire shared wall between adjacent rooms for open floor plans), and materials (floor and wall surfaces per room retrieved via semantic search).

For doors and windows, the designer specifies the wall segment, a discrete position (left, center, or right), and dimensions; windows additionally require sill height. The exact position is sampled uniformly within the specified segment, creating natural variation. The designer can adjust any element at any time, such as resizing a room after adding doors, changing materials after critique feedback, or regenerating the entire layout if the current design is unsatisfactory.

\subsection{Layout Validation}
\label{app:layout_validation}

The \texttt{validate} tool checks two properties. First, it verifies that room placement completed successfully and rooms exist. Second, it validates connectivity: starting from rooms with exterior doors, breadth-first search traverses interior doors and open connections to find all reachable rooms. At least one exterior door must exist, and any room not reachable from the exterior is reported as an error.

\subsection{Critic Evaluation}
\label{app:layout_critic}

The critic evaluates layouts across five categories, each scored 0--10: Room Proportions (appropriate dimensions for room types), Spatial Flow (logical connections and traffic patterns), Natural Lighting (window placement and exterior wall exposure), Material Consistency (materials matching room purposes), and Prompt Following (adherence to the scene description).

\subsection{Observation}
\label{app:layout_observation}

\begin{figure}[htbp]
    \centering
    \includegraphics[width=\textwidth]{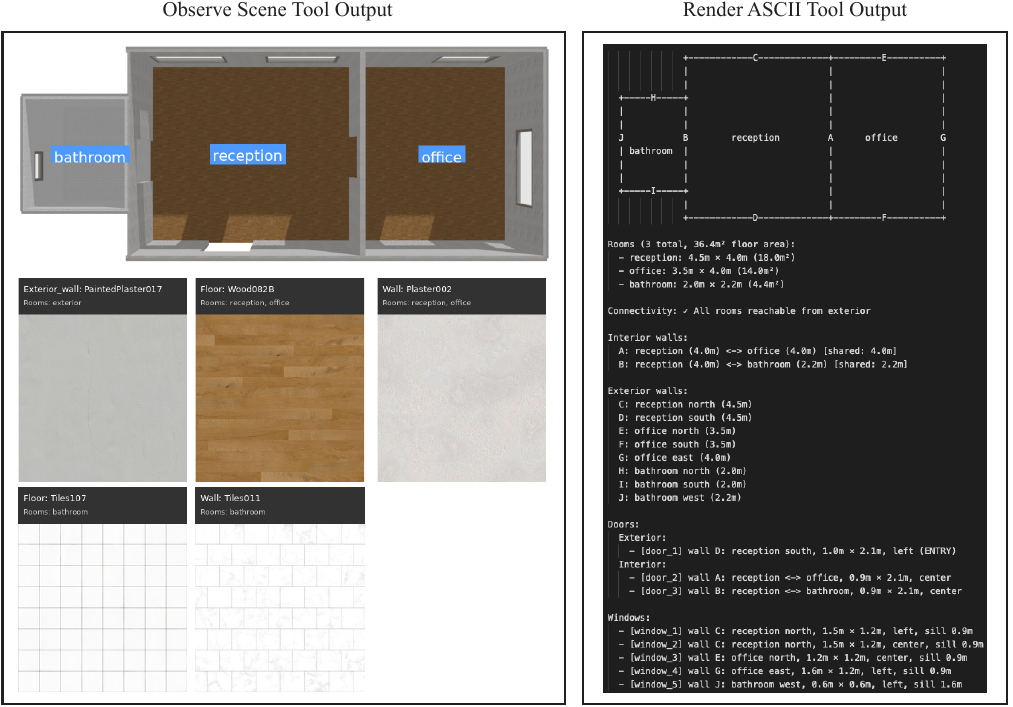}
    \caption{\textbf{Layout agent observation tools.} \textbf{Left:} The \texttt{observe\_scene} tool renders a top-down view with room labels and displays assigned materials with texture swatches. \textbf{Right:} The \texttt{render\_ascii} tool provides an ASCII floor plan with labeled wall segments (A--J) and structured metadata including room dimensions, connectivity validation, and door/window specifications.}
    \label{fig:layout_observe}
\end{figure}

Figure~\ref{fig:layout_observe} shows the layout agent's observation output, including a rendered perspective view and ASCII layout with labeled wall segments for precise door and window placement.

\section{Furniture Placement}
\label{app:furniture}

The furniture agent populates each room $\mathcal{R}_j$ with floor-standing furniture. It receives a room scene containing the architectural geometry $\mathcal{G}_j$ from the layout agent and the room-specific prompt $\mathcal{T}_j$ describing the intended furnishings. Each room is processed independently with a fresh agent instance. This decomposition enables parallel generation, simplifies the task for the agent by limiting scope to a single room, and reduces inference costs by keeping context sizes smaller. Table~\ref{tab:furniture_tools} lists the available tools.

\textbf{Coordinate System.}
The designer specifies furniture poses as $(x, y, \theta)$ tuples in floor coordinates, where $\theta$ is the yaw rotation in degrees. The room coordinate system is centered at the floor origin, with $x \in [-\ell/2, \ell/2]$ and $y \in [-w/2, w/2]$ where $\ell$ and $w$ are the room length and width respectively. Since floor-standing furniture rests on the floor surface, the full $SE(3)$ pose $\mathcal{X}_i$ is constructed by setting $z = 0$ with roll and pitch fixed to zero.

\begin{table}[htbp]
\centering
\small
\caption{Furniture agent tools. Access: D = Designer, C = Critic.}
\label{tab:furniture_tools}
\begin{tabular}{llcp{4.0cm}}
\toprule
\textbf{Tool} & \textbf{Category} & \textbf{Access} & \textbf{Description} \\
\midrule
\texttt{observe\_scene} & Visual & D, C & Multi-view renders with annotations \\
\texttt{get\_current\_scene\_state} & State & D, C & Object positions and bounding boxes \\
\texttt{generate\_assets} & Asset & D & Create 3D furniture from text \\
\texttt{list\_available\_assets} & State & D, C & List generated assets \\
\texttt{add\_furniture} & Modification & D & Place at floor position \\
\texttt{move\_furniture} & Modification & D & Reposition existing furniture \\
\texttt{remove\_furniture} & Modification & D & Remove from scene \\
\texttt{rescale\_furniture} & Modification & D & Uniform scaling \\
\texttt{check\_facing} & Specialized & D, C & Verify orientation relationships \\
\texttt{snap\_to\_object} & Specialized & D & Align and eliminate gaps \\
\texttt{check\_physics} & Feasibility & D, C & Detect collisions \\
\texttt{check\_reachability} & Feasibility & D, C & Verify traversability \\
\bottomrule
\end{tabular}
\end{table}

The following subsections detail the algorithms behind the specialized and feasibility tools.

\subsection{\texttt{check\_facing} Algorithm}
\label{app:check_facing}

Many furniture pieces have functional orientations: chairs should face tables, sofas should face entertainment centers, and storage furniture (wardrobes, dressers) should face away from walls so doors and drawers remain accessible. Visual assessment from rendered views can be unreliable due to camera angle ambiguities. The \texttt{check\_facing} tool provides precise orientation verification and returns the exact rotation needed for alignment.

Given source object $A$ and target $B$, the algorithm determines if $A$ faces toward/away from $B$:

\begin{enumerate}[nosep]
    \item Extract yaw $\theta_A$ from object $A$'s transform
    \item Compute forward direction: $\mathbf{d} = R_z(\theta_A) \cdot [0, 1, 0]^\top$, where $R_z$ denotes rotation about the vertical axis
    \item Select target point on $B$: center for circular objects, closest axis-aligned bounding box (AABB) point to $A$ otherwise
    \item Perform 2D ray-AABB intersection to check if $A$'s forward direction intersects $B$
    \item Compute optimal rotation: $\theta^* = \operatorname{atan2}(\Delta y, -\Delta x)$
\end{enumerate}

Circular objects are detected via volume ratio: $V_{\text{mesh}} / V_{\text{AABB}} < 0.80$, where $V_{\text{mesh}}$ is the mesh volume and $V_{\text{AABB}}$ is the axis-aligned bounding box volume.

\subsection{\texttt{snap\_to\_object} Algorithm}
\label{app:snap_to_object}

Placing furniture against walls or adjacent to other furniture typically requires multiple operations: approximate placement, orientation check, rotation correction, and position adjustment to eliminate gaps. The \texttt{snap\_to\_object} tool combines these into a single operation. It also handles cases like pushing a chair under a desk, where the correct position is difficult to determine from vision alone: 2D bounding boxes appear to intersect in the top-down view even though the 3D geometries do not collide.

The tool supports three orientation modes: \texttt{toward} rotates the source to face the target (e.g., chairs facing tables), \texttt{away} rotates the source to face away from the target (e.g., wardrobes with doors facing into the room), and \texttt{none} performs gap elimination without orientation change.

Given a source object to be moved and a stationary target, the algorithm proceeds in three phases:

\textbf{Phase 1: Collision Resolution.}
If objects overlap, push the source out along the axis with minimum overlap. The source's bounding box is treated as a square using $\max(\text{width}, \text{depth})$ to ensure subsequent rotation will not reintroduce collision.

\textbf{Phase 2: Orientation Alignment.}
For \texttt{toward} and \texttt{away} modes, use \texttt{check\_facing} to compute and apply the optimal yaw rotation. Skipped for \texttt{none} mode.

\textbf{Phase 3: Gap Elimination.}
Iteratively move the source toward the target in small steps, checking for collision at each step. Stop when collision is detected and revert to the last collision-free position, leaving a small margin between objects.

\subsection{\texttt{check\_reachability} Algorithm}
\label{app:reachability}

For generated scenes to be useful in robotics applications, all room areas must remain traversable by the target robot. The robot footprint half-width $h_r$ is a configurable parameter that users set based on their robot platform. This configurability is essential because mobile robots vary significantly in size: a humanoid robot may need only 20cm of clearance, while larger mobile manipulators require substantially more; for example, the Rainbow Robotics RB-Y1\footnote{\url{https://www.rainbow-robotics.com/en_rby1}} has a 60$\times$69cm base, requiring approximately 35cm clearance. By specifying the appropriate footprint, users ensure that generated scenes are actually navigable by their specific robot.

To verify traversability, the algorithm models the robot as a disk $B_{h_r}$ of radius $h_r$ and computes the walkable area via Minkowski sum operations:
\begin{enumerate}[nosep]
    \item Shrink floor polygon: $F' = F \ominus B_{h_r}$
    \item Expand each furniture OBB: $O_i' = O_i \oplus B_{h_r}$
    \item Compute walkable area: $W = F' \setminus \bigcup_i O_i'$
    \item Verify $W$ is a single connected region
\end{enumerate}
where $\ominus$ denotes Minkowski difference (erosion) and $\oplus$ denotes Minkowski sum (dilation). Erosion shrinks the floor boundary inward so the robot center cannot get too close to walls, while dilation expands furniture outward to account for the robot's physical extent.

\textbf{Blocking Identification:} For each furniture piece, the tool tests if removing it reduces the number of disconnected regions. Pieces that block connectivity are identified as potential repositioning candidates.

\textbf{Feedback to Agent:} The tool returns whether the room is fully reachable, the number of disconnected regions, the reachability ratio (largest region area divided by total walkable area), and the list of blocking furniture IDs. This feedback enables the designer to identify and reposition problematic pieces.

\subsection{\texttt{check\_physics} Tool}
\label{app:check_physics}

The \texttt{check\_physics} tool detects geometric and functional issues that would make the scene infeasible. During furniture placement, the tool reports:

\begin{itemize}[nosep]
    \item \textbf{Furniture collisions}: Interpenetration between furniture pieces or with walls, detected via signed distance queries on collision geometry using Drake~\citepalias{drake}.
    \item \textbf{Floor covering overlaps}: Overlapping rugs or mats, detected via 2D oriented bounding box intersection since these objects lack collision geometry.
    \item \textbf{Floor covering boundary}: Rugs extending beyond room walls.
    \item \textbf{Door blockage}: Furniture blocking the clearance zone in front of doors, which would prevent passage.
    \item \textbf{Open connection blockage}: Furniture blocking open connections between rooms, verified by checking if a robot-sized passage remains clear.
    \item \textbf{Window clearance} (warning): Furniture placed in front of windows above sill height. Reported as a warning for the agent to judge whether the placement is acceptable (e.g., a desk slightly overlapping a window may be fine, while a tall wardrobe fully blocking it is problematic).
\end{itemize}

Issues are reported with penetration depths and object identifiers, enabling the designer to make targeted corrections.

\subsection{Critic Evaluation}
\label{app:furniture_critic}

The critic evaluates furniture placement across six categories, each scored 0--10: Realism (natural placement patterns, scene collisions), Functionality (items support intended activities with correct facing relationships), Layout (logical arrangement with proper spacing and clear traffic flow), Holistic Completeness (room feels appropriately furnished for its type and size), Prompt Following (literal adherence to specified furniture and quantities), and Reachability (all areas remain traversable by the robot). Holistic Completeness and Prompt Following are complementary: the former evaluates whether the room is appropriately furnished based on common expectations for the room type, while the latter checks whether explicitly requested items are present. For example, a prompt ``living room with a chair'' would score high on Prompt Following if the chair is placed, but low on Holistic Completeness since living rooms typically contain more furniture than a single chair.

\subsection{Observation}
\label{app:furniture_observation}

The \texttt{observe\_scene} tool renders the room from multiple viewpoints: one top-down view and four corner views. For corner views, walls between the camera and room center are hidden to reveal the interior. Inspired by set-of-mark prompting~\cite{yang2023setofmark} and SceneWeaver~\cite{yang2025sceneweaver}, the top-down view includes annotations to support spatial reasoning:
\begin{itemize}[nosep]
    \item Coordinate grid with $(x, y)$ position markers and coordinate frame
    \item Bounding boxes around each furniture piece
    \item Object labels with unique identifiers
    \item Direction arrows showing the forward-facing direction of each object
    \item Door, window, and open connection labels on architectural elements
\end{itemize}
Corner views include coordinate markers for spatial reference.

\begin{figure}[htbp]
    \centering
    \includegraphics[width=\textwidth]{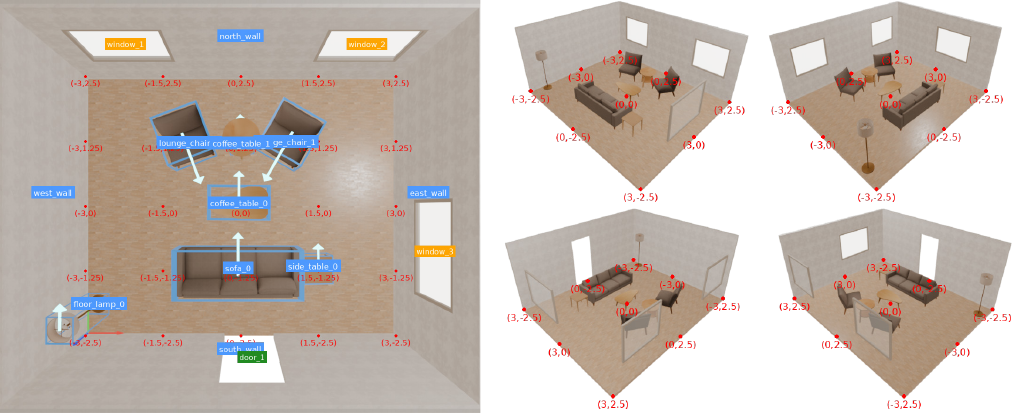}
    \caption{\textbf{Furniture agent \texttt{observe\_scene} output.} \textbf{Left:} Top-down view with coordinate grid, coordinate frame, and labeled furniture showing bounding boxes and direction arrows. Walls, doors, and windows are labeled for placement reference. \textbf{Right:} Four corner views with coordinate markers for depth perception and alignment.}
    \label{fig:furniture_observe}
\end{figure}

Object identifiers combine a human-readable name with a base-36 sequential suffix (e.g., \texttt{chair\_0}, \texttt{chair\_a}, \texttt{chair\_10}), reducing visual clutter while remaining easy for the agent to reference. Figure~\ref{fig:furniture_observe} shows an example observation.

\section{Wall Object Placement}
\label{app:wall}

The wall agent adds wall-mounted objects $(\mathcal{A}_i, \mathcal{X}_i)$ to $\mathcal{O}_j$, including art, mirrors, shelves, and clocks. It receives the room-specific prompt $\mathcal{T}_j$ and the room with furniture already placed. Each room is processed independently. Table~\ref{tab:wall_tools} lists the available tools.

\textbf{Coordinate System.}
Each wall surface in the room has its own local coordinate frame and a unique identifier. Wall objects use SE(2) coordinates $(x, z, \theta)$ within a specified wall's frame: $x$ runs along the wall from start (0) to end (wall length in meters), $z$ runs vertically from floor (0) to ceiling height, and $\theta$ specifies rotation in degrees around the wall normal. Since objects mount flush to the wall surface, the $y$-coordinate is implicitly zero. The agent populates all walls jointly during a single session, specifying both the target wall identifier and the local pose for each placement. The system automatically excludes regions occupied by doors, windows, and open connections, preventing invalid placements.

\begin{table}[htbp]
\centering
\small
\caption{Wall agent tools. Access: D = Designer, C = Critic.}
\label{tab:wall_tools}
\begin{tabular}{llcp{4.0cm}}
\toprule
\textbf{Tool} & \textbf{Category} & \textbf{Access} & \textbf{Description} \\
\midrule
\texttt{observe\_scene} & Visual & D, C & Multi-view renders with annotations \\
\texttt{get\_current\_scene\_state} & State & D, C & Wall surfaces and placed objects \\
\texttt{list\_wall\_surfaces} & State & D, C & Walls with excluded regions \\
\texttt{generate\_wall\_assets} & Asset & D & Create wall-mounted assets \\
\texttt{list\_available\_assets} & State & D, C & List generated assets \\
\texttt{place\_wall\_object} & Modification & D & Place using wall-local SE(2) \\
\texttt{move\_wall\_object} & Modification & D & Reposition (can change walls) \\
\texttt{remove\_wall\_object} & Modification & D & Remove from scene \\
\texttt{rescale\_wall\_object} & Modification & D & Uniform scaling \\
\texttt{check\_physics} & Feasibility & D, C & Detect collisions \\
\bottomrule
\end{tabular}
\end{table}

\subsection{Physics Validation}
\label{app:wall_physics}

The \texttt{check\_physics} tool detects geometric issues specific to wall placements:
\begin{itemize}[nosep]
    \item \textbf{Wall object collisions}: Interpenetration between wall-mounted objects
    \item \textbf{Boundary violations}: Objects extending beyond wall edges or above ceiling height
    \item \textbf{Excluded region violations}: Placement overlapping door, window, or open connection regions
\end{itemize}

\subsection{Critic Evaluation}
\label{app:wall_critic}

The critic evaluates wall object placement across five categories, each scored 0--10: Realism (objects placed at realistic heights for their purpose), Functionality (objects accessible and not blocked by furniture), Layout (distribution across walls with appropriate spacing from openings), Holistic Completeness (walls appropriately decorated for room type), and Prompt Following (requested wall items present).

\subsection{Observation}
\label{app:wall_observation}

The \texttt{observe\_scene} tool provides two complementary views: a top-down context view showing the full room layout, and per-wall orthographic views for precise placement. Each wall view includes:
\begin{itemize}[nosep]
    \item Coordinate grid with position markers showing wall-local $(x, z)$ coordinates and coordinate frame
    \item Wall surface identifier for unambiguous wall references
    \item Doors, windows, and open connections visible as openings in the wall
    \item Wall object labels with unique identifiers
    \item Nearby furniture rendered for spatial context (unlabeled to reduce clutter)
\end{itemize}

\begin{figure}[htbp]
    \centering
    \includegraphics[width=\textwidth]{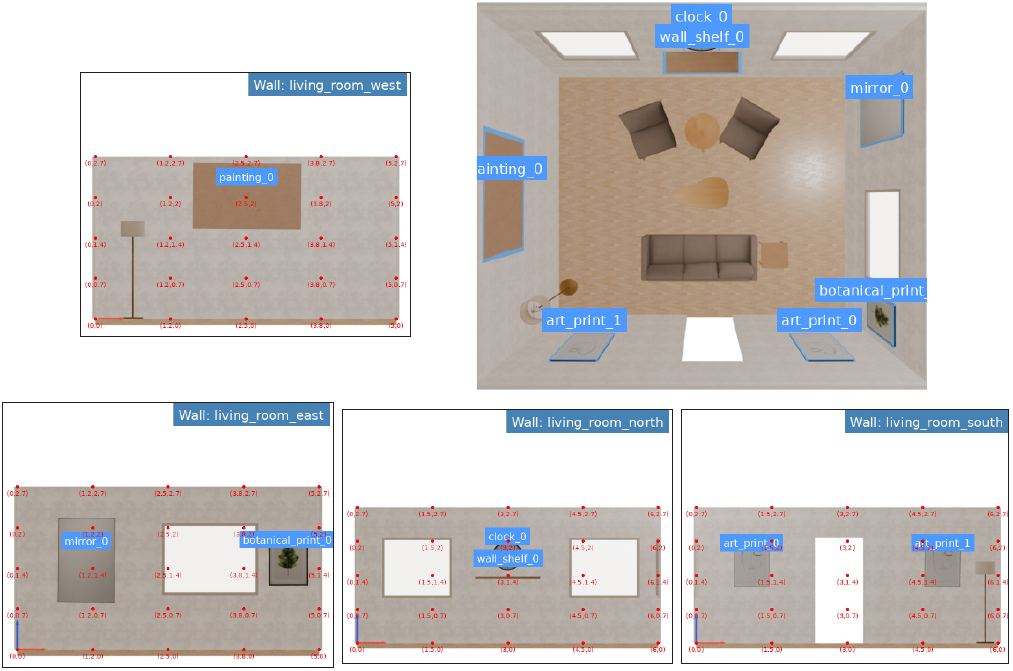}
    \caption{\textbf{Wall agent \texttt{observe\_scene} output.} \textbf{Top-right:} Top-down context view showing the full room with labeled wall objects. \textbf{Remaining panels:} Orthographic views of each wall surface with coordinate grids, wall identifiers, door/window openings, and object labels. Nearby furniture is rendered for spatial context.}
    \label{fig:wall_observe}
\end{figure}

Figure~\ref{fig:wall_observe} shows an example observation.

\section{Ceiling Object Placement}
\label{app:ceiling}

The ceiling agent adds ceiling-mounted objects $(\mathcal{A}_i, \mathcal{X}_i)$ to $\mathcal{O}_j$, including lights, fans, and chandeliers. It receives the room-specific prompt $\mathcal{T}_j$ and the room with furniture and wall objects already placed. Each room is processed independently. Table~\ref{tab:ceiling_tools} lists the available tools.

\textbf{Coordinate System.}
Ceiling objects use room-local SE(2) coordinates $(x, y, \theta)$ where $x$ and $y$ specify position within the room bounds and $\theta$ is the yaw rotation in degrees. Since fixtures mount to the ceiling surface, the $z$-coordinate is implicitly fixed at ceiling height.

\begin{table}[htbp]
\centering
\small
\caption{Ceiling agent tools. Access: D = Designer, C = Critic.}
\label{tab:ceiling_tools}
\begin{tabular}{llcp{4.0cm}}
\toprule
\textbf{Tool} & \textbf{Category} & \textbf{Access} & \textbf{Description} \\
\midrule
\texttt{observe\_scene} & Visual & D, C & Multi-view renders with annotations \\
\texttt{get\_current\_scene\_state} & State & D, C & Room bounds and ceiling objects \\
\texttt{generate\_ceiling\_assets} & Asset & D & Create ceiling-mounted assets \\
\texttt{list\_available\_assets} & State & D, C & List generated assets \\
\texttt{place\_ceiling\_object} & Modification & D & Place using room-local SE(2) \\
\texttt{move\_ceiling\_object} & Modification & D & Reposition fixture \\
\texttt{remove\_ceiling\_object} & Modification & D & Remove from scene \\
\texttt{rescale\_ceiling\_object} & Modification & D & Uniform scaling \\
\texttt{check\_physics} & Feasibility & D, C & Detect collisions \\
\bottomrule
\end{tabular}
\end{table}

\subsection{Physics Validation}
\label{app:ceiling_physics}

The \texttt{check\_physics} tool detects collisions between ceiling fixtures and other objects in the scene.

\subsection{Critic Evaluation}
\label{app:ceiling_critic}

The critic evaluates ceiling fixture placement across four categories, each scored 0--10: Realism (natural fixture placement centered over functional areas), Functionality (adequate lighting coverage for the room), Layout (symmetry, spacing, and clearance from tall furniture), and Prompt Following (requested fixtures present).

\subsection{Observation}
\label{app:ceiling_observation}

The \texttt{observe\_scene} tool renders one top-down view and two side views. Annotations include:
\begin{itemize}[nosep]
    \item Coordinate grid at ceiling height with $(x, y)$ position markers
    \item Furniture and wall-mounted objects rendered for spatial context (unlabeled to reduce clutter)
    \item Placed ceiling fixtures labeled with unique identifiers
\end{itemize}

\begin{figure}[htbp]
    \centering
    \includegraphics[width=\textwidth]{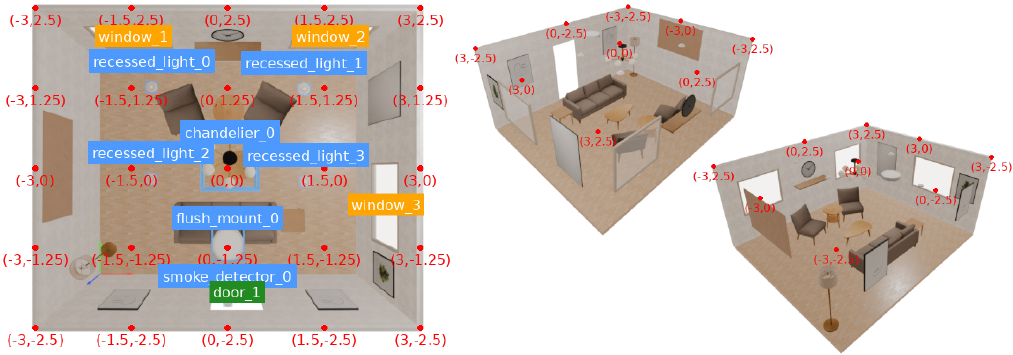}
    \caption{\textbf{Ceiling agent \texttt{observe\_scene} output.} \textbf{Left:} Top-down view with coordinate grid and labeled ceiling fixtures. \textbf{Right:} Two corner views with coordinate markers for verifying fixture heights and clearance from tall furniture.}
    \label{fig:ceiling_observe}
\end{figure}

Figure~\ref{fig:ceiling_observe} shows an example observation.

\section{Manipuland Placement}
\label{app:manipuland}

The manipuland agent adds small manipulable objects $(\mathcal{A}_i, \mathcal{X}_i)$ to $\mathcal{O}_j$, creating the dense arrangements and fine-grained clutter characteristic of real indoor environments. Specialized tools enable realistic object configurations including vertical stacks, random piles, and filled containers (Section~\ref{app:composite_tools}). Objects are placed on support surfaces of existing furniture, wall-mounted objects (e.g., shelves), and architectural elements (e.g., floors). Before agent execution, a VLM-based analysis step (Section~\ref{app:furniture_selection}) selects which entities should receive manipulands and derives entity-specific prompts $\mathcal{T}_{j,k}$ for each supporting entity $k$. Each supporting entity is then processed independently with a fresh agent instance. As described in the main text (Section~\ref{sec:scene_rep}), this forms the second level of branching in the tree-structured construction process (the first being the house-to-room decomposition used by furniture, wall, and ceiling agents), enabling parallel generation and reducing context size. Table~\ref{tab:manipuland_tools} lists the available tools.

\begin{table}[htbp]
\centering
\small
\caption{Manipuland agent tools. Access: D = Designer, C = Critic.}
\label{tab:manipuland_tools}
\begin{tabular}{llcp{4.0cm}}
\toprule
\textbf{Tool} & \textbf{Category} & \textbf{Access} & \textbf{Description} \\
\midrule
\texttt{observe\_scene} & Visual & D, C & Multi-view renders with annotations \\
\texttt{get\_current\_scene\_state} & State & D, C & Surface bounds and placed objects \\
\texttt{list\_support\_surfaces} & State & D, C & Surfaces with area and clearance \\
\texttt{generate\_manipuland\_assets} & Asset & D & Create small object assets \\
\texttt{list\_available\_assets} & State & D, C & List generated assets \\
\texttt{place\_manipuland} & Modification & D & Place using surface-local SE(2) \\
\texttt{move\_manipuland} & Modification & D & Reposition object \\
\texttt{remove\_manipuland} & Modification & D & Remove from scene \\
\texttt{rescale\_manipuland} & Modification & D & Uniform scaling \\
\texttt{create\_stack} & Specialized & D & Vertical stacking \\
\texttt{fill\_container} & Specialized & D & Fill cavity with objects \\
\texttt{create\_arrangement} & Specialized & D & Controlled placement on container \\
\texttt{create\_pile} & Specialized & D & Random scattered arrangement \\
\texttt{check\_physics} & Feasibility & D, C & Detect collisions \\
\bottomrule
\end{tabular}
\end{table}

\textbf{Coordinate System.}
Each support surface on an entity has its own local 2D coordinate frame and a unique identifier (e.g., \texttt{S\_0}, \texttt{S\_1}, \texttt{S\_a}). Manipulands use SE(2) coordinates $(x, y, \theta)$ within the specified surface's frame: $x$ runs left-to-right across the surface, $y$ runs front-to-back, and $\theta$ specifies rotation in degrees around the surface normal. The origin $(0, 0)$ is located at the surface center. Since objects rest on the surface, the $z$-coordinate is determined by the surface height. The agent populates all surfaces of an entity jointly during a single session (e.g., all shelf levels of a bookcase), specifying both the target surface identifier and the local pose for each placement. Surface boundaries are validated automatically, ensuring objects remain within the support region.

\subsection{Entity Selection and Prompt Derivation}
\label{app:furniture_selection}

A VLM analyzes the room prompt $\mathcal{T}_j$ and rendered furniture and wall-mounted objects to determine which entities should receive manipulands. For each selected entity, the VLM derives three structured outputs that together form the entity-specific prompt $\mathcal{T}_{j,k}$:
\begin{itemize}[nosep]
    \item \textbf{Suggested items}: Objects to place, distinguished as required (explicitly mentioned in $\mathcal{T}_j$) or optional (contextually appropriate)
    \item \textbf{Prompt constraints}: Explicit requirements extracted from $\mathcal{T}_j$, including quantity constraints (e.g., ``three books'') and negative constraints (e.g., ``only a laptop'')
    \item \textbf{Style notes}: Placement guidance affecting density and aesthetic (e.g., minimalist, cluttered, organized)
\end{itemize}
The same VLM call also identifies nearby context furniture (e.g., chairs around a dining table) that informs placement decisions, such as positioning place settings in front of chairs or orienting user-facing objects toward seating. This context furniture is included in visual observations to help guide manipuland placement.

\subsection{Support Surface Detection}
\label{app:support_surfaces}

We use the support surface identification algorithm from HSM~\cite{pun2026hsm} to extract horizontal placement surfaces from furniture and wall object meshes. For articulated objects, we merge all link geometries into a single mesh at zero joint positions (closed drawers/doors), ensuring correct clearance computation.

\subsection{Physics Validation}
\label{app:manipuland_physics}

The \texttt{check\_physics} tool detects collisions relevant to the current placement session:
\begin{itemize}[nosep]
    \item \textbf{Manipuland-manipuland collisions}: Interpenetration between manipulands on the current entity
    \item \textbf{Manipuland-furniture collisions}: Manipulands extending into nearby furniture
\end{itemize}
The tool applies context-aware filtering to report only violations the agent can address, excluding collisions involving objects from other entities. Violations are reported with object identifiers, enabling targeted corrections.

\subsection{Critic Evaluation}
\label{app:manipuland_critic}

The critic evaluates manipuland placement across five categories, each scored 0--10: Realism (natural placement patterns on surfaces), Functionality (items support intended activities), Layout (logical arrangement with visual balance), Holistic Completeness (surfaces appropriately populated for the furniture type), and Prompt Following (required items placed with constraints respected).

\subsection{Observation}
\label{app:manipuland_observation}

\begin{figure}[htbp]
    \centering
    \includegraphics[width=\textwidth]{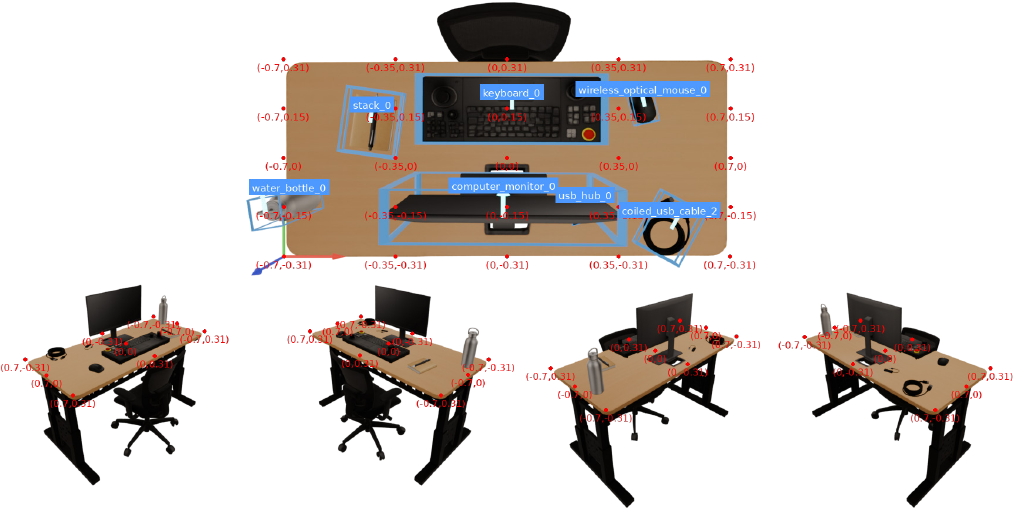}
    \caption{\textbf{Manipuland agent observation for single-surface entity.} \textbf{Top:} Top-down view with coordinate grid, surface-local position markers, and coordinate frame. Manipulands are labeled with bounding boxes. \textbf{Bottom:} Four side views with coordinate markers. The office chair is included as context furniture to inform object orientation.}
    \label{fig:manipuland_observe_single}
\end{figure}

\begin{figure}[htbp]
    \centering
    \includegraphics[width=\textwidth]{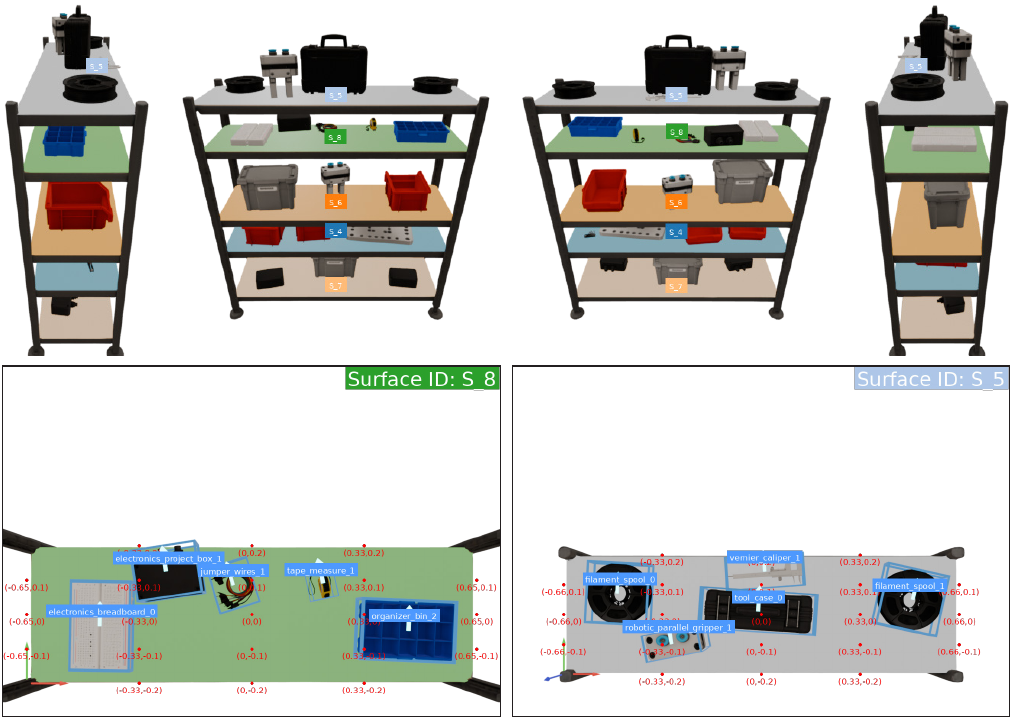}
    \caption{\textbf{Manipuland agent observation for multi-surface entity.} \textbf{Top:} Four side views of a shelving unit with color-coded surface identifiers (S\_4 through S\_8). \textbf{Bottom:} Top-down renders for individual surfaces (two of five shown), colored to match the side views. The agent populates all surfaces jointly in a single session.}
    \label{fig:manipuland_observe_multi}
\end{figure}

\begin{figure}[htbp]
    \centering
    \includegraphics[width=\textwidth]{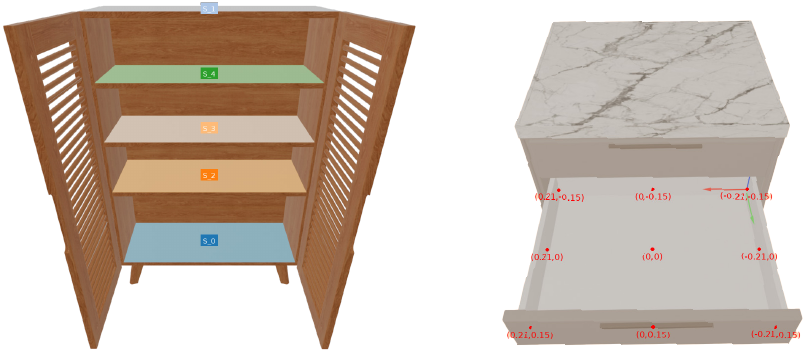}
    \caption{\textbf{Manipuland agent observation for articulated entities.} \textbf{Left:} Doors are opened during rendering to reveal interior surfaces. \textbf{Right:} For entities with drawers (prismatic joints), separate renders show each drawer individually to avoid occlusion from stacked drawers (one of multiple drawer renders shown).}
    \label{fig:manipuland_observe_articulated}
\end{figure}

The \texttt{observe\_scene} tool renders focused views of the current entity (furniture or wall-mounted object). The base output includes four side views with coordinate markers. Additional views depend on entity type:
\begin{itemize}[nosep]
    \item \textbf{Single-surface}: One top-down view with coordinate grid, surface-local $(x, y)$ position markers, and coordinate frame.
    \item \textbf{Multi-surface}: One top-down render per surface, each with coordinate grid and surface identifier. The agent populates all surfaces jointly in a single session.
    \item \textbf{Articulated}: Doors are opened during rendering to reveal interior surfaces. For entities with drawers, one render per drawer shows only that drawer open to avoid occlusion from stacked drawers.
\end{itemize}
Context furniture identified during entity selection (Section~\ref{app:furniture_selection}) may be included to inform object orientation (e.g., chairs around a dining table). Placed manipulands are labeled with unique identifiers, bounding boxes, and direction arrows. Figures~\ref{fig:manipuland_observe_single}, \ref{fig:manipuland_observe_multi}, and~\ref{fig:manipuland_observe_articulated} show example observations.

\subsection{Composite Object Tools}
\label{app:composite_tools}

\begin{figure}[htbp]
    \centering
    \includegraphics[width=\textwidth]{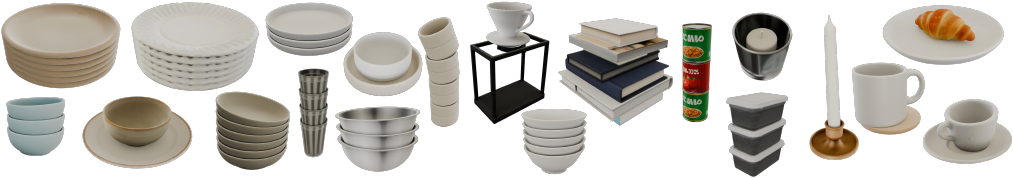}
    \caption{\textbf{Examples of the \texttt{create\_stack} tool.} Stacks include same-type objects (plates, bowls, cups), mixed objects (books, storage containers), and object-on-object arrangements (cup on saucer, candle on holder, croissant on plate). Physics simulation settles objects into stable configurations.}
    \label{fig:composite_stack}
\end{figure}

\begin{figure}[htbp]
    \centering
    \includegraphics[width=\textwidth]{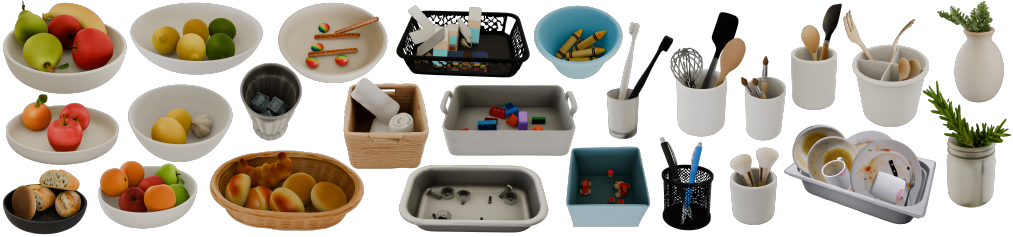}
    \caption{\textbf{Examples of the \texttt{fill\_container} tool.} Containers include fruit bowls, utensil crocks, pencil cups, toy bins, vases with plants, and bread baskets. Physics simulation with iterative retry settles objects naturally within container cavities.}
    \label{fig:composite_fill}
\end{figure}

\begin{figure}[htbp]
    \centering
    \includegraphics[width=\textwidth]{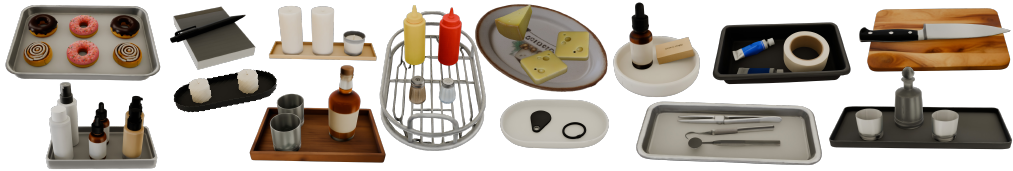}
    \caption{\textbf{Examples of the \texttt{create\_arrangement} tool.} Items are placed at agent-specified positions on flat containers: baking trays, cheese boards, condiment caddies, candle trays, bathroom trays, and cutting boards. Unlike \texttt{fill\_container}, this tool provides precise control over item placement.}
    \label{fig:composite_arrangement}
\end{figure}

\begin{figure}[htbp]
    \centering
    \includegraphics[width=\textwidth]{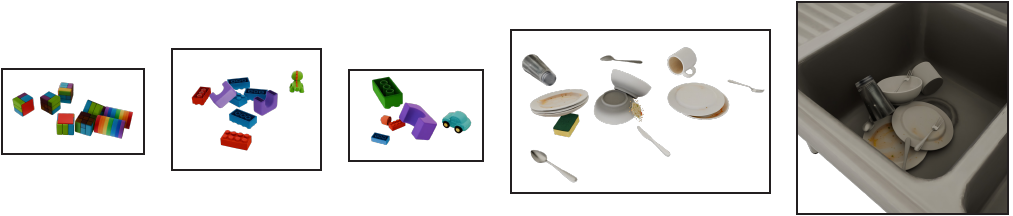}
    \caption{\textbf{Examples of the \texttt{create\_pile} tool.} \textbf{Left:} Floor piles with random positions and orientations: scattered toys and a dish mess on a kitchen floor. \textbf{Right:} Dirty dishes in a sink basin. The sink is built into the counter, so \texttt{fill\_container} cannot be used since it requires a separate container asset.}
    \label{fig:composite_pile}
\end{figure}

The specialized composite tools (Table~\ref{tab:manipuland_tools}) create multi-object arrangements using physics simulation. These tools are essential for creating interesting arrangements, as the standard placement tool only supports $SE(2)$ poses on surfaces. Each tool assembles objects into a composite that is placed at an $SE(2)$ pose on the support surface. In the final scene, individual objects are stored independently with full $SE(3)$ poses computed from the surface transform and each object's pose within the composite. Figures~\ref{fig:composite_stack}, \ref{fig:composite_fill}, \ref{fig:composite_arrangement}, and~\ref{fig:composite_pile} show examples of each tool's output.

\subsubsection{\texttt{create\_stack}}
Vertically stacked objects (e.g., books, plates, mug on coaster, candle on plate). Takes a list of objects (same or different types) and stacks them bottom-to-top in input order:
\begin{enumerate}[nosep]
    \item \textbf{Initial placement}: Compute initial transforms using collision geometry bounding box heights. Objects are positioned sequentially along the Z-axis based on their vertical extents.
    \item \textbf{Physics simulation and stability check}: Run Drake simulation to settle objects into static equilibrium. Objects may nest (e.g., stacked bowls) or shift laterally. Objects exceeding a displacement threshold are marked unstable; if any object falls, the entire stack is rejected and the tool reports how many items remained stable, enabling informed retry with fewer items.
    \item \textbf{Clearance validation}: Verify the final stack height fits the available surface clearance.
\end{enumerate}

\subsubsection{\texttt{fill\_container}}
Fills containers with multiple objects using physics simulation (e.g., fruit bowls, pencil cups, utensil crocks, vases with flowers, baskets). Takes a container asset and list of fill assets:
\begin{enumerate}[nosep]
    \item \textbf{Cavity detection}: Identify the container interior using a top-rim heuristic. Vertices in the top fraction of the container height are projected to the XY plane, and their convex hull defines the opening. The hull is scaled inward to provide a safety margin from container walls.
    \item \textbf{Object rotation}: Objects are rotated based on aspect ratio. Elongated objects (aspect ratio above threshold) are oriented to stand upright with their longest axis vertical, aligned with the container's length direction. Asymmetric objects like utensils undergo a thick-end-up check that compares top and bottom half footprints, flipping the object if the bottom is thicker (ensuring spatula heads face up in utensil crocks).
    \item \textbf{Spawn position generation}: For each fill object, sample an XY position within the interior hull using rejection sampling. Objects are spawned at staggered heights above the rim, with spacing based on each object's rotated height to prevent initial overlap.
    \item \textbf{Physics simulation}: Run Drake simulation to let objects fall and settle naturally within the container. Objects may shift, nest, or fall out depending on geometry and packing.
    \item \textbf{Inside/outside classification}: After simulation, objects are classified based on their final Z position relative to a configured threshold. Objects below the threshold are considered to have fallen out.
    \item \textbf{Iterative retry}: Objects that fall out are respawned in subsequent iterations with new random positions. The loop continues until all objects settle inside or the iteration limit is reached. Previously settled objects remain in place during retries but may be pushed out by new objects, in which case they are added back to the retry list.
\end{enumerate}
The tool succeeds as long as at least one object remains inside; objects that cannot settle after the maximum iterations are removed from the final composite. The tool reports which objects remained and which were removed, enabling the agent to retry with different object combinations if needed.

\subsubsection{\texttt{create\_arrangement}}
Places items at agent-specified positions on flat containers (e.g., place settings on placemats, cheese on cheeseboards, items on serving trays, toiletries on bathroom trays). Unlike \texttt{fill\_container} which uses random positions, this tool provides precise control over item placement. Takes a container asset and list of items with their local $(x, y, \theta)$ coordinates relative to the container center:
\begin{enumerate}[nosep]
    \item \textbf{Position validation}: Determine whether the container is circular or rectangular using mesh volume ratio analysis, then verify each item's center position lies within bounds. For circular containers, check Euclidean distance from center against the radius; for rectangular containers, check axis-aligned containment against half-extents. Only the center position is validated, allowing elongated items (e.g., forks, knives) to be placed near edges when oriented parallel to the edge.
    \item \textbf{Pre-collision check}: Before physics simulation, detect item-item collisions using Drake's signed distance queries. If any items overlap, the tool fails immediately with feedback identifying the colliding pair and penetration depth.
    \item \textbf{Physics Z-settling}: Spawn items above the container surface at their specified XY positions, then run Drake simulation to let them settle under gravity. Items may shift slightly as they come to rest.
    \item \textbf{All-or-nothing validation}: After simulation, classify items by final Z position. If any item falls off the container (Z below threshold), the entire arrangement fails. The tool reports which items fell and the container bounds, enabling the agent to adjust positions closer to the center.
\end{enumerate}
This all-or-nothing semantic ensures complete, coherent arrangements. There is no retry mechanism; if items fall off, the agent must explicitly retry with adjusted positions.

\subsubsection{\texttt{create\_pile}}
Serves two primary use cases: (1) random scattered arrangements on surfaces (e.g., paperclips on a desk, scattered toys, piles of laundry), and (2) placing objects into built-in furniture cavities that cannot be used with \texttt{fill\_container} (e.g., dirty dishes in a sink basin), since \texttt{fill\_container} requires a separate container asset. For built-in cavities, the cavity geometry naturally keeps objects inside during physics simulation. Takes a list of assets (minimum 2, same or different types):
\begin{enumerate}[nosep]
    \item \textbf{Spawn radius computation}: Compute the average bounding box diagonal across all objects and multiply by a configured factor to determine the spawn radius, ensuring objects spread naturally without excessive gaps.
    \item \textbf{Position and rotation generation}: For each object, sample an XY position uniformly from a disk of the spawn radius, with a square root transform on the radial component ensuring uniform area coverage. Orientations are sampled uniformly at random over SO(3).
    \item \textbf{Staggered Z spawning}: Objects spawn at staggered heights above the surface, with spacing based on each object's bounding box diagonal. This prevents initial overlap before physics settles the arrangement.
    \item \textbf{Physics simulation}: Run Drake simulation to let objects fall and settle under gravity. Objects may shift, tumble, nest, or fall off the surface depending on geometry and initial positions.
    \item \textbf{Surface classification}: After simulation, classify objects by final Z position. Objects below a configured threshold are considered to have fallen off the placement surface.
    \item \textbf{Minimum object validation}: The tool requires at least 2 objects to remain on the surface. If fewer than 2 stay on, the tool fails with feedback suggesting the agent try placing the pile further from edges or using fewer objects.
\end{enumerate}
Unlike \texttt{fill\_container}, there is no iterative retry. The tool returns which objects remained on the surface and which fell off, enabling informed retry with different configurations.

\section{Physical Feasibility Post-Processing}
\label{app:physics}

While SceneSmith's agentic construction process encourages physically reasonable placements through iterative feedback, agents are not required to satisfy physical constraints exactly during generation. Generated scenes may contain inter-object penetrations or objects in configurations that are not statically stable. To ensure environments are directly simulatable, we apply a two-stage post-processing pipeline after furniture placement and again after manipuland placement. Wall-mounted and ceiling-mounted objects do not require post-processing as they are welded to the world and remain fixed under simulation.

\subsection{Non-Penetration Projection}
\label{app:projection}

We resolve inter-object penetrations by projecting object poses to the nearest collision-free configuration using nonlinear optimization. Following \citet{pfaff2025_steerable_scene_generation}, we solve:
\begin{equation}
    \min_{\mathbf{p}} \|\mathbf{p} - \mathbf{p}_0\|_2^2 \quad
    \text{s.t.} \quad d(i,j) \ge \epsilon, \quad \forall i \neq j,
\end{equation}
where $\mathbf{p}$ and $\mathbf{p}_0$ denote the optimized and original object translations, $d(i,j)$ is the signed distance between objects $i$ and $j$ computed via Drake's collision checker, and $\epsilon = 10^{-5}$m provides a small separation tolerance. Orientations remain fixed during projection to prevent rotational adjustments from destabilizing objects that would otherwise tip over during subsequent gravity simulation. The optimization is solved using SNOPT~\cite{snopt} with Drake's inverse kinematics framework.

\subsection{Gravity Settling}
\label{app:gravity_settling}

After projection resolves penetrations, we simulate the scene under gravity using Drake. Objects settle into statically stable configurations.

\subsection{Pipeline Integration}
\label{app:physics_pipeline}

The pipeline applies projection followed by gravity settling at three points: (1) after furniture placement, where all furniture is free to translate; (2) after each supporting entity's manipulands are placed, using a subset scene containing only that entity and its manipulands with the entity welded; and (3) after all manipuland placement completes, where all furniture is welded and only manipulands are optimized.

\subsection{Fallen Object Removal}
\label{app:fallen_objects}

Physics simulation may cause objects to tip over or fall off surfaces. The system automatically detects and removes fallen objects using type-specific criteria:

\textbf{Furniture}: Compute tilt angle between the object's Z-axis and world vertical. Objects exceeding a threshold are marked as fallen and removed.

\textbf{Manipulands}: Two detection methods identify fallen small objects: (1) floor penetration detection removes objects that clip through the floor, indicating physics instability; (2) surface departure detection removes manipulands that started on furniture but ended near floor level with significant downward displacement, indicating they fell off their intended support surface.

\section{Stochastic Placement}
\label{app:stochastic}

We observed that agents tend to select round position and rotation values when placing objects (e.g., exactly 1.0m from a wall, precisely 90° rotation), resulting in unnaturally precise arrangements compared to real-world scenes. To address this, we apply Gaussian noise to object placements, effectively treating placement as noisy actions, creating realistic variation.

\subsection{Mode Selection}

The orchestrator analyzes prompt keywords to select placement style:
\begin{itemize}[nosep]
    \item \textbf{Natural}: Keywords like ``lived-in'', ``cozy'', ``casual''
    \item \textbf{Perfect}: Keywords like ``pristine'', ``showroom'', ``gallery''
\end{itemize}

Default is natural mode if no keywords match. In practice, we use small noise values (1--3cm position, 0.5--3° rotation for natural mode) to introduce subtle imperfections without disrupting functional arrangements; see Table~\ref{tab:noise_profiles} for the full noise profiles.

\begin{table}[htbp]
\centering
\small
\caption{Gaussian noise standard deviations ($\sigma$) for stochastic placement by object type and placement style.}
\label{tab:noise_profiles}
\begin{tabular}{llcc}
\toprule
\textbf{Object Type} & \textbf{Parameter} & \textbf{Natural $\sigma$} & \textbf{Perfect $\sigma$} \\
\midrule
Furniture & Position XY & 3 cm & 1 mm \\
Furniture & Yaw & 1.0° & 0.1° \\
Wall & Position X (along wall) & 2 cm & 1 mm \\
Wall & Position Z (height) & 1 cm & 1 mm \\
Wall & Rotation & 0.5° & 0.1° \\
Ceiling & Position XY & 2 cm & 1 mm \\
Ceiling & Yaw & 0.5° & 0.1° \\
Manipuland & Position XY & 1 cm & 1 mm \\
Manipuland & Yaw & 3.0° & 0.1° \\
\bottomrule
\end{tabular}
\end{table}

\subsection{Style-Aware Evaluation}

The critic receives placement style to calibrate expectations. Natural style accepts subtle imperfections (e.g., dining chairs at slightly varied angles). Perfect style expects precise alignment within tolerances.

\section{Export and Simulator Compatibility}
\label{app:export}

SceneSmith performs simulator-in-the-loop scene creation using Drake as its native simulation backend. Collision detection, non-penetration projection, and gravity settling are all performed in Drake throughout the generation pipeline, ensuring physical consistency during construction. Scenes are natively represented using SDFormat with Drake model directives.

For compatibility with other simulators, SceneSmith can export scenes to MJCF and USD formats. These exports include visual geometry, collision geometry, physical properties (mass, center of mass, moment of inertia, friction), and articulated joints for articulated objects. The exported formats work with most common robotics simulators including MuJoCo~\citep{mujoco}, PyBullet~\citep{coumans2021}, Isaac Sim\footnote{\url{https://developer.nvidia.com/isaac/sim}}, and Genesis\footnote{\url{https://genesis-embodied-ai.github.io/}}. Blender export is also supported for visualization and rendering, containing visual geometry only.

\section{Comparison to Prior Scene Generation Systems}
\label{app:feature_comparison}

Table~\ref{tab:feature_comparison} summarizes high-level capabilities of SceneSmith and representative prior indoor scene-generation systems.

\newcommand{\featureyes}{\textcolor{green!45!black}{\ensuremath{\checkmark}}}
\newcommand{\featureno}{\textcolor{red!70!black}{\ensuremath{\times}}}

\begin{table}[htbp]
\centering
\caption{Feature comparison with prior scene-generation systems. Green checkmarks indicate support; red crosses indicate no support.}
\label{tab:feature_comparison}
\small
\resizebox{\textwidth}{!}{%
\begin{tabular}{lccccccc}
\toprule
\textbf{Feature} & \textbf{Ours} & \textbf{Holodeck} & \textbf{HSM} & \textbf{SceneWeaver} & \textbf{I-Design} & \textbf{LayoutVLM} & \textbf{ProcTHOR} \\
\midrule
Novel static asset generation & \featureyes & \featureno & \featureno & \featureno & \featureno & \featureno & \featureno \\
Language-conditioned generation & \featureyes & \featureyes & \featureyes & \featureyes & \featureyes & \featureyes & \featureno \\
Multi-room scenes & \featureyes & \featureyes & \featureno & \featureno & \featureno & \featureno & \featureyes \\
Floor/wall materials and textures & \featureyes & \featureyes & \featureno & \featureno & \featureno & \featureno & \featureyes \\
Separable multi-object compositions & \featureyes & \featureno & \featureno & \featureno & \featureno & \featureno & \featureno \\
Articulated objects & \featureyes & \featureno & \featureno & \featureno & \featureno & \featureno & \featureno \\
Estimated physical properties & \featureyes & \featureno & \featureno & \featureno & \featureno & \featureno & \featureno \\
Multiple simulator exports & \featureyes & \featureno & \featureno & \featureno & \featureno & \featureno & \featureno \\
Iterative critique/refinement & \featureyes & \featureno & \featureno & \featureyes & \featureno & \featureno & \featureno \\
\bottomrule
\end{tabular}}
\end{table}

``Novel static asset generation'' refers to generating new static object assets on demand rather than selecting them from a fixed retrieval library. ``Separable multi-object compositions'' refers to arrangements such as fruit bowls, plate stacks, and utensil crocks where each component remains individually manipulable rather than fused into a single mesh. ProcTHOR is included as a widely used procedural scene-generation system.

\section{Robot Policy Evaluation}
\label{app:robot_eval}

\begin{figure}[htbp]
    \centering
    \includegraphics[width=\textwidth]{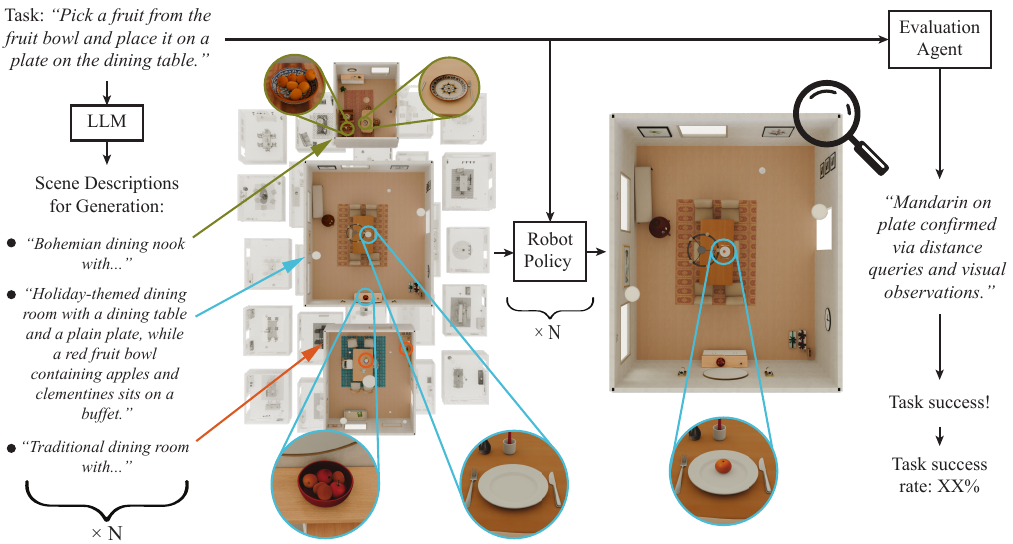}
    \caption{\textbf{Robot manipulation evaluation pipeline.} Given a manipulation task (e.g., ``Pick a fruit from the fruit bowl and place it on a plate''), an LLM generates diverse scene prompts specifying scene constraints implied by the task. SceneSmith generates scenes from each prompt. A robot policy attempts the task in simulation, and an evaluation agent verifies success using simulator state queries and visual observations. This enables scalable policy evaluation without manual environment or success predicate design.}
    \label{fig:robot_eval}
\end{figure}

This section provides implementation details for the robot policy evaluation pipeline described in Section~\ref{sec:task_eval} of the main text. Figure~\ref{fig:robot_eval} illustrates the pipeline.

\subsection{Scene Prompt Generation}
\label{app:scene_prompt_gen}

Given a natural-language task description (e.g., ``pick a fruit and place it on a plate''), an LLM generates diverse scene prompts that enable the task without pre-solving it. The generator first analyzes the task to extract:
\begin{itemize}[nosep]
    \item \textbf{Room requirement:} Whether specific room types are required (explicit: ``in the kitchen''), implied (``dining table'' suggests dining room), or flexible. House-level tasks may require multiple rooms.
    \item \textbf{Required objects:} Objects that must be present (e.g., fruit, plate).
    \item \textbf{Initial state constraint:} What must \emph{not} be true initially. For example, for ``place fruit on plate,'' the fruit must not already be on the plate.
    \item \textbf{Flexible dimensions:} Aspects that can vary across prompts (style, object variants, additional furniture).
\end{itemize}
The generator then produces $N$ diverse scene prompts describing a room with the required objects in non-goal positions plus contextual objects for diversity. Prompts vary across style (modern, rustic, minimalist), object variants (apple vs.\ banana vs.\ orange), and placement configurations. Each prompt is passed to SceneSmith to generate a complete scene.

\subsection{Evaluator Agent}
\label{app:evaluator_agent}

After each policy execution, an evaluator agent determines task success. The agent is provided with the task description and final scene state, then uses two categories of tools to gather evidence:
\begin{itemize}[nosep]
    \item \textbf{State tools:} Geometric queries including object listing, signed distance computation between object pairs, and support verification (contact detection and surface overlap percentage).
    \item \textbf{Vision tools:} Multi-view renders of the scene or specific objects, providing top-down and side views for visual verification of spatial relationships. These reuse the rendering infrastructure from SceneSmith's scene observation tools.
\end{itemize}
The agent exhaustively tests candidate object pairs that could satisfy the task requirement, using state tools for efficiency and falling back to vision tools for ambiguous cases such as containment relationships. This approach avoids hand-crafted success predicates and generalizes across diverse task specifications. Table~\ref{tab:evaluator_tools} lists the available tools.

\begin{table}[htbp]
\centering
\small
\caption{Evaluator agent tools.}
\label{tab:evaluator_tools}
\begin{tabular}{llp{5.5cm}}
\toprule
\textbf{Tool} & \textbf{Category} & \textbf{Description} \\
\midrule
\texttt{list\_objects} & State & List all objects with positions \\
\texttt{get\_object\_info} & State & Object dimensions and orientation \\
\texttt{get\_distance} & State & Signed distance between object pair \\
\texttt{get\_spatial\_relation} & State & Vertical/horizontal gaps, overlap \\
\texttt{get\_support} & State & Contact detection and surface overlap \\
\texttt{observe\_scene} & Vision & Multi-view renders of entire scene \\
\texttt{observe\_objects} & Vision & Focused renders of specific objects \\
\bottomrule
\end{tabular}
\end{table}

\subsection{Example Policy}
\label{app:example_policy}

To demonstrate the pipeline, we implement a model-based pick-and-place policy using a KUKA LBR iiwa14 mounted on a mobile base with a telescoping lift joint. The policy first uses a VLM to parse the task into structured components: a goal predicate (\emph{on}, \emph{inside}, or \emph{near}), target and reference object categories, and optional preconditions. The VLM then matches these categories to specific objects in the scene, producing ranked candidate bindings that are attempted in order. A true generalist policy could operate directly from the natural language task without this structured interface.

Our policy is given a segmented point cloud of each candidate object, and iteratively samples antipodal grasps \citep[\S 5.4]{manipulation} until one is found such that the gripper is collision-free.
It then solves an inverse kinematics optimization problem with collision-avoidance constraints to determine a complete robot posture for the pick configuration \citep[\S 6.1]{manipulation}.
To determine the place configuration, gripper poses are sampled within the target region and a similar inverse kinematics problem is solved.
It also solves for a heuristic pre-grasp and pre-place configuration, where the gripper pose is moved away from the manipuland along the direction of grasping.
These steps are interleaved and run repeatedly until success or timeout.

Planning is performed in segments, connecting (in order) the robot's start configuration, a pre-grasp configuration, the grasp configuration, the post-grasp (same as pre-grasp) configuration, the pre-place configuration, and the place configuration.
Planning is performed using RRT-Connect \citep{kuffner2000rrtconnect} with randomized shortcutting \citep[\S6.4.1]{sekhavat1998shortcutting}.
During the post-grasp to pre-place segment, we require that the trajectory have greater clearance of obstacles around the gripper, to help prevent dropping the manipuland.
These path segments are post-processed with time-optimal path parameterization \citep{verscheure2009topp} to ensure the resulting trajectory satisfies the velocity, acceleration, and torque limits of the robot.
We additionally enforce spatial velocity and acceleration limits on the pose of the gripper, to prevent sudden movements that may cause the robot's grasp to slip.

Finally, we create a deliberately-inferior policy to demonstrate that our agentic policy evaluation can be used for performance comparisons.
The inferior policy does not have the gripper pose velocity and acceleration limits, making it more likely to drop the object.
It also does not have the increased gripper clearance enforced in the planning stage, making the manipuland more likely to collide with the environment.

\subsection{Experimental Results}
\label{app:robot_results}

\textbf{Tasks.}
We evaluate on four pick-and-place tasks, with 25 generated scenes per task:
\begin{enumerate}[nosep]
    \item Pick a fruit from the fruit bowl and place it on a plate on the dining table.
    \item Pick a coke can from the shelf and place it on the table.
    \item Pick a cup from the floor and place it in the sink.
    \item Bring me the water bottle from the kitchen and place it on the coffee table in the living room. (house-level)
\end{enumerate}

\textbf{Evaluator Accuracy.}
To validate the evaluator agent, we manually labeled all 300 evaluations (100 scenes $\times$ 3 states: initial condition, standard policy result, degraded policy result). The evaluator agreed with human judgment in 299/300 cases (99.7\%). The single disagreement was an ambiguous case: a fruit landed on the edge of a plate, which the human labeled as success but the evaluator judged as failure.

\textbf{Policy Comparison.}
Table~\ref{tab:robot_results} presents per-task success rates. The standard policy achieves 16\% overall success compared to 12\% for the degraded variant, confirming the two policies differ in performance as intended. This demonstrates that our automatic evaluation system can detect meaningful differences in policy quality.

\begin{table}[htbp]
\centering
\caption{Robot policy evaluation results across 100 generated scenes (25 per task).}
\label{tab:robot_results}
\small
\begin{tabular}{lccc}
\toprule
\textbf{Task} & \textbf{Standard} & \textbf{Degraded} & \textbf{$\Delta$} \\
\midrule
Fruit $\rightarrow$ plate & 3/25 (12\%) & 3/25 (12\%) & 0\% \\
Coke can $\rightarrow$ table & 6/25 (24\%) & 4/25 (16\%) & +8\% \\
Cup $\rightarrow$ sink & 4/25 (16\%) & 3/25 (12\%) & +4\% \\
Water bottle $\rightarrow$ coffee table & 3/25 (12\%) & 2/25 (8\%) & +4\% \\
\midrule
\textbf{Overall} & 16/100 (16\%) & 12/100 (12\%) & +4\% \\
\bottomrule
\end{tabular}
\end{table}

\section{Evaluation Methodology}
\label{app:eval_methodology}

This section provides details on the experimental setup and evaluation metrics.

\subsection{Experimental Setup}
\label{app:experimental_setup}

\subsubsection{Baselines}

We compare against five external baselines, all of which use LLMs or VLMs for scene generation but differ in their architectural approaches. Holodeck is the only baseline that supports house-level generation. LayoutVLM is evaluated with two different asset libraries. All baselines are evaluated using their open-source codebases.

\textbf{HSM}~\cite{pun2026hsm} is a hierarchical scene generation framework that recursively decomposes scene generation from room-level furniture to small objects on surfaces. It uses learned ``scene motifs'' (recurring spatial patterns like dining chairs around a table or stacked books) to capture arrangement structures. Assets are retrieved from the HSSD library~\cite{khanna2023hssd} using CLIP similarity, and collisions are resolved through spatial optimization.

\textbf{Holodeck}~\cite{yang2024holodeck} prompts an LLM to generate spatial relations between objects, then uses a constraint satisfaction solver to convert these into physically plausible layouts. It decouples semantic understanding from geometric optimization to avoid collisions and boundary violations. Holodeck is the only baseline that supports both room-level and house-level generation.

\textbf{I-Design}~\cite{IDesign_2024} is an LLM-based pipeline that uses a sequence of specialized prompts (designer, architect, engineer, corrector, refiner) to progressively build a scene graph from user preferences. Each role refines the scene description: the designer selects objects, the architect specifies spatial relationships, and correction stages resolve conflicts. The scene graph is converted to object poses using a backtracking algorithm that samples valid positions from constraint intersections.

\textbf{LayoutVLM}~\cite{sun2024layoutvlm} leverages vision-language models for layout generation by combining visual prompting with differentiable optimization. It generates both numerical pose estimates and spatial relations from visually marked images, filtering for self-consistency before jointly optimizing for semantic coherence and collision avoidance. Originally designed for placing a given set of objects, we implement the open-vocabulary extension from their Appendix B.1, which first retrieves objects from a library. We evaluate two asset libraries: \textbf{LayoutVLM (Curated)} uses the small curated library provided by the authors, while \textbf{LayoutVLM (Objaverse)} uses the Objathor subset of Objaverse~\cite{deitke2022objaverse, procthor}, a curated collection of assets used by ProcTHOR and Holodeck.

\textbf{SceneWeaver}~\cite{yang2025sceneweaver} is a single-agent LLM framework that synthesizes 3D scenes through iterative refinement using a ``reason-act-reflect'' paradigm. A separate VLM evaluates rendered scenes using physical and perceptual metrics. The agent then uses these evaluation results to select tools for progressive refinement. It uses a standardized tool interface with feedback-driven planning to maintain spatial consistency.

\subsubsection{Ablations}

We evaluate six ablations to understand component contributions:

\begin{itemize}[nosep]
    \item \textbf{NoCritic}: No critic iterations (initial design only)
    \item \textbf{NotGenerated}: HSSD \cite{khanna2023hssd} retrieved assets instead of generated assets
    \item \textbf{NoAssetValidation}: No VLM-based asset validation (routing components retained)
    \item \textbf{NoSpecializedTools}: No specialized placement tools (snap, facing, stack, fill, arrangement, pile)
    \item \textbf{NoObserveScene}: No visual observations (structured state only)
    \item \textbf{NoAgentMemory}: No memory of previous iterations for designer and critic
\end{itemize}

\subsubsection{Prompt Corpus}

We evaluate on 210 total prompts organized into five categories:
\begin{itemize}[nosep]
    \item \textbf{SceneEval-100 (Room)}: 94 room-level prompts from the SceneEval benchmark~\cite{tam2025sceneeval}, primarily bedrooms, living rooms, and dining rooms with varying complexity (the remaining 6 SceneEval prompts are house-level and included in the House-Level category)
    \item \textbf{Type Diversity}: 50 prompts each with a different room type not well-represented in SceneEval-100 (pet stores, pharmacies, yoga studios, wine cellars, etc.)
    \item \textbf{Object Density}: 25 prompts specifically designed to stress-test high object count scenarios (e.g., ``a kitchen shelf with at least 10 plates, 8 bowls, and 8 cups'')
    \item \textbf{Themed Scenes}: 10 prompts with specific stylistic constraints (Star Wars themed bedroom, steampunk styled home office, Harry Potter library, etc.)
    \item \textbf{House-Level}: 31 prompts for multi-room scenes (SceneEval-100 house subset plus additional house prompts)
\end{itemize}

Since most baselines only support room-level generation, we evaluate room-level prompts (179 total) and house-level prompts (31 total) separately. Room-level results comparing all methods are presented in Sections~\ref{app:human_evaluation} and~\ref{app:automated_evaluation}. House-level results comparing SceneSmith against Holodeck (the only baseline with multi-room support) are presented in Tables~\ref{tab:user_study} and~\ref{tab:sceneeval} in the main paper.

\subsection{SceneEval Metrics}
\label{app:sceneeval_metrics}

We use the following metrics from the SceneEval benchmark~\cite{tam2025sceneeval}:

\begin{itemize}[nosep]
    \item \textbf{CNT (Object Count)}: Satisfaction rate for object count requirements. A VLM maps scene objects to objects specified in the prompt, then compares instance counts to annotated quantities (exact or relative).
    \item \textbf{ATR (Object Attribute)}: Satisfaction rate for object attributes (e.g., colors, materials). Renders a front view and a reference view with a human figure for scale, then uses a VLM to evaluate if objects satisfy the specified attributes.
    \item \textbf{OOR (Object-Object Relationship)}: Satisfaction rate for spatial relationships between objects (e.g., ``chair in front of desk''). Evaluates predefined relationship types.
    \item \textbf{OAR (Object-Architecture Relationship)}: Satisfaction rate for object-architecture relationships (e.g., ``sofa against wall'', ``rug in middle of room''). Checks relationship types between objects and architectural elements.
    \item \textbf{ACC (Accessibility)}: Measures whether functional sides of objects are accessible. A VLM identifies functional sides (e.g., front of sofa, sides of bed), then a 2D occupancy analysis checks if those sides are unobstructed.
    \item \textbf{NAV (Navigability)}: Ratio of the largest connected free floor space to the total free space. Uses a 2D occupancy projection and connected component analysis to measure if object placements create isolated regions.
    \item \textbf{OOB (Out of Bounds)}: Fraction of objects outside the floor plan boundary. Samples surface points and casts rays toward the floor; objects with $<$99\% of rays hitting the floor are considered out-of-bounds.
\end{itemize}

\subsubsection{Physics Metrics for Simulation Readiness}

In addition to the semantic metrics above, we propose two physics-based metrics using Drake~\citepalias{drake} to evaluate whether generated scenes are ready for robotic simulation.

None of the baselines produce scenes that can be directly simulated in physics simulators, as they lack collision geometry and physical properties (mass, center of mass, inertia, friction). To enable evaluation, we first make baseline scenes simulator-compatible: we generate collision geometries using V-HACD convex decomposition (128 max hulls, matching our pipeline) and estimate mass, center of mass, and inertia assuming a uniform density of 1000 kg/m$^3$ (water density). We exclude carpets from these metrics as they are mainly visual elements, following SceneWeaver~\cite{yang2025sceneweaver}. This postprocessing step reflects the work required to use baseline outputs in robotic simulation. SceneSmith scenes are simulation-ready out of the box and require no such postprocessing.

\textbf{COL (Collision Rate):} Uses Drake's signed distance queries on collision geometry. Reports the fraction of objects in collision with penetration depth exceeding 1mm. This metric reflects whether scenes are physically feasible. Deep penetrations can cause objects to explode apart due to high contact forces when simulating a scene, while slight penetrations are generally tolerable.

\textbf{STB (Static Equilibrium):} Runs a 5-second physics simulation with gravity (1ms time step). Objects are marked stable if displacement $< 1$cm and rotation $< 0.1$rad after settling. Wall and ceiling objects are welded to the world frame since they are mounted fixtures. For fairness, support types (ground, wall, ceiling, or object-supported) are determined by SceneEval's VLM classification for all methods, including ours, rather than using ground-truth labels from our pipeline.

\section{Human Evaluation}
\label{app:human_evaluation}

This section presents the methodology of our human evaluation study. See Table~\ref{tab:user_study} in the main paper for complete pairwise comparison results. P-values were computed using two-sided binomial tests against a null hypothesis of 50\% win rate, with Benjamini-Hochberg FDR correction for 28 simultaneous tests (14 comparisons $\times$ 2 dimensions). Effect sizes are reported using Cohen's $h$ (small $\approx 0.2$, medium $\approx 0.5$, large $\approx 0.8$). Our study was powered to detect win rates $\geq$63\% with 95\% power; detecting smaller effects (53--55\%) would require 6-18x more comparisons.

\subsection{Study Setup}
\label{app:user_study_methodology}

We conducted a human evaluation study with the following setup:

\begin{itemize}[nosep]
    \item \textbf{Platform}: Prolific\footnote{\url{https://www.prolific.co/}}, USA-based participants only
    \item \textbf{Participants}: 210 recruited, 205 with complete data
    \item \textbf{Compensation}: \$12.50 per participant ($\sim$32 minutes mean completion time)
    \item \textbf{Scenes}: 179 room-level + 31 house-level prompts (same as in Section~\ref{app:eval_methodology})
    \item \textbf{Design}: Pairwise comparison where each comparison includes SceneSmith vs one baseline or ablation
    \item \textbf{Comparisons per participant}: 15 scene pairs
    \item \textbf{Total valid responses}: 3,051
\end{itemize}

\subsection{Interface Design}

The interface was iteratively refined through multiple rounds of feedback from 12 independent testers who completed the study and provided detailed suggestions. The final design prioritizes accessibility for non-specialists, with an interactive tutorial, contextual tooltips, and multiple viewing modes to accommodate different preferences and device capabilities. We optimized for various internet speeds by compressing GLB files for fast download, implementing lazy loading, and adding loading indicators for both 3D viewers and images. See Appendix~\ref{app:interface_screenshots} for screenshots.

Each comparison presents two scenes side-by-side with the following components:

\begin{itemize}[nosep]
    \item \textbf{Primary view}: Custom Babylon.js 3D viewer with full interactivity (orbit, pan, zoom)
    \item \textbf{Object inspection}: Click to select objects with a visual highlight
    \item \textbf{Controls}: Hide Selected, Show All, Frame (focus camera on object), Reset view, Fullscreen
    \item \textbf{Alternative views}: Click-Through carousel (9 camera angles) and Grid View for image-based viewing
    \item \textbf{Prompt display}: Text description shown prominently above the comparison
\end{itemize}

\noindent The evaluation was fully blinded: method names were hidden, and scene positions (left/right) were randomized.

\subsection{Questions}

Participants answered two questions per comparison. Each question included a short hint visible below the question text and an expandable tooltip with detailed guidance. The welcome page also provided context for interpreting each question.

\paragraph{Q1: Realism} (forced choice: Scene A / Scene B)

\begin{quote}
``Which scene looks more realistic?''
\end{quote}

\noindent\textit{Hint:} ``Consider: realism, object placement, layout coherence''

\noindent\textit{Tooltip:} ``Consider the realism of each scene: Does it look like a realistic room? Are the objects and furniture well-placed and appropriately sized? Does the overall layout feel coherent and plausible?''

\noindent\textit{Welcome page:} ``Which scene looks more realistic? Consider whether the room looks realistic, objects are well-placed and appropriately sized, and the layout feels coherent and plausible.''

\paragraph{Q2: Prompt Faithfulness} (three options: Scene A / Scene B / Equal)

\begin{quote}
``Which scene better follows the requirements in the text description?''
\end{quote}

\noindent\textit{Hint:} ``Consider: requested objects, room type, described details''

\noindent\textit{Tooltip:} ``Consider how well each scene matches the specific requirements mentioned in the prompt: Does the scene include the requested objects? Is the room type correct? Are details like colors, quantities, and arrangements as described? Additional objects that fit the room are fine.''

\noindent\textit{Welcome page:} ``Which scene better follows the requirements in the text description? Consider whether the requested objects are present, the room type is correct, and details like colors, quantities, and arrangements match what was described. Additional objects that fit the room are fine.''

\noindent\textit{Welcome page example:} ``If the prompt says `A living room with a sofa and a coffee table with a book on it', the scene should include a sofa and a coffee table with a book on top. Additional items typical of a living room (like a TV, rug, or lamp) are fine. We don't expect an empty room with just the mentioned objects.''

\noindent\textit{Equal confirmation:} When selecting ``Equal,'' participants saw: ``Are you sure both scenes follow the description equally well? Select `Equal' only if you genuinely cannot distinguish which scene better matches the prompt.''

\subsection{Randomization and Counterbalancing}

\begin{itemize}[nosep]
    \item \textbf{Left/right position}: Randomized per comparison to prevent positional bias
    \item \textbf{Baseline assignment}: Each baseline is evaluated by an equal number of participants
    \item \textbf{Comparison order}: The 15 comparisons are presented in random order per participant
    \item \textbf{Question order}: Randomized per comparison to prevent order effects between Q1 and Q2
    \item \textbf{No prompt repeats}: Each participant sees each prompt at most once
\end{itemize}

\subsection{Results by Prompt Category}

Table~\ref{tab:user_study_category} shows win rates broken down by prompt category for room-level scenes. Object Density prompts show the highest realism win rate (94.7\%), validating our method's ability to handle complex, object-rich scenes. Themed Scenes show the highest faithfulness win rate (94.9\%), indicating strong adherence to specific stylistic constraints. All room-level categories exceed 89\% on both metrics, demonstrating consistent performance across diverse scene types. House-level results are reported in Table~\ref{tab:user_study} in the main paper.

\begin{table}[htbp]
\centering
\caption{User study win rates by prompt category (room-level scenes, average across all external baselines). $n$ denotes the number of pairwise comparisons (excluding ties for faithfulness).}
\label{tab:user_study_category}
\small
\begin{tabular}{lcccc}
\toprule
\textbf{Category} & \textbf{Realism Win\%} & \textbf{Faith. Win\%} & \textbf{$n$ (Real.)} & \textbf{$n$ (Faith.)} \\
\midrule
SceneEval-100 & 92.2\% & 92.0\% & 831 & 808 \\
Type Diversity & 90.3\% & 89.8\% & 309 & 304 \\
Object Density & 94.7\% & 89.7\% & 151 & 146 \\
Themed Scenes & 89.8\% & 94.9\% & 59 & 59 \\
\bottomrule
\end{tabular}
\end{table}

\section{Automated Evaluation}
\label{app:automated_evaluation}

This section presents results from automated evaluation metrics including SceneEval benchmark scores and generation costs.

\subsection{SceneEval Benchmark Results}
\label{app:sceneeval_results}

Table~\ref{tab:sceneeval_full} presents SceneEval metrics and our physics metrics for all methods on the 179 room-level prompts.

\textbf{Error bars.} All values are reported as mean $\pm$ 95\% confidence interval (CI), computed using the t-distribution: $\text{CI} = t_{0.975, n-1} \times (s / \sqrt{n})$, where $s$ is the sample standard deviation and $n$ is the number of scenes. Non-overlapping CIs suggest statistically significant differences between methods.

\begin{table}[htbp]
\centering
\caption{SceneEval benchmark results (179 room-level scenes). Metrics: CNT = object count, ATR = attributes, OOR = object-object relationships, OAR = object-architecture relationships, ACC = accessibility, NAV = navigability, COL = collision rate, STB = stability, OOB = out-of-bounds. Values are mean $\pm$ 95\% t-distribution CI. $\uparrow$ = higher is better, $\downarrow$ = lower is better. Non-overlapping CIs suggest statistically significant differences.}
\label{tab:sceneeval_full}
\resizebox{\textwidth}{!}{
\begin{tabular}{l c ccccccccc}
\toprule
\textbf{Method} & \textbf{\#Obj} & \textbf{CNT}$\uparrow$ & \textbf{ATR}$\uparrow$ & \textbf{OOR}$\uparrow$ & \textbf{OAR}$\uparrow$ & \textbf{ACC}$\uparrow$ & \textbf{NAV}$\uparrow$ & \textbf{COL}$\downarrow$ & \textbf{STB}$\uparrow$ & \textbf{OOB}$\downarrow$ \\
\midrule
\multicolumn{11}{l}{\textit{External Baselines}} \\
HSM & 22.7$\pm$2.6 & 60.6$\pm$4.2 & 61.5$\pm$7.3 & 30.6$\pm$5.7 & 66.2$\pm$6.6 & 88.9$\pm$1.7 & 99.4$\pm$0.4 & 20.6$\pm$3.0 & 45.2$\pm$3.9 & 5.5$\pm$1.1 \\
Holodeck & 23.0$\pm$1.4 & 44.2$\pm$4.3 & 44.4$\pm$7.2 & 16.8$\pm$4.6 & 38.0$\pm$7.1 & 84.9$\pm$1.4 & 99.6$\pm$0.2 & 12.3$\pm$2.3 & 31.9$\pm$2.7 & 0.8$\pm$0.4 \\
IDesign & 13.0$\pm$1.0 & 69.3$\pm$4.0 & 50.3$\pm$7.2 & 28.6$\pm$5.5 & 66.6$\pm$6.7 & 70.1$\pm$2.8 & 95.9$\pm$1.4 & 3.0$\pm$1.5 & 60.8$\pm$4.6 & 4.3$\pm$2.0 \\
LayoutVLM (Curated) & 11.2$\pm$1.6 & 41.0$\pm$4.0 & 25.6$\pm$6.6 & 14.1$\pm$4.0 & 22.4$\pm$6.1 & 93.8$\pm$1.9 & 99.7$\pm$0.2 & 25.9$\pm$4.4 & 19.4$\pm$3.6 & 6.2$\pm$2.1 \\
LayoutVLM (Objaverse) & 14.2$\pm$1.7 & 55.5$\pm$4.7 & 34.3$\pm$7.0 & 20.7$\pm$5.0 & 17.4$\pm$5.6 & 91.5$\pm$1.6 & 98.7$\pm$0.8 & 28.9$\pm$3.5 & 8.1$\pm$1.7 & 5.1$\pm$1.4 \\
SceneWeaver & 13.5$\pm$1.0 & 41.8$\pm$3.9 & 29.1$\pm$6.7 & 15.3$\pm$4.2 & 31.6$\pm$6.5 & 77.5$\pm$2.7 & 98.1$\pm$1.2 & 12.5$\pm$2.5 & 37.3$\pm$4.4 & 0.0$\pm$0.0 \\
\midrule
\multicolumn{11}{l}{\textit{Ablations}} \\
NoCritic & 54.0$\pm$8.3 & 83.5$\pm$3.0 & 72.4$\pm$6.3 & 64.5$\pm$5.8 & 73.5$\pm$6.2 & 85.8$\pm$1.6 & 99.0$\pm$0.8 & 0.8$\pm$0.4 & 96.1$\pm$1.5 & 0.1$\pm$0.1 \\
NoObserveScene & 69.7$\pm$15.4 & 81.1$\pm$3.0 & 75.9$\pm$6.0 & 59.6$\pm$5.8 & 71.7$\pm$6.7 & 84.7$\pm$1.7 & 98.8$\pm$0.8 & 0.9$\pm$0.4 & 95.3$\pm$1.5 & 0.0$\pm$0.0 \\
NoAgentMemory & 78.9$\pm$19.7 & 82.5$\pm$3.1 & 72.6$\pm$6.5 & 66.0$\pm$5.8 & 81.5$\pm$5.3 & 84.5$\pm$1.6 & 97.6$\pm$1.2 & 1.7$\pm$0.7 & 94.1$\pm$1.9 & 0.1$\pm$0.1 \\
NoSpecializedTools & 61.5$\pm$14.1 & 83.1$\pm$2.9 & 75.2$\pm$6.1 & 66.0$\pm$5.5 & 81.2$\pm$5.4 & 83.8$\pm$1.5 & 97.7$\pm$1.1 & 0.5$\pm$0.3 & 97.6$\pm$1.2 & 0.0$\pm$0.0 \\
NoAssetValidation & 72.7$\pm$16.3 & 81.6$\pm$3.0 & 72.4$\pm$5.9 & 65.5$\pm$5.5 & 79.5$\pm$5.4 & 83.6$\pm$1.4 & 97.5$\pm$1.2 & 0.8$\pm$0.4 & 95.6$\pm$1.5 & 0.1$\pm$0.1 \\
NotGenerated & 57.7$\pm$9.9 & 75.7$\pm$3.1 & 64.2$\pm$6.9 & 53.7$\pm$6.2 & 72.4$\pm$6.3 & 83.9$\pm$1.6 & 97.3$\pm$1.3 & 7.4$\pm$1.3 & 88.4$\pm$2.4 & 0.1$\pm$0.1 \\
\midrule
SceneSmith (Ours) & 71.1$\pm$13.0 & 82.9$\pm$3.4 & 74.4$\pm$6.2 & 67.6$\pm$5.8 & 80.6$\pm$5.5 & 83.4$\pm$1.5 & 97.6$\pm$1.1 & 1.2$\pm$0.6 & 95.6$\pm$1.7 & 0.2$\pm$0.4 \\
\bottomrule
\end{tabular}}
\end{table}

See Table~\ref{tab:sceneeval} in the main paper for baseline comparison results. For ablations, most semantic metrics have overlapping CIs with SceneSmith, indicating these automated metrics lack sensitivity to differentiate system components. The exception is NotGenerated, which shows statistically significant degradation on CNT (75.7\% vs 82.9\%), OOR (53.7\% vs 67.6\%), and STB (88.4\% vs 95.6\%), all with non-overlapping CIs. This suggests that generated assets provide meaningful improvements over a fixed asset library. NoObserveScene and NoAssetValidation show significant user study differences (Table~\ref{tab:user_study}) not captured by these automated metrics. Other ablations (NoCritic, NoAgentMemory, NoSpecializedTools) show user study win rates near 50\% (52--55\% realism) that did not reach significance with our study power, and similarly indistinguishable automated metrics.

The NotGenerated ablation also controls for asset-library effects. It preserves the SceneSmith construction pipeline while replacing generated assets with retrieved HSSD assets, the same asset source used by HSM. Its improvement over HSM indicates that SceneSmith's gains are not solely due to generated asset appearance or vocabulary, while its gap to full SceneSmith shows that generated assets still contribute.

\subsection{Base Model Swap Control}
\label{app:model_swap}

For the main comparisons, we evaluate each baseline in its released configuration, including its original models, prompts, and settings. This preserves each method as designed, since prompts and tool calls are part of the system and are often tuned to the model used by the authors. As a base-model control, we also rerun HSM and Holodeck, our strongest baselines, with GPT-5.2 at high reasoning effort on the 179 room-level prompts, while keeping their prompts and tools fixed. Table~\ref{tab:model_swap_control} shows that the model swap changes both baselines, but not consistently for the better: HSM-GPT-5.2 increases \#Obj, CNT, OOR, and OAR, but lowers ATR, ACC, NAV, and STB and worsens COL and OOB; Holodeck-GPT-5.2 increases \#Obj and slightly improves CNT, ATR, NAV, COL, and OOB, but degrades OOR, OAR, ACC, and STB. SceneSmith remains ahead on most metrics, with especially large gaps on CNT, ATR, OOR, OAR, COL, and STB, indicating that its gains are not explained by the base model alone.

\begin{table}[htbp]
\centering
\caption{Base model swap control for the strongest baselines (179 room-level scenes). Values are mean $\pm$ 95\% t-distribution CI. $\uparrow$ = higher is better, $\downarrow$ = lower is better. Non-overlapping CIs suggest statistically significant differences. Each baseline is grouped with its GPT-5.2 rerun. In GPT-5.2 rows, green/red entries indicate improved/degraded mean estimates relative to the corresponding original baseline, accounting for metric direction; \#Obj is left uncolored.}
\label{tab:model_swap_control}
\small
\resizebox{\textwidth}{!}{
\begin{tabular}{lcccccccccc}
\toprule
\textbf{Method} & \textbf{\#Obj} & \textbf{CNT}$\uparrow$ & \textbf{ATR}$\uparrow$ & \textbf{OOR}$\uparrow$ & \textbf{OAR}$\uparrow$ & \textbf{ACC}$\uparrow$ & \textbf{NAV}$\uparrow$ & \textbf{COL}$\downarrow$ & \textbf{STB}$\uparrow$ & \textbf{OOB}$\downarrow$ \\
\midrule
HSM & 22.7$\pm$2.6 & 60.6$\pm$4.2 & 61.5$\pm$7.3 & 30.6$\pm$5.7 & 66.2$\pm$6.6 & 88.9$\pm$1.7 & 99.4$\pm$0.4 & 20.6$\pm$3.0 & 45.2$\pm$3.9 & 5.5$\pm$1.1 \\
HSM-GPT-5.2 & 31.2$\pm$2.9 & \textcolor{green!45!black}{63.8$\pm$4.1} & \textcolor{red!70!black}{57.8$\pm$7.3} & \textcolor{green!45!black}{37.1$\pm$6.5} & \textcolor{green!45!black}{66.3$\pm$6.7} & \textcolor{red!70!black}{83.3$\pm$2.1} & \textcolor{red!70!black}{99.1$\pm$0.6} & \textcolor{red!70!black}{31.8$\pm$3.5} & \textcolor{red!70!black}{36.7$\pm$4.0} & \textcolor{red!70!black}{7.2$\pm$1.1} \\
\midrule
Holodeck & 23.0$\pm$1.4 & 44.2$\pm$4.3 & 44.4$\pm$7.2 & 16.8$\pm$4.6 & 38.0$\pm$7.1 & 84.9$\pm$1.4 & 99.6$\pm$0.2 & 12.3$\pm$2.3 & 31.9$\pm$2.7 & 0.8$\pm$0.4 \\
Holodeck-GPT-5.2 & 27.6$\pm$1.5 & \textcolor{green!45!black}{45.2$\pm$4.1} & \textcolor{green!45!black}{44.9$\pm$7.3} & \textcolor{red!70!black}{15.6$\pm$4.3} & \textcolor{red!70!black}{33.6$\pm$6.5} & \textcolor{red!70!black}{80.4$\pm$1.4} & \textcolor{green!45!black}{99.7$\pm$0.1} & \textcolor{green!45!black}{10.8$\pm$1.9} & \textcolor{red!70!black}{27.4$\pm$2.5} & \textcolor{green!45!black}{0.4$\pm$0.2} \\
\midrule
SceneSmith (Ours) & 71.1$\pm$13.0 & 82.9$\pm$3.4 & 74.4$\pm$6.2 & 67.6$\pm$5.8 & 80.6$\pm$5.5 & 83.4$\pm$1.5 & 97.6$\pm$1.1 & 1.2$\pm$0.6 & 95.6$\pm$1.7 & 0.2$\pm$0.4 \\
\bottomrule
\end{tabular}}
\end{table}

\subsection{Equilibrium Statistics}

While the STB metric reports the \textit{fraction} of stable objects, it does not capture the \textit{severity} of instability. Table~\ref{tab:equilibrium_stats} presents detailed equilibrium statistics from our 5-second physics simulation, showing how much objects displace and rotate after settling.

\begin{table}[htbp]
\centering
\caption{Equilibrium statistics for room scenes (179 prompts). Values are mean $\pm$ 95\% t-distribution CI, averaged across scenes. MD = mean displacement across non-welded objects per scene; XD = max displacement per scene; MR = mean rotation per scene. Lower values indicate objects settle with minimal movement, critical for simulation-ready scenes. Non-overlapping CIs suggest statistically significant differences. CIs are symmetric approximations; actual values are bounded at zero.}
\label{tab:equilibrium_stats}
\small
\begin{tabular}{lccc}
\toprule
\textbf{Method} & \textbf{MD (mm)} & \textbf{XD (m)} & \textbf{MR (rad)} \\
\midrule
\multicolumn{4}{l}{\textit{SceneSmith + Ablations}} \\
SceneSmith (Ours) & 12.3$\pm$4.5 & 0.35$\pm$0.13 & 0.014$\pm$0.004 \\
NoCritic & 28.5$\pm$34.0 & 1.31$\pm$1.40 & 0.011$\pm$0.005 \\
NoObserveScene & 61.9$\pm$83.7 & 1.14$\pm$1.31 & 0.020$\pm$0.011 \\
NoAgentMemory & 795.1$\pm$1078.8 & 1.92$\pm$1.94 & 0.020$\pm$0.006 \\
NoSpecializedTools & 22.6$\pm$24.4 & 0.67$\pm$0.83 & 0.007$\pm$0.003 \\
NotGenerated & 45.3$\pm$19.4 & 0.94$\pm$0.59 & 0.049$\pm$0.015 \\
NoAssetValidation & 33.3$\pm$31.5 & 1.19$\pm$1.31 & 0.011$\pm$0.003 \\
\midrule
\multicolumn{4}{l}{\textit{Baselines}} \\
HSM & 654.8$\pm$164.1 & 5.02$\pm$1.88 & 0.193$\pm$0.037 \\
Holodeck & 932.2$\pm$233.6 & 7.95$\pm$2.69 & 0.217$\pm$0.032 \\
IDesign & 382.3$\pm$106.1 & 2.06$\pm$0.50 & 0.114$\pm$0.027 \\
LayoutVLM (Curated) & 684.5$\pm$117.1 & 2.49$\pm$0.46 & 0.208$\pm$0.040 \\
LayoutVLM (Objaverse) & 906.3$\pm$114.7 & 3.51$\pm$0.44 & 0.299$\pm$0.049 \\
SceneWeaver & 575.8$\pm$103.2 & 3.30$\pm$0.55 & 0.193$\pm$0.033 \\
\bottomrule
\end{tabular}
\end{table}

SceneSmith achieves 12.3 $\pm$ 4.5mm mean displacement with a tight CI, indicating consistent stability across scenes: objects reliably settle within $\sim$1cm. Baselines show 30-75x higher mean displacement (382--932mm) with non-overlapping CIs, confirming statistically significant differences. The contrast in variance is also informative: ablations like NoAgentMemory (795 $\pm$ 1079mm) have CIs larger than their means, indicating inconsistent behavior where some scenes work well while others contain objects that fail significantly. Note that some instability stems from VLM misclassification of object support types (e.g., wall-mounted objects incorrectly classified as floor-supported, which then fall). This is expected to affect all methods equally.

\subsection{Collision Depth Statistics}
\label{app:collision_depth}

While the COL metric reports the \textit{fraction} of colliding objects, it does not capture \textit{how severely} objects interpenetrate. Table~\ref{tab:collision_depth_stats} presents penetration depth statistics, showing the mean depth among colliding objects per scene.

\begin{table}[htbp]
\centering
\caption{Collision depth statistics for room scenes (179 prompts). Values are mean $\pm$ 95\% t-distribution CI, averaged across scenes. COL = fraction of objects in collision ($>$1mm penetration); MPD = mean penetration depth among colliding objects per scene. Lower values indicate less severe interpenetration. Non-overlapping CIs suggest statistically significant differences. CIs are symmetric approximations; actual values are bounded at zero.}
\label{tab:collision_depth_stats}
\small
\begin{tabular}{lcc}
\toprule
\textbf{Method} & \textbf{COL (\%)}$\downarrow$ & \textbf{MPD (mm)}$\downarrow$ \\
\midrule
\multicolumn{3}{l}{\textit{SceneSmith + Ablations}} \\
SceneSmith (Ours) & 1.2$\pm$0.6 & 3.77$\pm$1.28 \\
NoCritic & 0.8$\pm$0.4 & 3.03$\pm$1.44 \\
NoObserveScene & 0.9$\pm$0.4 & 4.44$\pm$3.67 \\
NoAgentMemory & 1.7$\pm$0.7 & 3.99$\pm$1.27 \\
NoSpecializedTools & 0.5$\pm$0.3 & 4.48$\pm$3.13 \\
NotGenerated & 7.4$\pm$1.3 & 6.14$\pm$1.91 \\
NoAssetValidation & 0.8$\pm$0.4 & 2.32$\pm$0.70 \\
\midrule
\multicolumn{3}{l}{\textit{Baselines}} \\
HSM & 20.6$\pm$3.0 & 14.54$\pm$3.46 \\
Holodeck & 12.3$\pm$2.3 & 12.94$\pm$3.88 \\
IDesign & 3.0$\pm$1.5 & 47.05$\pm$25.31 \\
LayoutVLM (Curated) & 26.1$\pm$4.4 & 15.00$\pm$4.01 \\
LayoutVLM (Objaverse) & 31.4$\pm$3.5 & 12.78$\pm$1.78 \\
SceneWeaver & 12.6$\pm$2.6 & 10.87$\pm$3.21 \\
\bottomrule
\end{tabular}
\end{table}

SceneSmith's collisions are not only rare (1.2\%) but shallow: colliding objects penetrate by only 3.77 $\pm$ 1.28mm on average, consistent with slight overlaps among densely packed manipulands (e.g., fruit in bowls) that do not cause simulator instability, as contact forces are proportional to penetration depth. Baselines exhibit 3-12x deeper penetrations (10.87--47.05mm), with IDesign showing particularly severe interpenetration (47.05 $\pm$ 25.31mm) despite a relatively low collision rate (3.0\%), indicating that its few collisions involve deeply overlapping objects. The NotGenerated ablation stands out among ablations with both the highest collision rate (7.4\%) and deepest penetrations (6.14mm), suggesting that retrieved HSSD assets have visual geometry that produces worse V-HACD collision decompositions, leading to more penetrations during physics post-processing.

\subsection{Generation Cost and Time}
\label{app:generation_costs}

Table~\ref{tab:cost_analysis} reports per-scene generation cost and latency over the same 179 room-level scenes used for the automated metrics. The time statistics are measured in our throughput-oriented evaluation setting, where 25 scenes are generated concurrently on 8 L40S GPUs; they should therefore not be read as optimized single-scene latency.

\begin{table}[htbp]
\centering
\caption{Generation cost and time statistics per scene over the 179 room-level scenes. Time is measured with 25 scenes generated concurrently on 8 L40S GPUs.}
\label{tab:cost_analysis}
\small
\begin{tabular}{lccccc}
\toprule
\textbf{Method} & \textbf{Mean} & \textbf{Std} & \textbf{Min} & \textbf{Max} \\
\midrule
\multicolumn{5}{l}{\textit{Cost per scene (\$)}} \\
SceneSmith (Ours) & 13.98 & 9.98 & 4.59 & 101.93 \\
NoCritic & 4.24 & 2.15 & 1.73 & 22.70 \\
NoObserveScene & 9.62 & 7.46 & 3.94 & 75.56 \\
NoSpecializedTools & 14.31 & 9.48 & 5.07 & 72.54 \\
NoAssetValidation & 14.85 & 9.24 & 5.52 & 81.99 \\
NoAgentMemory & 16.41 & 10.76 & 5.56 & 96.62 \\
NotGenerated & 15.70 & 9.99 & 4.42 & 104.79 \\
\midrule
\multicolumn{5}{l}{\textit{Time per scene (hours:minutes)}} \\
SceneSmith (Ours) & 3:26 & 1:57 & 1:17 & 15:45 \\
NoCritic & 1:40 & 1:04 & 0:42 & 12:17 \\
NoObserveScene & 2:10 & 0:49 & 0:48 & 5:24 \\
NoSpecializedTools & 3:14 & 1:41 & 1:20 & 17:43 \\
NoAssetValidation & 3:11 & 1:52 & 1:19 & 20:22 \\
NoAgentMemory & 4:14 & 2:01 & 1:48 & 15:35 \\
NotGenerated & 3:33 & 1:49 & 1:11 & 13:55 \\
\bottomrule
\end{tabular}
\end{table}

NoCritic achieves 70\% cost reduction (\$4.24 vs \$13.98) and 51\% faster generation (1:40 vs 3:26) but produces sparser scenes (54.0 vs 71.1 objects on average), presenting a cost-density trade-off discussed in the main paper. NoObserveScene is 31\% cheaper (\$9.62 vs \$13.98) and 37\% faster (2:10 vs 3:26) because images consume many tokens. Conversely, NoAgentMemory is 17\% more expensive (\$16.41 vs \$13.98) and 23\% slower (4:14 vs 3:26) because designers and critics retry previously failed approaches without remembering what was attempted.

Table~\ref{tab:cost_breakdown} breaks down SceneSmith cost by agent component over these 179 scenes. Designer calls account for 75.9\% of cost, making the refinement loop the main target for further cost reduction. Replacing some designer or critic calls with smaller or open-weight models is a promising direction for decreasing generation cost and time.

\begin{table}[htbp]
\centering
\caption{Cost breakdown by agent component for SceneSmith over the 179 room-level scenes. Designer includes Initial, Change, and Layout sub-agents. Orchestrator includes Manipuland planner. Other includes VLM calls for asset physics analysis and furniture selection.}
\label{tab:cost_breakdown}
\small
\begin{tabular}{lcc}
\toprule
\textbf{Component} & \textbf{Cost (\$)} & \textbf{\% of Total} \\
\midrule
Designer & 10.61 & 75.9\% \\
Critic & 2.79 & 20.0\% \\
Orchestrator & 0.24 & 1.7\% \\
Other & 0.34 & 2.4\% \\
\midrule
\textbf{Total} & \textbf{13.98} & \textbf{100\%} \\
\bottomrule
\end{tabular}
\end{table}

\section{Evaluation Prompts}
\label{app:prompts}

This section lists all 210 evaluation prompts organized by category.

\subsection{SceneEval-100 (Room)~\cite{tam2025sceneeval} (94 prompts)}

\begin{enumerate}[nosep, leftmargin=*]
\footnotesize
    \item A bedroom with a bed, two nightstands, and a wardrobe in the corner of the room.
    \item A bedroom with a bed and a wardrobe.
    \item A bedroom with a king-size bed in the corner of the room, two large blue armchairs, and a floor lamp near a armchair.
    \item A bedroom with a double bed and a mini fridge near the bed, a table across from the door, and a painting on the wall above the bed.
    \item A bedroom with two bunk beds in the corners of the room, a desk placed against the wall far from the door, and a light hanging from the ceiling in the center of the room.
    \item A simple bedroom featuring a twin bed against the wall, with a wardrobe positioned in the corner of the room, and a desk next to the window.
    \item A minimal bedroom with a queen bed against the wall, flanked by two nightstands on each side.
    \item A living room with a TV, sofa, and bookshelf. There is no coffee table in the room.
    \item A living room with a TV, sofa, bookshelf, and coffee table.
    \item A living room with a sofa in front of a wall-mounted TV, a lamp behind the sofa, and a stool near the TV against the wall.
    \item A living room with two bookcases along the wall far from the door, a wooden chair near the bookcases, and a painting on the wall opposite the bookcases.
    \item A living room with a two-seater sofa against the wall, a square rug in the middle in front of the sofa, and two large plants on the floor near the sofa.
    \item A living room with a bean bag near a wall cabinet, which is across from the window. There is a stool next to the bean bag.
    \item A living room with a sofa against the wall across from the door, a painting on the wall above the sofa, and a pendant light hanging in the middle of the room.
    \item A dining room with two bar stools at the short sides of a bar table.
    \item A dining room with three dining chairs surrounding a dining table.
    \item A small dining room with a dining table next to a window, a chair on the long side of the table, and a rug in the middle of the room.
    \item A dining room with a cabinet next to the door against the wall, and two wine cabinets against the wall near a bar table.
    \item A dining room with a table in the corner of the room and a chair on the long side.
    \item A dining room with a ceiling light hanging above the table and two chairs on the long side of the table.
    \item A bedroom with a bed with two nightstands on each side, a wardrobe, and a TV stand positioned in front of the bed. There is a toy box with two dolls outside the box.
    \item A bedroom with a bed flanked by two nightstands, and an office chair in front of an office desk located in the corner of the room. There is no wardrobe in the room.
    \item A bedroom with a bed against the wall and at least one nightstand next to it, featuring a lamp on the nightstand. There is an office desk with a monitor on top and a sofa chair in front of it.
    \item A teenager's bedroom featuring a twin bed with an adjacent nightstand, a wardrobe for clothes, a small desk with a chair, and a bookshelf positioned next to the wardrobe. There are two books outside the bookshelf.
    \item A teenager's bedroom designed for both comfort and functionality includes a twin bed with an adjacent wooden nightstand, a large wardrobe with multiple drawers for clothes, and a small desk and chair equipped with an electronic setup featuring a monitor and peripherals. A large oak bookshelf is positioned next to the wardrobe.
    \item A bedroom with a king-size bed positioned against the wall across from the window. Two stools are placed in front of the bed. A painting hangs above the bed, flanked by a wall light on each side. There is a bookcase with two books on the left side.
    \item A bedroom with three extendable beds, one of which has a doll placed on top. Two of the beds are positioned in the corners of the room, and a dresser is located next to the door. There are two dolls outside a toy box.
    \item A living room with a coffee table in front of a large sofa, a TV on a TV stand facing the sofa, and a chair on the short side facing the coffee table. There is a cup on the right side of the TV stand.
    \item A living room with a coffee table in front of a large sofa, a TV on a TV stand facing the sofa, and two small tables on each side of the sofa.
    \item A living room featuring two bookshelves along the wall, with only one book on the bookshelf nearest the door. Across from the bookshelves, a painting is mounted on the wall. In the center of the room, a table with a tray holding a potted plant sits near an L-shaped sofa.
    \item A living room with a marble table in front of a four-seater sofa. The sofa is positioned in the corner of the room, with two bean bags placed next to the table. A lamp hangs from the ceiling.
    \item A large living room with three display shelves against the wall, with a long sofa in front. There are two tables in front of the sofa and another table right of the sofa with a lamp on top. There is no TV in the room.
    \item A Japanese-style living room featuring a coffee table next to the window with two floor cushions placed beside it. A sofa is positioned across from the window, and in front of the sofa is a table with a teapot on top, completing the serene and minimalist setup.
    \item A living room featuring an irregular-shaped table in the middle of the room with a sofa positioned in front of it. Across the table are two sofa chairs with a small wooden coffee table placed between them. A clock is mounted on the wall far from the door.
    \item A dining room with a bar table positioned in the middle of the room. A wooden shelf is mounted on the wall, holding a potted plant and a jug, adding a touch of decor and functionality. Along the wall, there is a fridge and a wine cabinet, providing ample storage and a cohesive design.
    \item A dining room with a fridge positioned near the door against the wall. A table is placed in the middle of the room with two chairs on its long sides. In the corner of the room across from the door, there is a cabinet.
    \item A dining room with six wooden dining chairs surrounding a round wooden table in the middle of the room. There is no coffee table in the room.
    \item A dining room with a square dining table, featuring two chairs on each long side and one chair on each short side of the table.
    \item A dining room with a dining table featuring four chairs on the long side and one on a short side. A vase is placed in the middle of the table. A display cabinet is positioned next to a window against the wall.
    \item Generate a dining room. The room should have a rectangular, wooden dining table with three chairs surrounding it. There is a utility cart near the table with three plates inside.
    \item A spacious master bedroom featuring a king-size bed against the wall, flanked by two nightstands, one holding a modern table lamp and a small book. A large wardrobe is positioned in the corner of the room, and a wooden dresser with three picture frames on it. Near the window, a plush armchair with a throw blanket provides a cozy seating area. A round area rug is placed at the foot of the bed, adding warmth and style to the space.
    \item A teenager's bedroom features a comfortable twin bed with a backboard in the far corner, with boxes underneath it. At the foot of the bed is a small desk equipped with a monitor, an external keyboard and mouse, and a desk lamp on the right for visibility, accompanied by a rolling chair. Next to the bed, a nightstand with an additional floor lamp nearby provides space for a phone and other valuables. A sizable wooden wardrobe with multiple drawers offers ample storage for clothes, while a coffee table beside it holds books and board games. In the center of the room, a tan-colored rug creates a cozy spot to sit, and the walls are adorned with various posters and pictures.
    \item A bedroom featuring a bunk bed positioned next to the window, adorned with plush toys on the bed. The room includes two desks placed against the wall, each accompanied by a chair. Over ten plush toys are scattered across the desks, adding a playful and cozy touch to the space.
    \item A modern living room featuring an L-shaped sofa facing a wall-mounted TV above a floating media cabinet. Two leather accent chairs face a glass coffee table in the center, which displays two art books. A bar cart stocked with glasses and bottles stands in the corner, while three pendant lights hang gracefully from the ceiling. The sofa is flanked by two side tables, one of which holds a table lamp, and a large abstract painting decorates the wall.
    \item The room is dimly lit, creating a somber atmosphere in a cozy and well-furnished living room. In the center of the room, there is a dining table with four wooden dining chairs arranged neatly around it. The table appears to be set, indicating a recent meal or gathering. Adjacent to the dining area, There is a glass coffee table serving as a centerpiece for the seating arrangement. It is both stylish and practical, providing a surface for drinks or decor. A multi-seat sofa is in front of the coffee table, providing ample seating for guests. Completing the seating options, two comfortable armchairs also face the coffee table on opposite sides near the sofa. Next to the sofa and each armchair is a corner side table, adding a touch of functionality and convenience. Each table has a lamp on it for lighting. The sofa and seating arrangement should face an opposite wall, against which someone in the scene could view a large flatscreen TV. The TV is supported by a TV stand with storage underneath for various books. The room appears to reflect the inhabitants' taste for a warm and inviting environment, despite the unsettling news program playing on the TV in the background. The juxtaposition of the serene living room with the chaos and screams on the TV screen creates a sense of tension and unease.
    \item A modern living room featuring four sofa chairs surrounding a circular ceramic table, with two floor lamps positioned adjacent to the table. All the furniture is set on a large rectangular rug in the middle of the room. A large wooden tray with two glass bowls rests on top of the table. A mirror is mounted on the wall next to the window, flanked by a wall lamp on each side.
    \item An open-plan dining area featuring a round table surrounded by six mid-century modern chairs. A grand chandelier illuminates the table, while a sideboard with four drawers rests against the wall, topped with three potted plants. Two wall sconces provide additional lighting, flanking a large mirror that hangs between them. A wine cabinet is positioned in the corner, with a serving cart placed nearby for added convenience.
    \item A cozy bedroom featuring a queen-sized bed against a soft pastel wall, with a small desk and bookshelf tucked neatly into the corner.
    \item A simple bedroom featuring a full bed flanked by two bedside lamps, with a large wardrobe positioned directly facing the bed.
    \item A bright bedroom with a king-sized bed, a plush area rug under it, and a dresser along the wall.
    \item A large bedroom with a California king bed, a reading table in the corner with a comfy chair, and a wall-mounted TV for entertainment.
    \item A living room featuring an overstuffed sofa, a vintage wooden table, alongside a bicycle hung decoratively on the wall.
    \item A chic living room with a bright rug, a gallery wall of family photos, and a comfortable hammock chair.
    \item A contemporary living room with a leather sofa, a small coffee table, and a vintage typewriter.
    \item A sunny dining room with a large picnic-style table, a colorful tablecloth, and a lovely herb garden next to the window.
    \item A simple dining room featuring a farmhouse table, a rustic chandelier, and a small cart placed against the wall.
    \item A minimalist dining room with a glass table and an industrial-style metal chair next to it, accented by a unique wall-mounted wine rack.
    \item In a cheerful playroom, a spacious foam play mat lies in the middle, with a small climbing structure positioned nearby for active play, and a cozy bench situated against the wall.
    \item This playroom features a large colorful rug at its center, with a play kitchen set against one wall and a rocking horse placed on the rug.
    \item A compact gaming room is designed with an adjustable computer desk and chair for PC gaming, positioned next to a large TV and a comfortable couch.
    \item A minimalist gaming room features a wall-mounted screen paired with a low-profile media console below, while a compact gaming desk with an ergonomic chair is positioned nearby.
    \item A kitchen featuring a bistro table with two chairs in a corner and a refrigerator positioned nearby.
    \item A stylish kitchen with an elegant pendant light hanging over a marble-topped kitchen counter. A single cabinet is positioned against the wall, providing ample storage space.
    \item A bathroom with a freestanding bathtub positioned beside a frosted window and a double vanity located across from it.
    \item In a classic bathroom, a double-sink vanity is set against one wall with two round mirrors hanging above, while a toilet is positioned in the corner.
    \item In this basement, a pool table occupies the center of the room, with two bar stools lined up at a nearby bar table, allowing for social gatherings and entertainment.
    \item A gym-purpose basement featuring a treadmill and an exercise bike positioned against one wall, with a tall mirror mounted on a wall directly facing the exercise bike.
    \item This is a small, plain bedroom. Upon entering through the door, a white desk with a black rolling chair in front of it is positioned against the wall to the right. A queen bed is centered against a wall, with two modern nightstands beside it, one of which holds a stylish lamp.
    \item A large bedroom with a lofted twin bed in one corner, freeing up the floor space below for a comfy armchair and a small side table. A bookcase stands against the wall next to the armchair, while a small desk, equipped with a pen and some papers along with a lamp for studying, is positioned near the window.
    \item A bedroom with a modern queen bed against the main wall. Two bedside tables stand on either side, while a small writing desk is positioned in the corner next to the window, equipped with a cute desk lamp. A wicker basket lies at the foot of the bed.
    \item A bedroom featuring a queen-sized bed against one wall, with a whimsical painting hung above it. A small bedside table sits to the left of the bed, holding a cute alarm clock, while a dresser stands across from the door.
    \item A living room with a sectional sofa, a wooden coffee table in the middle, and a retro record player in the corner with some records nearby.
    \item A living room with a large wall-mounted screen, several chairs arranged in front of it, and a collection of board games on a shelf in the corner.
    \item A cozy living room with an L-shaped oversized sofa against one wall, adorned with three colorful pillows. In front of the sofa, a wooden coffee table holds a small vase with fresh flowers.
    \item A dining room featuring a simple rectangular table, a corner shelf filled with cookbooks, and a pendant light hanging from the ceiling.
    \item A simple dining room with a picnic table, illuminated by string lights overhead, and a small cooler holding utensils and napkins.
    \item A dining room with a round wooden table in the middle of the room, surrounded by four wooden chairs, and a large vintage map displayed on the wall.
    \item A bright playroom featuring a low table with board games in the center, surrounded by three small chairs for group activities, while an adjacent puppet theater is positioned against the wall.
    \item This engaging playroom features a small painting stand positioned next to the window, a bookshelf full of toys standing against the nearby wall, and a cozy reading chair set across from the door.
    \item A tech-styled gaming room featuring a glass gaming desk with dual monitors and a powerful gaming PC, with an ergonomic chair positioned directly in front.
    \item A welcoming gaming room featuring a U-shaped sofa with a coffee table in front, facing a large TV on the wall. In the corner, there is a mini arcade machine and a PlayStation.
    \item In this modern kitchen, a large bar table with four high-backed stools stands in the center, facilitating social gatherings, while a stainless steel fridge is positioned against the far wall, next to a sleek pantry shelf. A small bowl of fruit sits on the table's surface.
    \item This cozy kitchen features a rustic wooden dining table positioned against the window, surrounded by four chairs. A sideboard against the wall is equipped with a coffee maker and a small herb planter.
    \item In this bathroom, a wide vanity sits against the wall, paired with a tall mirror mounted above. Nearby, a stylish laundry hamper is positioned next to the door, while a small basket containing bath essentials rests beside the tub.
    \item A bright bathroom features a spacious bathtub positioned beside a large window, while a stylish shelving unit stands against the wall, displaying towels and small decorative items. The toilet is placed across from the tub, with a small basket beside it.
    \item This versatile basement features a large sofa and an area rug in the center, creating a cozy movie area, while a small treadmill is positioned in the corner for quick exercise sessions. A mini chalkboard hangs on the wall next to a wall clock.
    \item This entertaining basement layout features a large gaming setup with two monitors on a desk against one wall, while a comfy bean bag chair is positioned nearby for casual seating. Across from the gaming area, a small cabinet with a mini fridge and a popcorn machine on top completes the setup.
    \item A welcoming bedroom with a full bed placed against the wall, flanked by two nightstands topped with a bedside clock and a decorative candle. A reading chair sits in the corner next to a tall bookshelf filled with novels.
    \item A bedroom featuring a vintage double bed dressed in soft linens, with a painting hanging on the wall above it. On either side of the bed are wooden nightstands, each with a small decorative vase. A tall chest of drawers stands against the opposite wall. A small round table in the corner holds a lamp, colorful cobblestones, and photo frames.
    \item A living room featuring a comfortable recliner chair placed next to a tall bookshelf filled with books and small sculptures. Across from the chair, a table next to a large window holds a collection of candles. In the center of the room, a round coffee table is surrounded by a simple sofa and two cube ottomans.
    \item A living room with a comfortable loveseat against the wall, adorned with plush cushions. Next to the loveseat, a small table holds a cup of tea and a remote control. A mid-century style coffee table sits in front of the loveseat with art magazines spread out. A large bookshelf containing books and newspapers lines the wall across from the door.
    \item A dining room with a circular table surrounded by six vintage chairs, and an old wooden ladder against the wall displaying plants and decorative jars.
    \item In this playroom, a large toy chest sits against the wall, overflowing with colorful building blocks, while a small craft table nearby is equipped with papers, crayons, and glue sticks for hands-on projects.
    \item This gaming room features powerful speakers positioned in every corner to create an immersive audio experience, with a large screen mounted opposite a plush sofa. A small table beside the sofa holds various game controllers.
\end{enumerate}

\subsection{Type Diversity (50 prompts)}

\begin{enumerate}[nosep, leftmargin=*]
\footnotesize
    \item A pet store with aquariums along one wall, exactly two shelf units where one displays pet food bags and the other displays pet toys, pet beds on the floor, and a checkout counter near the entrance.
    \item A bookstore with at least four tall bookshelves, each containing at least five books. A central display table showcases featured titles, and two comfortable reading chairs sit in one corner.
    \item A small grocery store with two refrigerated display cases along one wall. Two tables near the entrance hold produce bins, each containing at least three fruits. At least four shopping baskets are stacked by the door.
    \item A pharmacy with a pharmacy counter in the back of the room. Two shelves stand against the walls, each holding at least four medicine bottles. Three waiting chairs are arranged near the counter.
    \item A toy store with two large toy bins in the middle of the room, each containing at least five stuffed animals. Three shelves along the walls display multiple board games each, and a rocking horse sits in one corner.
    \item An electronics store with two display tables in the center, one showcasing three laptops and the other holding four tablets. Two large televisions are mounted on the wall, and a headphone display stand holds headphones.
    \item A furniture showroom with a sofa and coffee table arrangement in one corner, a dining table surrounded by four chairs in the center, and two floor lamps positioned near the walls. A large rug lies under the dining set.
    \item A flower shop with three flower buckets on the floor, each holding flowers. A wooden table displays potted plants. A potted bonsai tree sits in one corner, and two Christmas trees stand in another corner.
    \item A hotel lobby with a reception desk in the back of the room. A seating area in the center features two sofas facing each other with a coffee table between them. A luggage cart stands near the entrance, and a large chandelier hangs from the ceiling.
    \item A hotel room with a queen bed flanked by two nightstands, a desk with a chair, and a luggage rack in the corner.
    \item A restaurant with four dining tables, each surrounded by four chairs. A bar counter runs along one wall with six bar stools in front. Two potted plants flank the entrance, and pendant lights hang above each table.
    \item A cafe with a bar counter along one wall featuring a coffee machine, a filter coffee stand, and stacks of stone coffee cups. Three small round tables with two chairs each are arranged in the center. A chalkboard menu hangs on the wall behind the counter.
    \item A bar with a long wooden bar counter and five bar stools. Shelves behind the counter display bottles. Two high tables with two stools each are near the window.
    \item A diner with three booth tables against the wall, a jukebox in the corner, and a hostess stand near the entrance.
    \item A private office with a desk and office chair, a shelf with books against the wall, and two guest chairs facing the desk.
    \item An open office with four desks arranged in pairs facing each other. Each desk has a monitor and an office chair. A water cooler stands in the corner.
    \item A conference room with a long table surrounded by eight chairs. A projector screen hangs on the wall, and a whiteboard is mounted on another wall.
    \item A reception area with a reception desk, at least three waiting chairs, and a potted plant in the corner.
    \item A classroom with six student desks, each with a chair. A teacher's desk sits at the front near the chalkboard, which hangs on the wall.
    \item A library with shelves containing books, reading tables with chairs, and a commercial printer.
    \item A computer lab with six desks, each with a monitor, keyboard, and chair. A printer sits in the corner.
    \item An art studio with easels, stools, and a supply cabinet in the corner.
    \item A music room with a piano, a drum set, and music stands with chairs.
    \item A medical exam room with an exam table, a doctor's stool, and a cabinet with medical supplies and a blood pressure monitor.
    \item A hospital room with a hospital bed, an IV stand, a bedside table, and a visitor chair.
    \item A dental office with a dental chair, a dentist's stool, a dental light overhead, and a cabinet with dental tools.
    \item A waiting room with chairs along the walls, a coffee table with magazines, and a water dispenser in the corner.
    \item A physical therapy room with a treatment table, exercise mats on the floor, dumbbells, and a full-length mirror on the wall.
    \item A gym with two treadmills and two exercise bikes along the wall. A weight rack holds dumbbells. A squat rack has five 45-pound weight plates stacked next to it. Exercise balls are in the corner, and a large mirror covers one wall.
    \item A yoga studio with yoga mats on the floor, a mirror on the wall, and meditation cushions in the corner.
    \item A dance studio with a ballet barre along the wall, a large mirror, and a speaker in the corner.
    \item A spa treatment room with a massage table, a cabinet with towels, and candles on a round side table.
    \item A robotics lab with two workbenches with robotic arms. Tool shelves on the wall hold tools, and a 3D printer sits in the corner.
    \item A workshop with a wooden workbench, a stool, and a model airplane on the workbench. A tool box contains tools including a drill, hammer, screwdriver, and tape.
    \item A garage with a car, a workbench, tires stacked in the corner, and a ladder against the wall.
    \item A server room with three server racks, a fire extinguisher in the corner, and a desk with two monitors.
    \item An art gallery with at least five paintings on the walls, a sculpture on a pedestal in the center, and a bench for viewing.
    \item A commercial kitchen with a large stainless steel prep table, an industrial stove, a commercial refrigerator, and shelves with pots and pans. A utensil crock holds utensils including a spatula, tongs, and whisk. A cutting board with a chef's knife on top sits on the table.
    \item A bakery with a display case showing pastries, a counter with a cash register, and a bread rack with loaves of bread.
    \item A nail salon with two manicure tables, each with a chair and a stool, and a shelf with nail polish bottles.
    \item A barber shop with two barber chairs facing mirrors on the wall, and a waiting bench near the entrance.
    \item A tattoo parlor with a tattoo chair, a rolling cart with ink bottles, and framed artwork on the walls.
    \item A pet grooming salon with a grooming table, a bathtub, and a cage for pets in the corner.
    \item A laundromat with four washing machines along the wall, two dryers, a folding table, and a row of plastic chairs.
    \item A nursery with a crib, a rocking chair, a changing table, a baby mobile hanging from the ceiling, and a pile of wooden play blocks on the floor.
    \item A daycare room with small tables and chairs, a storage box with toys in the corner, and colorful bins on a shelf.
    \item A wine cellar with wooden shelves along the walls holding at least ten wine bottles. A tasting table with four stools sits in the center, with four wine glasses and a plate with cheese on it. A barrel is in the corner, and an antique chandelier hangs from the ceiling.
    \item A science lab with two lab benches, each with a microscope, a glass beaker, and a test tube stand. A fume hood hangs from the ceiling, and a safety shower is in the corner.
    \item An Egyptian museum room with a sarcophagus in the center. One display table holds a golden scarab beetle and a canopic jar. Another display table has an ancient scroll, a clay tablet, and a bronze amulet. Hieroglyphic panels are mounted on the walls, and two stone statues flank the entrance.
    \item An escape room with a locked chest, a bookshelf with books, a clock on the wall, and a mysterious painting.
\end{enumerate}

\subsection{Object Density (25 prompts)}

\begin{enumerate}[nosep, leftmargin=*]
\footnotesize
    \item A pantry with a single three-level shelf. The first level holds at least 10 cans, the second level holds at least 10 pasta packs, and the third level holds at least 10 potatoes.
    \item A kitchen with a table. There is a fruit bowl on the table with a mix of apples and oranges. The bowl contains at least 6 fruits.
    \item A kitchen with a single shelf. The shelf holds at least 8 plates, 8 bowls, and 8 cups.
    \item A kitchen with a spice rack on the wall. The rack holds at least 10 spice jars.
    \item A kitchen with an open refrigerator. The refrigerator contains at least 6 bottles and 6 food containers.
    \item A kitchen. There is a utensil holder on the counter. The holder contains at least 4 forks, 4 knives, and 4 spoons.
    \item A dining room with a table set for 4 people. Each setting has a plate, a glass, a fork, and a knife.
    \item A dining room with a table set for 8 people. Each setting has a plate, a glass, a fork, and a knife.
    \item A dining room with a long table set for 12 people. Each setting has a small plate stacked on a large plate, a glass, a fork, and a knife.
    \item A kitchen with a mug rack on the wall holding at least 8 mugs.
    \item A bedroom with a breakfast tray on the bed. The tray has a plate, a cup, a bowl, and a glass.
    \item A living room with a coffee table. On the table is a tea set with a teapot, 4 teacups on saucers.
    \item A bar with a counter. On the counter are at least 8 glasses and 6 bottles.
    \item A dining room with a wine shelf holding at least 12 wine bottles.
    \item A pottery store with shelves along the walls. The shelves hold at least 30 cups and 30 bowls. There is a table with a large painted bowl.
    \item A bathroom with a counter. On the counter are at least 4 bottles, 2 tubes, and a soap dispenser.
    \item A bathroom with a medicine cabinet. The cabinet contains at least 6 medicine bottles, 4 pill boxes, 2 tubes, and a pair of scissors.
    \item A kitchen with a three-level shelf. The top level holds at least 10 jars. The other two levels hold stacks of plates with at least 20 plates in total.
    \item A closet with a shoe rack holding exactly 5 pairs of shoes.
    \item A bedroom with a vanity table. On the table are at least 6 makeup bottles, at least 2 brushes, and a small mirror. There is also a nightstand with a stack of 8 books.
    \item An office with a desk. On the desk is a pen holder with at least 5 pens and a laptop. There is a shelf with stacks of books, at least 15 books in total.
    \item A kitchen with a sink. The sink contains a mess of dishes with at least 5 plates, 3 glasses, and 5 bowls.
    \item A playroom with two bins. One bin contains 2 teddy bears. The other bin contains at least 5 wooden blocks. There is a carpet with a toy car and a toy train. There is a pile of lego blocks on the carpet containing at least 5 blocks.
    \item A restaurant with exactly 4 tables and 8 chairs. Each table is set for 2 people with a plate, a glass, a fork, and a knife.
    \item A bookstore with at least 50 books.
\end{enumerate}

\subsection{Themed Scenes (10 prompts)}

\begin{enumerate}[nosep, leftmargin=*]
\footnotesize
    \item A Star Wars themed teen bedroom. The room features a bed with Star Wars bedding showing characters and starships, and a Star Wars themed desk lamp with Darth Vader or Death Star designs.
    \item A playroom inspired by The Incredibles movie. The room features an Incredibles themed rug with the superhero family logo and two Incredibles themed bean bags for seating.
    \item An Art Deco styled living room. The room features an Art Deco style sofa with geometric patterns, an Art Deco style coffee table positioned in front of the sofa, and an Art Deco style floor lamp with a geometric shade.
    \item A Minecraft themed gaming room. The room features a Minecraft themed chair with Creeper designs and a Minecraft themed poster on the wall.
    \item A Picasso-inspired dining room. The room features a dining table with four chairs surrounding it. A Picasso-style cubist painting hangs on the wall, and a tablecloth with abstract geometric patterns covers the table.
    \item A steampunk styled home office. The room features a steampunk desk with brass fittings and exposed gears, a steampunk style desk lamp with copper pipes and vintage bulb, a steampunk themed wall clock with visible clockwork mechanisms, and a steampunk styled office chair with leather and metal accents.
    \item A Harry Potter themed home library. The room features a bookshelf styled like a Hogwarts library, a Harry Potter themed reading chair with a Harry Potter themed floor lamp next to it, and a Hogwarts trunk for storage.
    \item A Pop Art styled artist studio. The room features a Pop Art style desk with bold colors, a Pop Art style stool positioned in front of the desk, a large Pop Art style painting with comic-style imagery on the wall, and a Pop Art themed floor lamp.
    \item A Jurassic Park themed kids bedroom. The room features a bed with Jurassic Park dinosaur bedding, a Jurassic Park themed rug on the floor, and a Jurassic Park poster on the wall. Two dinosaur stuffed animals sit on the bed, and a dinosaur themed desk lamp rests on the nightstand.
    \item A Frozen themed nursery. The room features a Frozen themed crib with Elsa and Anna bedding, a Frozen themed rocking chair, a Frozen themed rug with snowflake patterns on the floor, a Frozen themed lamp on a side table, a Frozen poster on the wall, and a Frozen themed toy box for storage.
\end{enumerate}

\subsection{SceneEval-100 (House)~\cite{tam2025sceneeval} (6 prompts)}

\begin{enumerate}[nosep, leftmargin=*]
\footnotesize
    \item A medium-sized bedroom featuring a queen-size bed with a nightstand on each side. A TV is mounted on the wall directly across from the bed, providing convenient viewing. The room includes a walk-in closet equipped with three wardrobes for clothes and two large mirrors.
    \item A master bedroom with a king-size bed, an attached walk-in closet, and a master bathroom. A desk is positioned in one corner of the room for working, there is a laptop on the right side of the desk and a wallet on the left side. There are also two sofa chairs next to the window. There is a sink mounted in the bathroom with a soap on the left side. 
    \item The space includes three rooms: a spacious living room with an L-shaped sectional sofa facing a wall-mounted TV and a rectangular coffee table in front, holding a small serving tray with napkins. The dining room features an extending dining table surrounded by six chairs, a sleek sideboard with cups and sodas on top, two wall sconces providing soft lighting, and a small bar cart loaded with bottles in the corner. The modern kitchen boasts a central bar table with four bar stools on the long side, a bar table holding a charcuterie board with plates beside it, and a wine cooler in the corner.
    \item This is a modern bedroom with a spa-like bathroom. The bedroom has a queen-size bed with vibrant pillows. A sleek dresser sits against one wall, adorned with a small jewelry box and a decorative mirror. A reading corner includes an armchair and a small side table. Adjacent to the bedroom, the bathroom features a large bathtub next to a frosted window. A vanity opposite the tub holds two sinks with skincare products.
    \item This cozy bedroom features a full-size bed against a wall with two nightstands on each side where the left one has a small clock. A small desk sits in the corner with a comfortable chair for studying or working. Opposite the bed, a spacious dresser provides additional storage. Adjacent to the bedroom, the living room has a recliner and an ottoman facing a low coffee table with a small vase and magazines. A large bookshelf in the corner holds books and board games, while a wide couch provides space for gatherings. Just off the living room, a small gaming room with a gaming console and two beanbag chairs offers a dedicated space for entertainment, complete with a small shelf for controllers and headsets.
    \item This bedroom has a full-size bed against a wall with a cartoon painting above. Two nightstands on each side of the bed provide storage. A writing desk in the corner holds a modern lamp for task lighting. A large wardrobe stands next to the door for additional storage. Beside the bedroom is a cozy living room with a large sofa facing a sleek coffee table where puzzles and decorative coasters reside. A bookshelf in the corner displays books and family pictures. The stylish bathroom includes a bathtub, with a toilet in the corner. The vanity features a double sink, with toiletries and white candles on top. A wicker basket at the foot of the bathtub holds towels.
\end{enumerate}

\subsection{House-Level (25 prompts)}
\label{app:prompts_house}

\begin{enumerate}[nosep, leftmargin=*]
\footnotesize
    \item A compact studio apartment with a single open-plan room and a bathroom. The studio contains a bed against the wall, a sofa, and a desk with a chair. The bathroom has a toilet, a sink, and a blue bathtub.
    \item A hotel room with a bedroom and an en-suite bathroom. The bedroom has a bed with two nightstands on each side, a desk with a chair, and a TV mounted on the wall. The bathroom has a toilet, a sink, and a shower.
    \item A one-bedroom apartment with a bedroom, a living room, and a bathroom. The bedroom has a bed and a wardrobe. The living room has a sofa, a coffee table, and a TV on a TV stand. The bathroom has a toilet, a sink, and a bathtub.
    \item A cozy guest cottage with a bedroom, a sitting area, and a bathroom. The bedroom has a bed with a nightstand and a lamp on top. The sitting area has two armchairs and a wooden bookshelf full of books. A skull sits on the bookshelf. The bathroom has a toilet and a sink.
    \item A small office suite with a reception, a private office, and a bathroom. The reception has a reception desk with a computer on it and two chairs. The private office has an office desk with a chair and a filing cabinet. The bathroom has a toilet and a sink.
    \item A massage parlor with a treatment room, a waiting area, and a bathroom. The treatment room has a massage table and a shelf with bottles. The waiting area has a sofa and a coffee table with books on it. The bathroom has a toilet and a sink.
    \item A tiny house with a bedroom, a living-kitchen, and a bathroom. The bedroom has a bed and a wooden dresser. The living-kitchen has a sofa, a dining table with two chairs, and a refrigerator. The bathroom has a toilet and a sink.
    \item A hair salon with a salon floor, a waiting area, a storage room, and a bathroom. The salon floor has two salon chairs positioned in front of two mirrors on the wall, facing them. The waiting area has a sofa and a coffee table. The storage room has shelves with bottles. The bathroom has a toilet and a sink.
    \item A dorm suite with a bedroom, a study room, a common\_room, and a bathroom. The bedroom has a bed and a wardrobe. The study room has a desk with a chair and a bookshelf with books. The common\_room has a refrigerator and a microwave on a counter. The bathroom has a toilet and a sink.
    \item A small retail shop with a showroom, a fitting room, a storage room, and a bathroom. The showroom has a display table with clothes on it and a cash register on a counter. The fitting room has a mirror and a bench. The storage room has shelves with boxes. The bathroom has a toilet and a sink.
    \item A two-bedroom apartment with a master bedroom, a second bedroom, a living room, a hallway, and a bathroom. The master bedroom has a bed and a nightstand. The second bedroom has a bed and a desk with a chair. The living room has a sofa, a coffee table, and a TV. The hallway has a coat rack. The bathroom has a toilet, a sink, and a bathtub.
    \item A boutique hotel suite with a bedroom, a living area, a walk-in closet, a dressing room, and a bathroom. The bedroom has a king-size bed with two nightstands. The living area has a sofa and a coffee table. The walk-in closet has a wardrobe and a luggage rack. The dressing room has a vanity with a mirror. The bathroom has a toilet, a sink, and a bathtub.
    \item A small restaurant with a dining area, a kitchen, a bar, a storage room, and a bathroom. The dining area has four dining tables, each with two chairs. The kitchen has a stove and a refrigerator. The bar has a bar counter with three bar stools. The storage room has shelves with supplies including plates and cups. The bathroom has a toilet and a sink.
    \item A small family home with a master bedroom, a kids room, a living-dining room, a kitchen, a hallway, and a bathroom. The master bedroom has a bed and a dresser. The kids room has a bunk bed and a toy box. The living-dining room has a sofa, a dining table with four chairs, and a TV. The kitchen has a refrigerator and a stove. The hallway has a coat rack and a shoe cabinet. The bathroom has a toilet, a sink, and a bathtub.
    \item A dental office with a reception-waiting area, two exam rooms, an X-ray room, a corridor, and a bathroom. The reception-waiting area has a reception desk with a computer and three chairs. The first exam room has a dental chair and a cabinet. The second exam room has a dental chair and a sink. The X-ray room has an X-ray machine. The corridor has a bench. The bathroom has a toilet and a sink.
    \item An art gallery with three exhibition rooms, an office, a storage room, and a bathroom. The first exhibition room has two paintings on the wall and a bench. The second exhibition room has three sculptures on pedestals. The third exhibition room has a painting on the wall with a bench in front for viewing. The office has a desk with a chair and a computer. The storage room has shelves with crates. The bathroom has a toilet and a sink.
    \item A co-working space with an open office, two meeting rooms, a lounge, a kitchen, a reception, and a bathroom. The open office has four desks with chairs and computers. The first meeting room has a conference table with six chairs. The second meeting room has a whiteboard and four chairs. The lounge has two sofas and a coffee table. The kitchen has a refrigerator and a microwave on a counter. The reception has a reception desk with a computer. The bathroom has a toilet and a sink.
    \item A photography studio with a shooting area, an editing room, a client lounge, a dressing room, a props storage, an office, and a bathroom. The shooting area has two studio lights and a backdrop. The editing room has a desk with two monitors and a chair. The client lounge has a sofa and a coffee table. The dressing room has a vanity with a mirror on top and a clothing rack. The props storage has shelves with props including hats and bags. The office has a desk with a chair and a filing cabinet. The bathroom has a toilet and a sink.
    \item A small fitness studio with a workout area, a yoga room, a locker room, a reception, an office, a storage room, and a bathroom. The workout area has a treadmill, a weight rack with dumbbells, and a squat rack with a stack of weight plates next to it. The yoga room has yoga mats and a mirror on the wall. The locker room has lockers and a bench. The reception has a reception desk with a computer. The office has a desk with a chair. The storage room has shelves with towels. The bathroom has a toilet and a sink.
    \item A pet clinic with a reception, a waiting room, two exam rooms, a surgery room, a corridor, and a bathroom. The reception has a reception desk with a computer. The waiting room has four chairs and a water dispenser. The first exam room has an exam table and a cabinet. The second exam room has an exam table and a scale. The surgery room has a surgery table and a surgical light. The corridor has a bench. The bathroom has a toilet and a sink.
    \item A large family home with a master bedroom, two kids bedrooms, a living room, a kitchen, a dining room, a hallway, and two bathrooms. The master bedroom has a bed with two nightstands and a dresser. The first kids bedroom has a bed and a desk with a chair. The second kids bedroom has a bunk bed and a bookshelf with books. The living room has a sofa, a coffee table, and a TV. The kitchen has a refrigerator, a stove, and a dining table with four chairs. The dining room has a dining table with six chairs and a sideboard. The hallway has a coat rack and a mirror on the wall. The first bathroom has a toilet, a sink, and a bathtub. The second bathroom has a toilet and a sink.
    \item A medical clinic with a reception, a waiting room, four exam rooms, a lab, a pharmacy, a corridor, and two bathrooms. The reception has a reception desk with a computer. The waiting room has six chairs and a water dispenser. The first exam room has an exam table and a cabinet. The second exam room has an exam table and a scale. The third exam room has an exam table and a desk with a chair. The fourth exam room has an exam table and a stool. The lab has a lab bench and a microscope. The pharmacy has shelves with bottles. The corridor has a bench. The first bathroom has a toilet and a sink. The second bathroom has a toilet and a sink.
    \item A small office building with a reception, five offices, two conference rooms, a break room, a server room, a corridor, and a bathroom. The reception has a reception desk with a computer and two chairs. The first office has a desk with a chair and a computer. The second office has a desk with a chair and a filing cabinet. The third office has a desk with a chair and a bookshelf. The fourth office has a desk with a chair and a printer. The fifth office has a desk with a chair and a whiteboard. The first conference room has a conference table with eight chairs. The second conference room has a conference table with six chairs, a projector, and a whiteboard on the wall. The break room has a refrigerator, a microwave on a counter, and a table with four chairs. The server room has two server racks. The corridor has a water dispenser. The bathroom has a toilet and a sink.
    \item A community center with a main hall, four activity rooms, a kitchen, an office, a reception, a locker room, a hallway, two bathrooms, and a storage room. The main hall has a stage and rows of chairs. The first activity room has tables and chairs for crafts. The second activity room has yoga mats and a mirror on the wall. The third activity room has a ping pong table. The fourth activity room has easels and stools for art class. The kitchen has a refrigerator, a stove, and a counter. The office has a desk with a chair and a computer. The reception has a desk with a computer. The locker room has lockers and a bench. The hallway has a bulletin board on the wall. The first bathroom has a toilet and a sink. The second bathroom has a toilet and a sink. The storage room has shelves with boxes.
    \item A boutique hotel with a reception, four guest bedrooms with en-suite bathrooms, a restaurant, a kitchen, a gym, a lounge, and a corridor. The reception has a desk with a computer and two armchairs. The first guest bedroom has a bed with two nightstands and its bathroom has a toilet, a sink, and a shower. The second guest bedroom has a bed and a desk and its bathroom has a toilet, a sink, and a bathtub. The third guest bedroom has a bed and an armchair and its bathroom has a toilet, a sink, and a shower. The fourth guest bedroom has a bed and a wardrobe and its bathroom has a toilet, a sink, and a bathtub. The restaurant has four dining tables with chairs. The kitchen has a refrigerator, a stove, and a counter. The gym has a treadmill and a weight rack with dumbbells. The lounge has two sofas and a coffee table. The corridor has a console table with a vase on top.
\end{enumerate}

\section{User Study Interface Screenshots}
\label{app:interface_screenshots}

This section provides screenshots of the user study interface, illustrating the viewing modes, interactive features, and question format.

\begin{figure}[htbp]
    \centering
    \includegraphics[width=\textwidth]{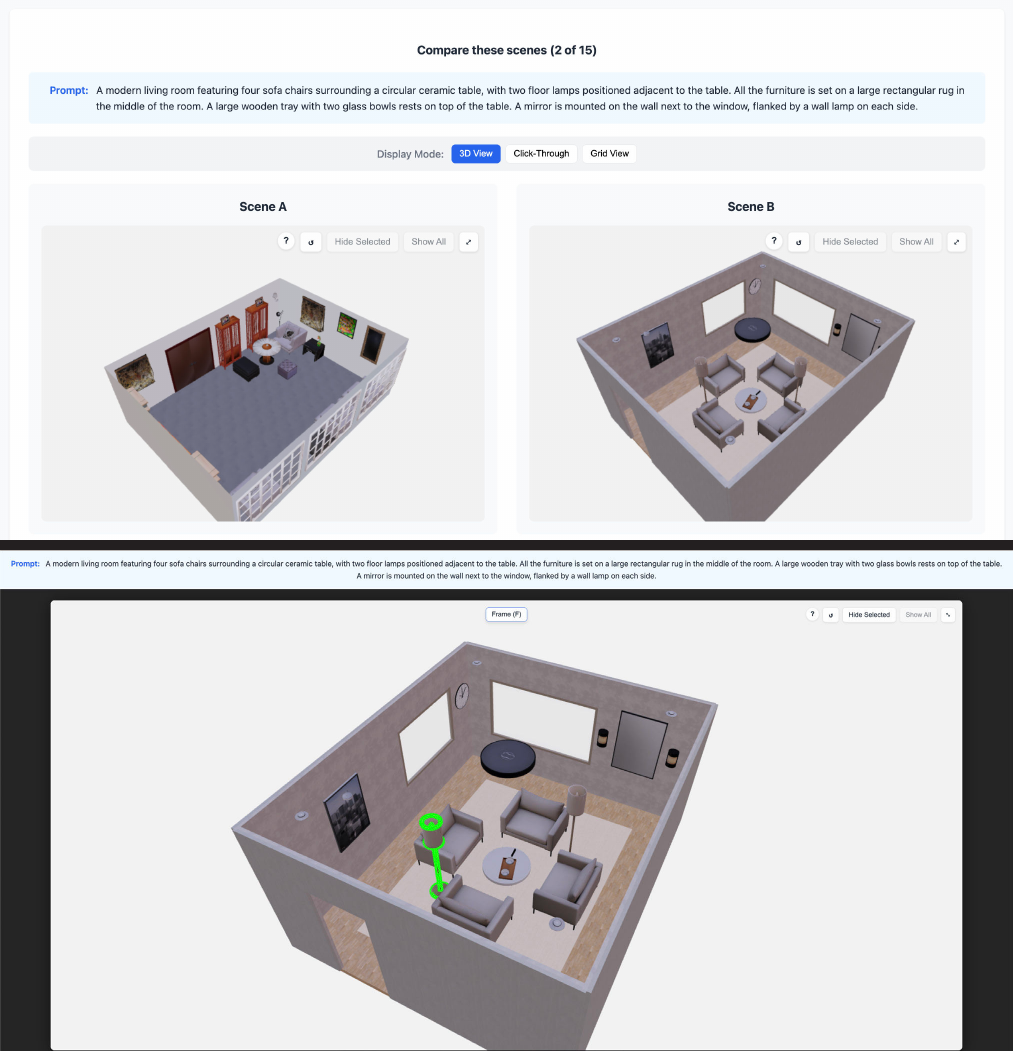}
    \caption{User study interface: 3D viewer mode. \textbf{Top:} Side-by-side comparison with interactive Babylon.js 3D viewers. Participants can orbit, pan, and zoom each scene independently. Controls include Hide Selected, Show All, Reset, and Fullscreen. \textbf{Bottom:} Fullscreen mode with object selection. Clicking an object highlights it and enables the Frame button, which focuses the camera on the selected object and zooms in.}
    \label{fig:user_study_3d}
\end{figure}

\begin{figure}[htbp]
    \centering
    \includegraphics[width=\textwidth]{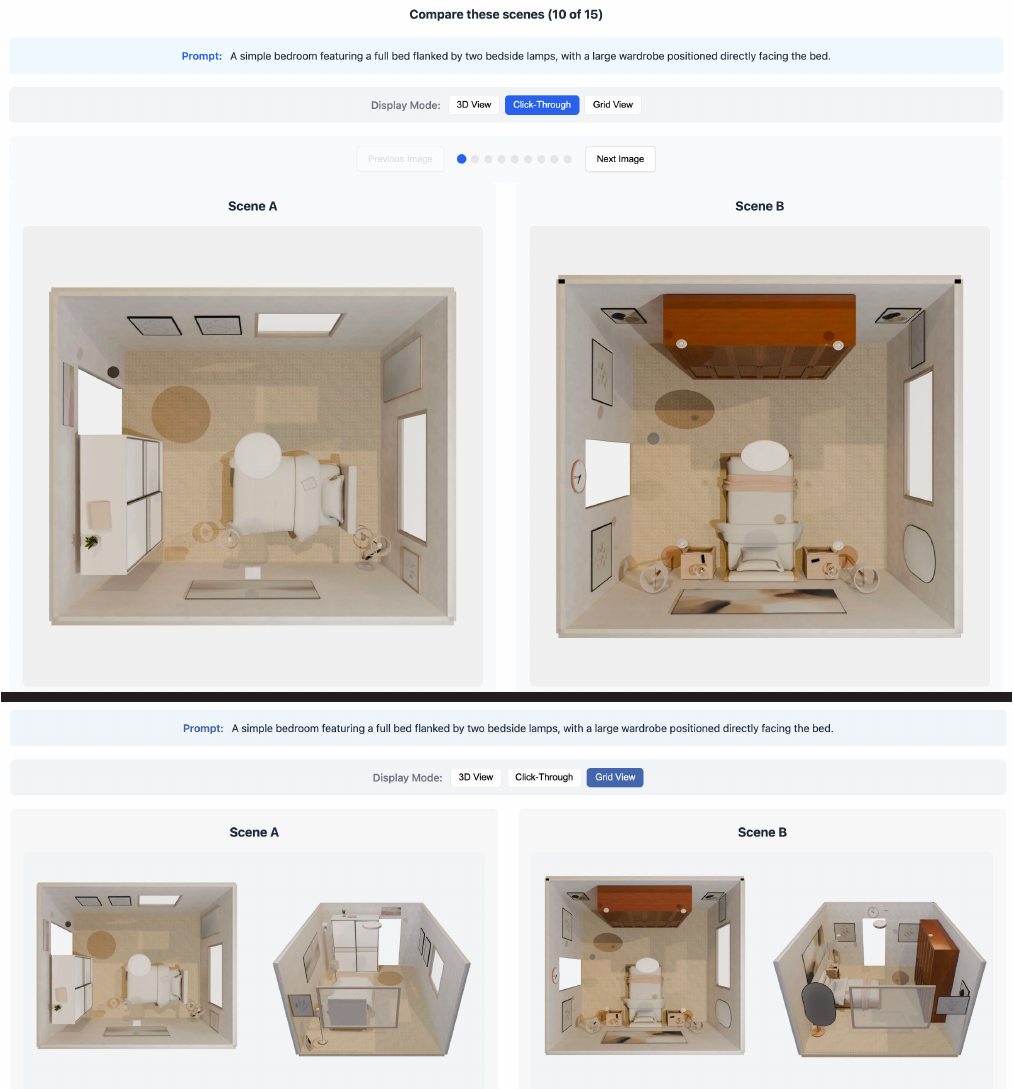}
    \caption{User study interface: Alternative viewing modes for participants who prefer static images or have slower devices. \textbf{Top:} Click-Through mode with carousel navigation through 9 pre-rendered camera angles. \textbf{Bottom:} Grid View mode displaying all angles simultaneously for quick comparison.}
    \label{fig:user_study_images}
\end{figure}

\begin{figure}[htbp]
    \centering
    \includegraphics[width=\textwidth]{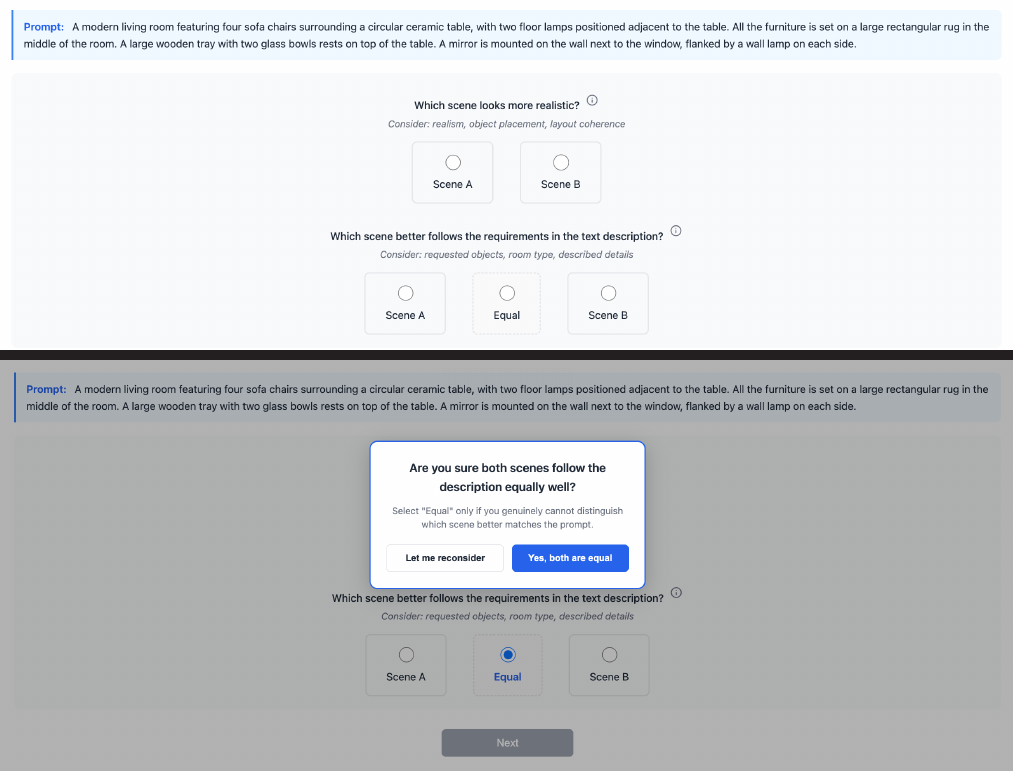}
    \caption{User study questions. \textbf{Top:} The two evaluation questions. Q1 (Realism) is forced-choice between Scene A and B; Q2 (Faithfulness) includes an ``Equal'' option. \textbf{Bottom:} Confirmation dialog shown when participants select ``Equal,'' encouraging them to reconsider and only confirm if they genuinely cannot distinguish which scene better matches the prompt.}
    \label{fig:user_study_questions}
\end{figure}

\begin{figure}[htbp]
    \centering
    \includegraphics[width=\textwidth]{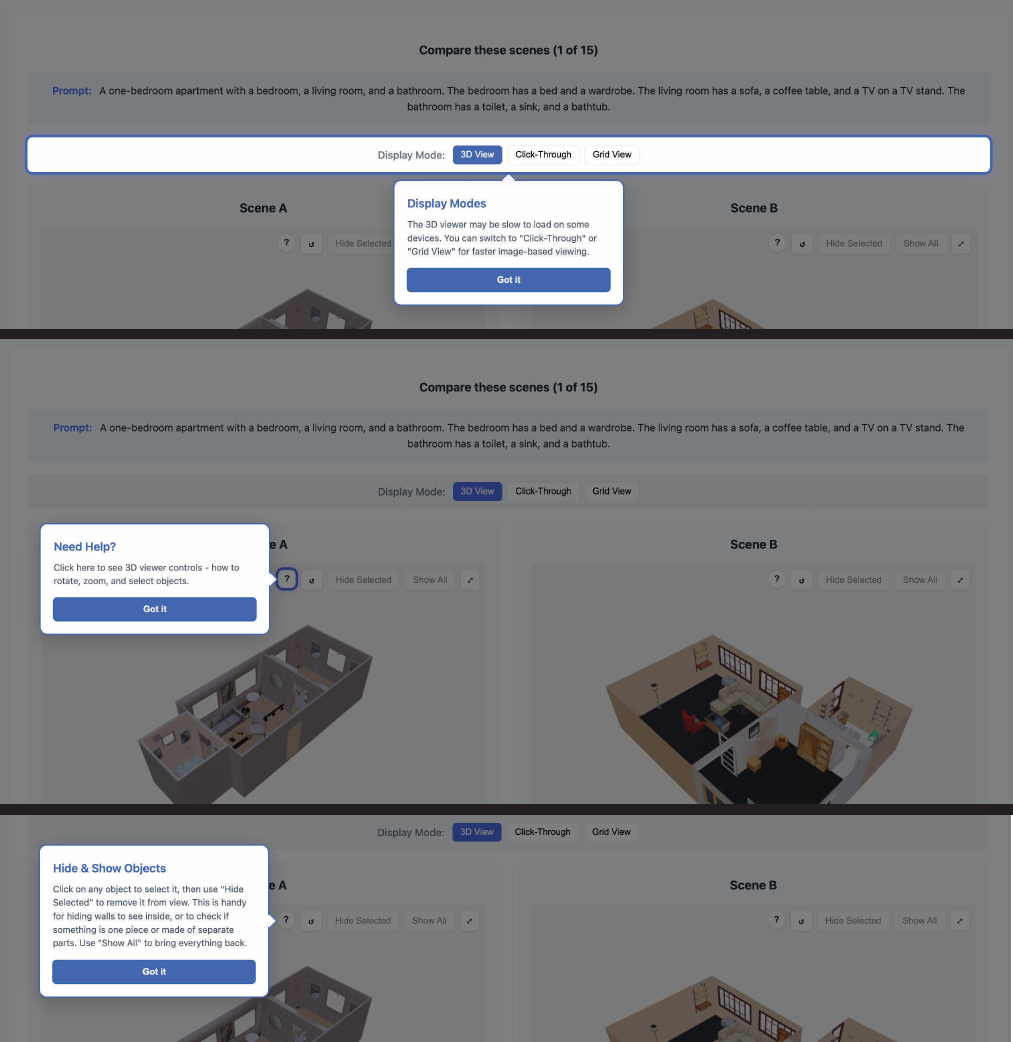}
    \caption{User study tutorial tooltips (selection). The interface includes an interactive onboarding sequence with contextual tooltips. Shown here: display mode switching (3D View, Click-Through, Grid View), the help button for 3D viewer controls, and the Hide Selected/Show All functionality for inspecting occluded objects or verifying object composition.}
    \label{fig:user_study_tutorial}
\end{figure}

\section{Additional Qualitative Results}
\label{app:qualitative_results}

\begin{figure}[htbp]
\centering
\includegraphics[width=\textwidth]{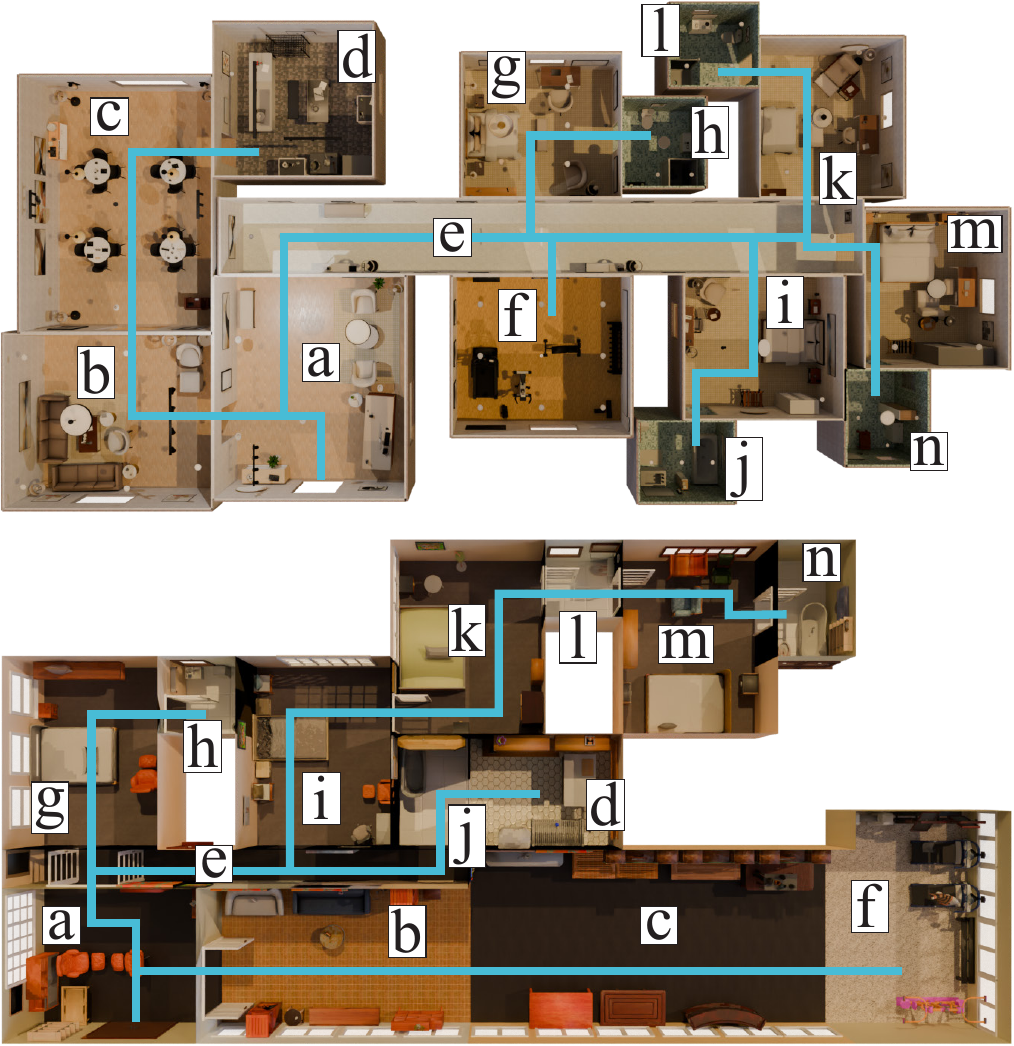}
\caption{\textbf{Room connectivity comparison for a hotel scene.} Both SceneSmith (top) and Holodeck (bottom) contain all the requested rooms (House-level prompt 25; Appendix~\ref{app:prompts_house}): reception (a), lounge (b), restaurant (c), kitchen (d), corridor (e), gym (f), four guest rooms (g, i, k, m), and four bathrooms (h, j, l, n). Cyan lines show room connectivity paths. SceneSmith generates realistic floor plans where the entrance leads through the reception, en suite bathrooms are only accessible through associated guest rooms, the kitchen is connected to the restaurant, and all rooms connect via a central corridor. Holodeck produces implausible layouts where rooms may only be reachable through other guest rooms or bathrooms. For example, the kitchen (d) is only reachable through bathroom (j), and guest room (m) is only reachable by traversing through guest rooms (i) and (k) and bathroom (l).}
\label{fig:hotel_connectivity}
\end{figure}

\subsection{Baseline Comparisons}
\begin{figure}[htbp]
\centering
\includegraphics[width=\textwidth]{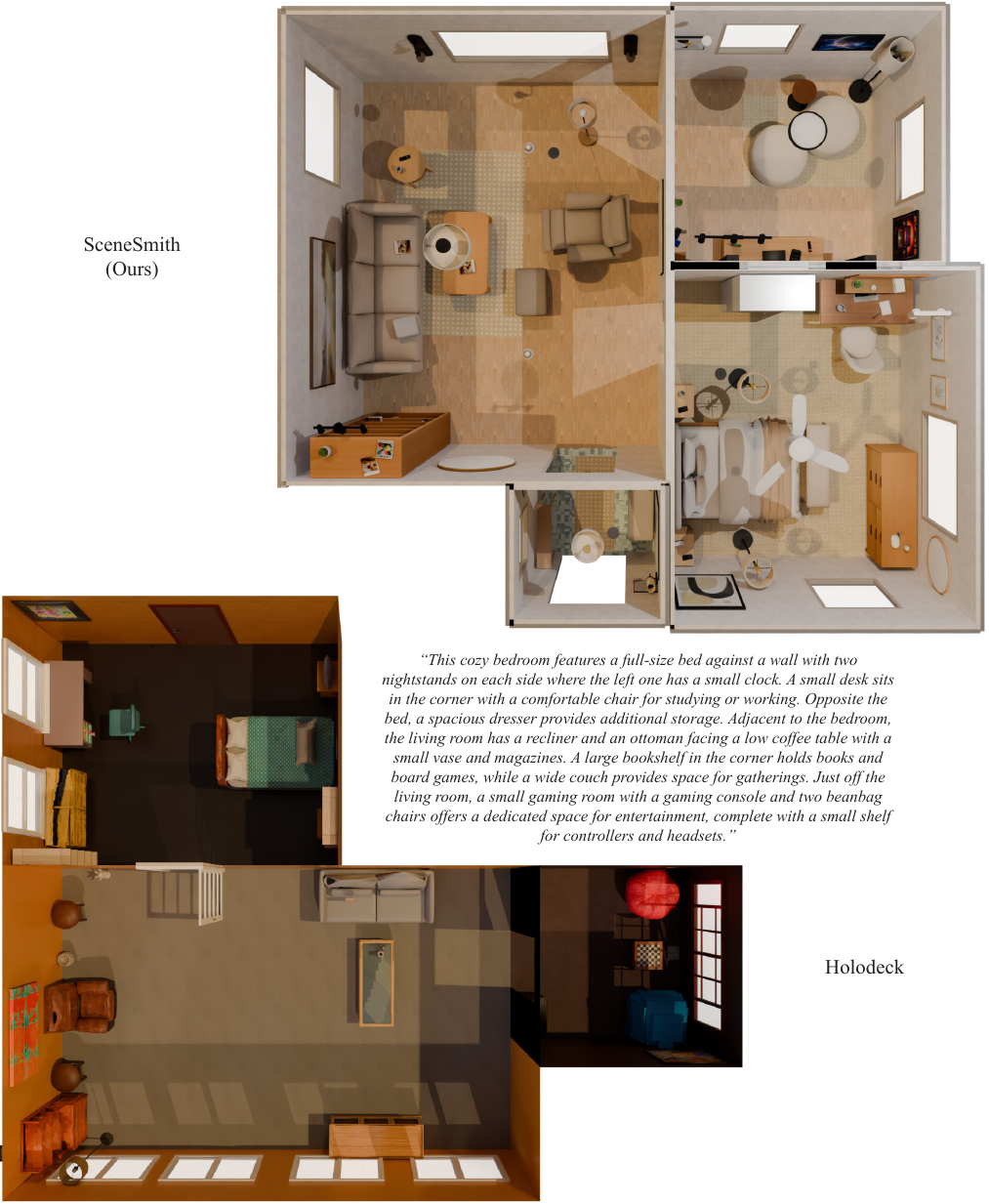}
\caption{\textbf{House-level qualitative comparison.} SceneSmith (top) vs Holodeck (bottom) for a multi-room prompt.}
\label{fig:baseline_comparison_house}
\end{figure}

\begin{figure}[htbp]
\centering
\includegraphics[width=\textwidth]{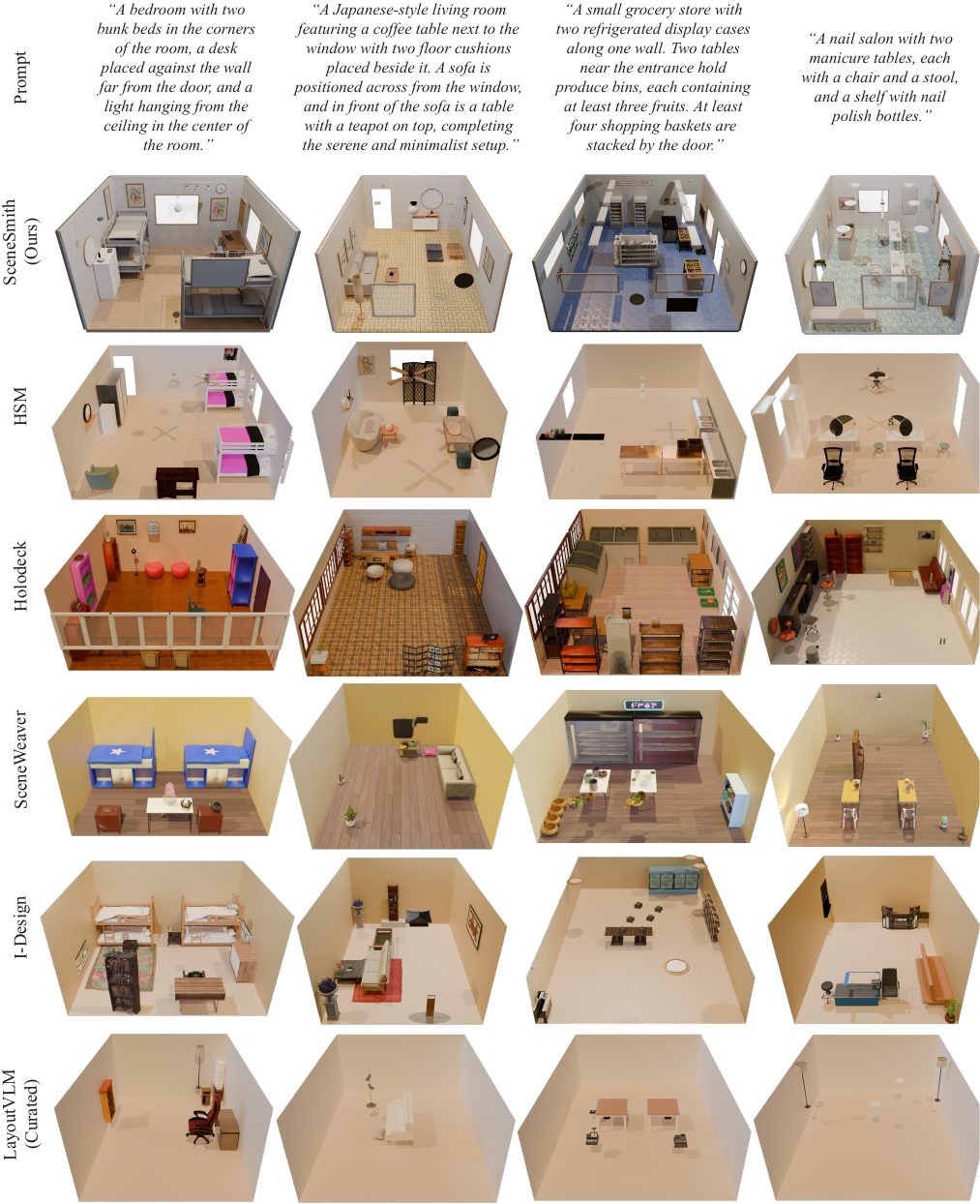}
\caption{\textbf{Room-level qualitative comparison.} SceneSmith compared to six baselines across four diverse prompts: bunk bed bedroom, Japanese-style living room, grocery store, and nail salon. Baselines are ordered by their user study realism score. SceneSmith generates scenes with appropriate object density, realistic arrangements, and faithful prompt following.}
\label{fig:baseline_comparison_room}
\end{figure}

Figure~\ref{fig:hotel_connectivity} compares room connectivity between SceneSmith and Holodeck for a hotel scene. Figures~\ref{fig:baseline_comparison_house} and~\ref{fig:baseline_comparison_room} provide qualitative comparisons between SceneSmith and baseline methods for house-level and room-level prompts respectively.

\subsection{Additional SceneSmith Examples}
\begin{figure}[htbp]
\centering
\includegraphics[width=\textwidth]{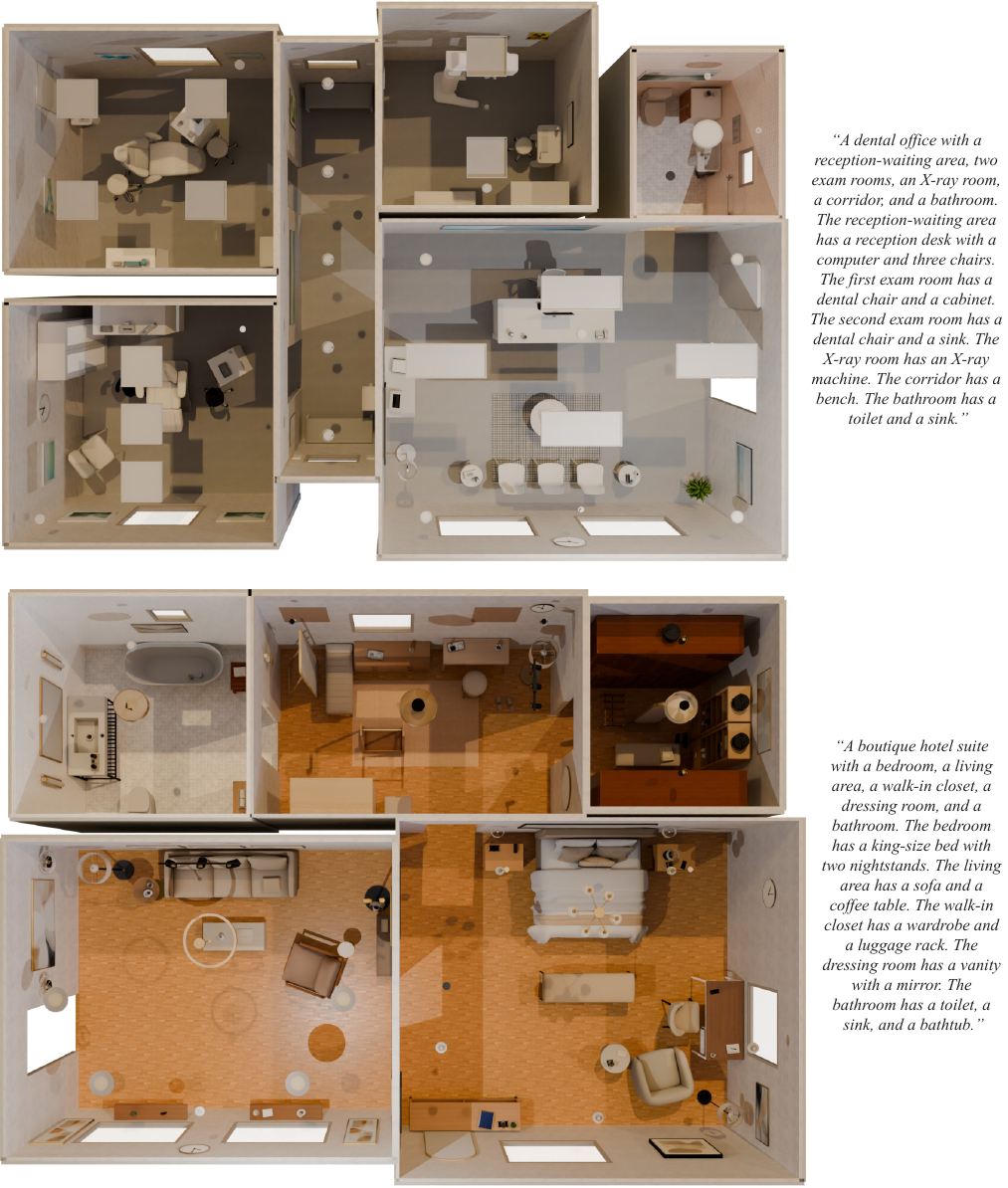}
\caption{\textbf{House-level generation examples.} Two multi-room scenes generated by SceneSmith: a dental office with reception area, exam rooms, X-ray room, and bathroom; and a boutique hotel suite with bedroom, living area, walk-in closet, dressing room, and bathroom.}
\label{fig:ours_house_qualitative}
\end{figure}

\begin{figure}[htbp]
\centering
\includegraphics[width=\textwidth]{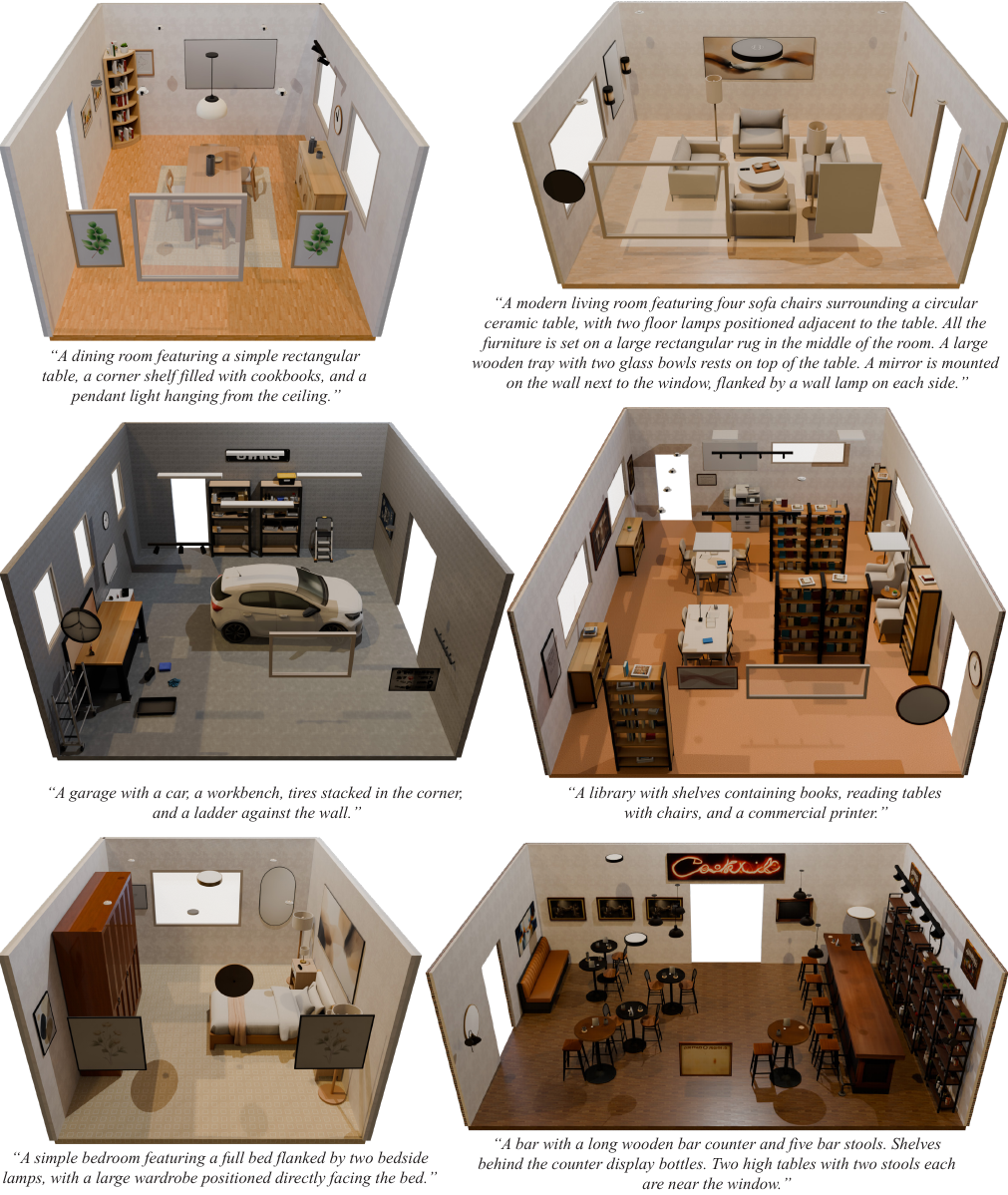}
\caption{\textbf{Additional room-level generation examples.} Each panel shows a text prompt and the corresponding scene generated by SceneSmith.}
\label{fig:ours_room_qualitative_1}
\end{figure}

\begin{figure}[htbp]
\centering
\includegraphics[width=\textwidth]{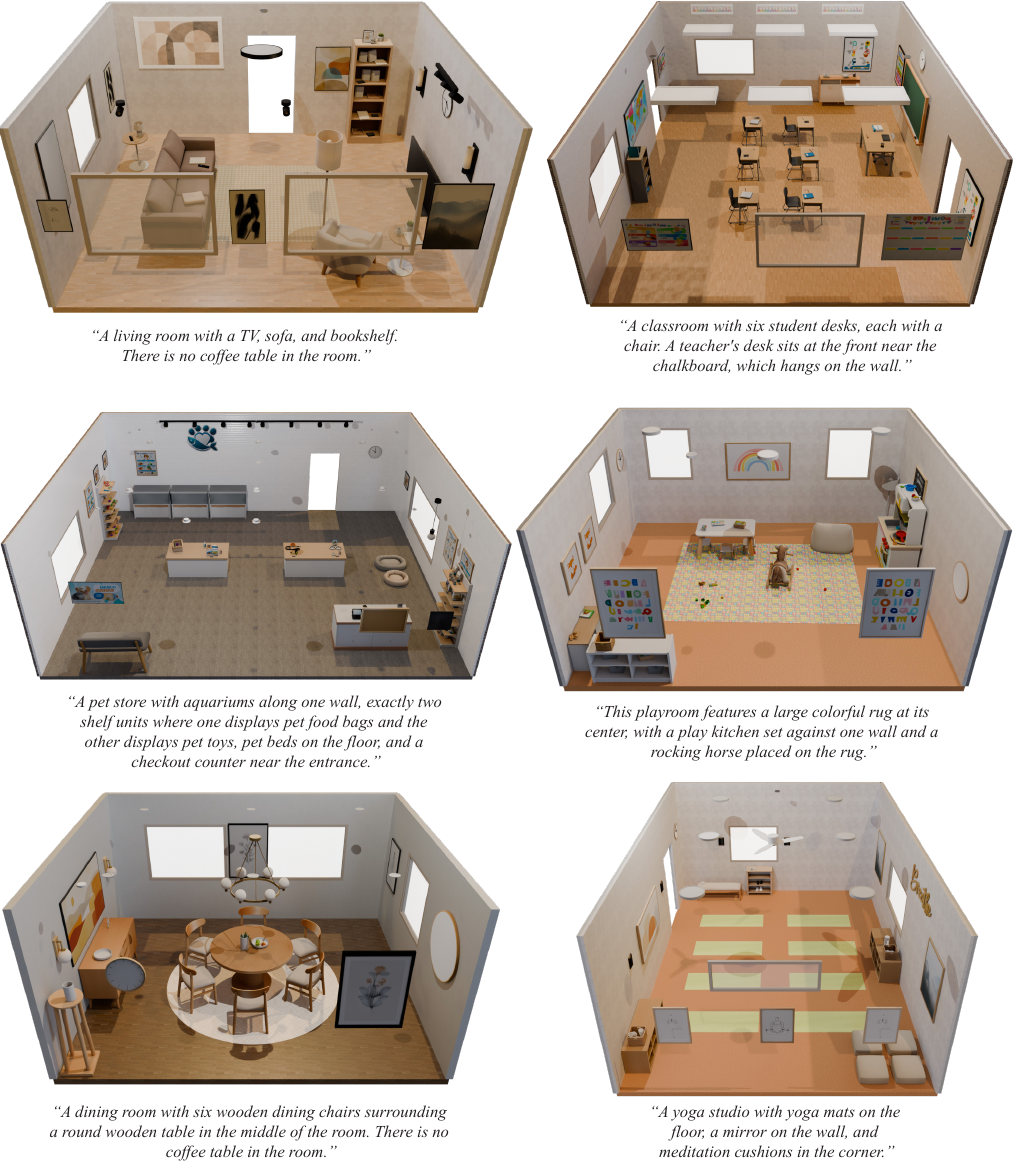}
\caption{\textbf{Additional room-level generation examples.} Each panel shows a text prompt and the corresponding scene generated by SceneSmith.}
\label{fig:ours_room_qualitative_2}
\end{figure}

\begin{figure}[htbp]
\centering
\includegraphics[width=\textwidth]{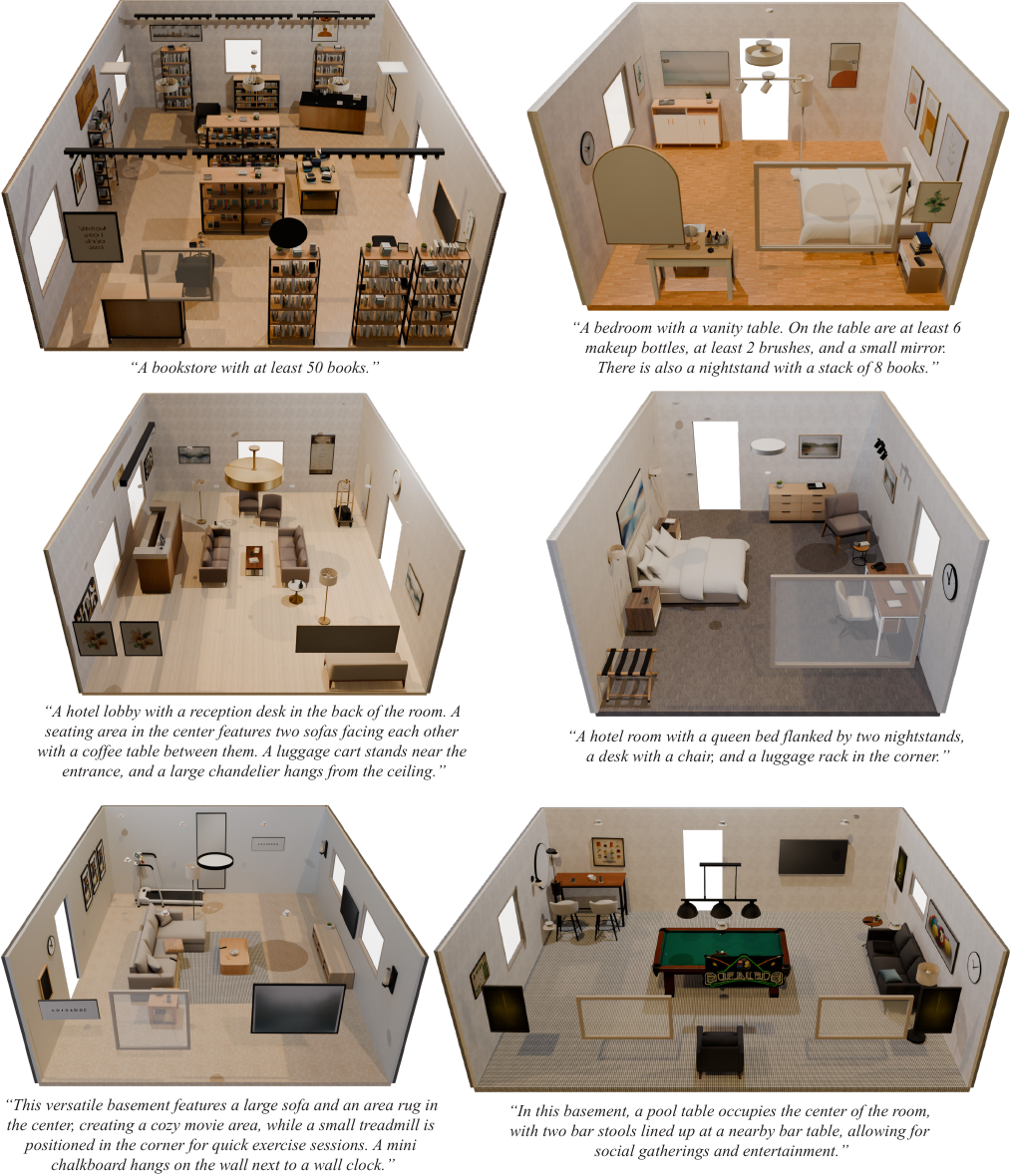}
\caption{\textbf{Additional room-level generation examples.} Each panel shows a text prompt and the corresponding scene generated by SceneSmith.}
\label{fig:ours_room_qualitative_3}
\end{figure}

\begin{figure}[htbp]
\centering
\includegraphics[width=\textwidth]{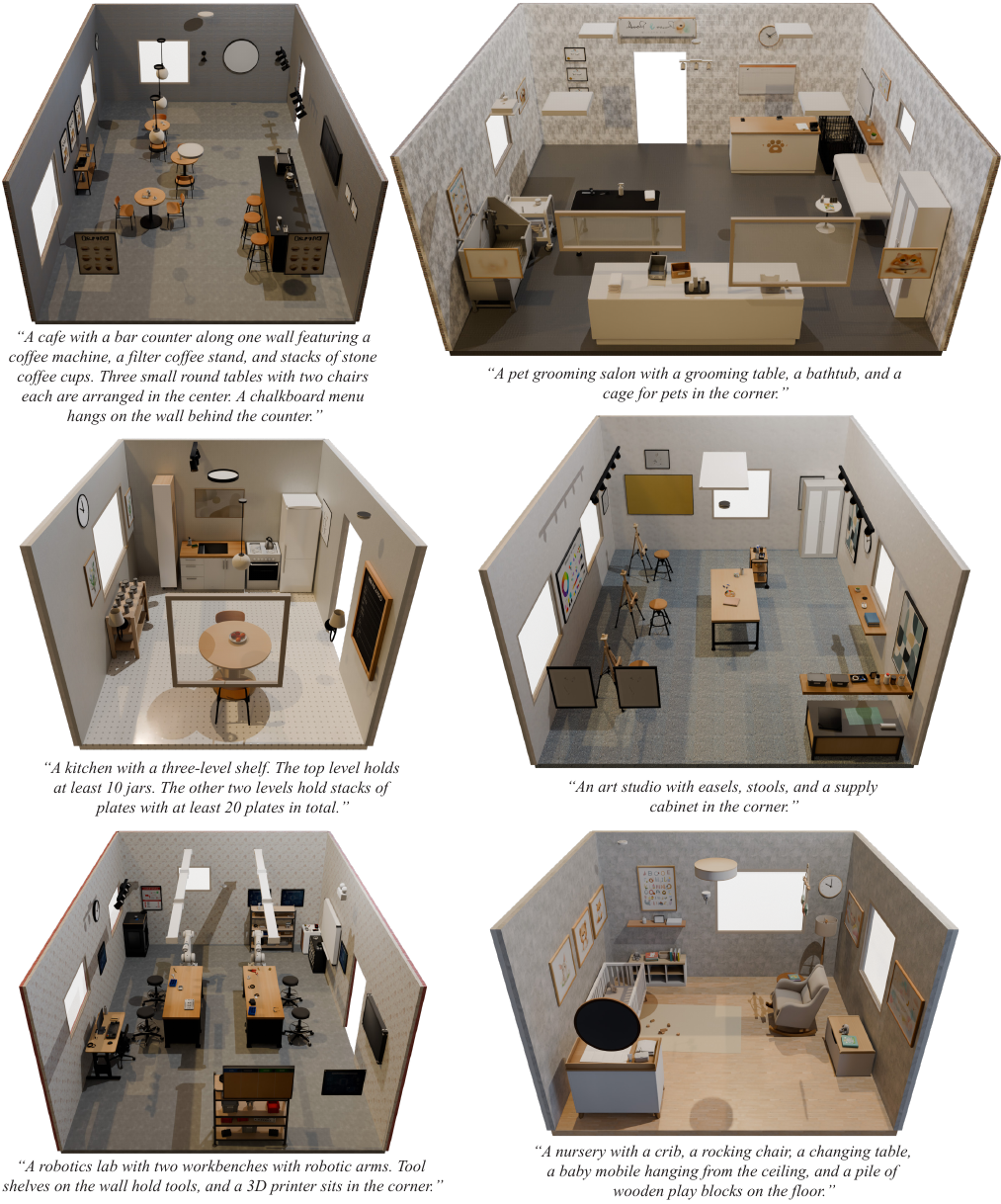}
\caption{\textbf{Additional room-level generation examples.} Each panel shows a text prompt and the corresponding scene generated by SceneSmith.}
\label{fig:ours_room_qualitative_4}
\end{figure}

Figures~\ref{fig:ours_house_qualitative}--\ref{fig:ours_room_qualitative_4} show additional scenes generated by SceneSmith across diverse room types and complexity levels.

\subsection{Manipuland and Themed Generation}
\begin{figure}[htbp]
\centering
\includegraphics[width=\textwidth]{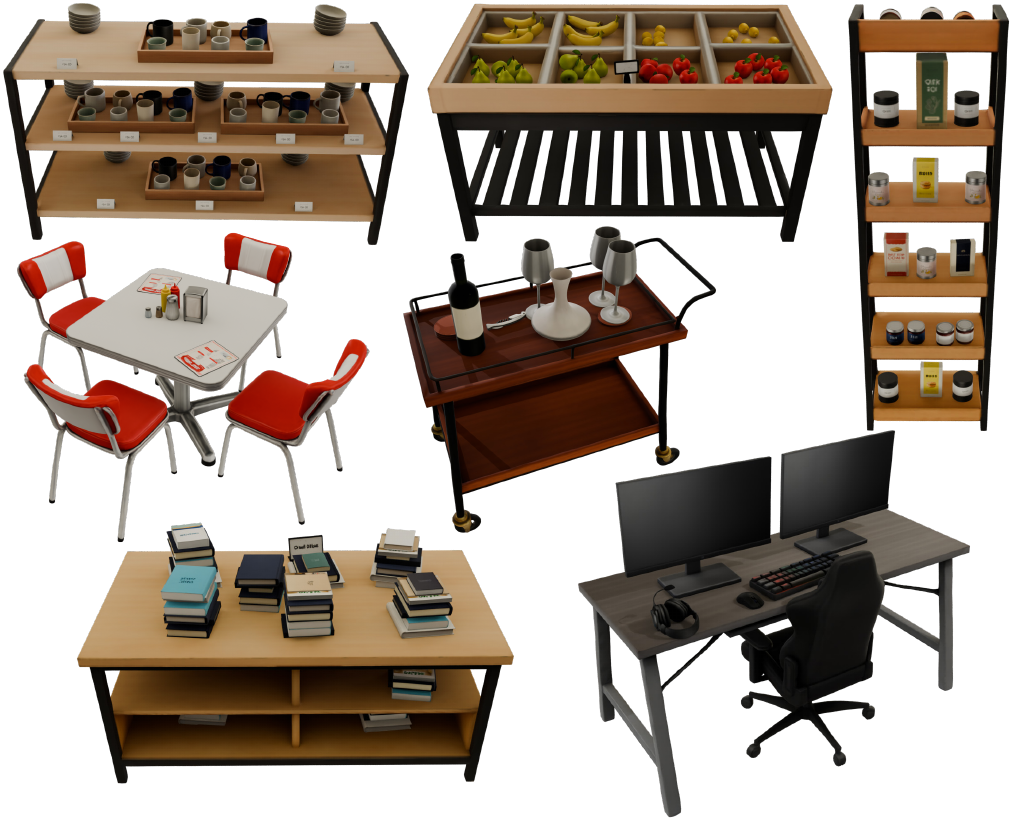}
\caption{\textbf{Manipuland placement examples.} Diverse furniture pieces populated with contextually appropriate manipulands: pottery store display table, produce stand with fruits, store shelf with products, diner table with menus, bar cart with bottles and glasses, library table with books, and gaming desk setup. Each object is a separate simulation asset that can be individually manipulated (e.g., each book in the stacks).}
\label{fig:manipuland_examples}
\end{figure}

\begin{figure}[htbp]
\centering
\includegraphics[width=\textwidth]{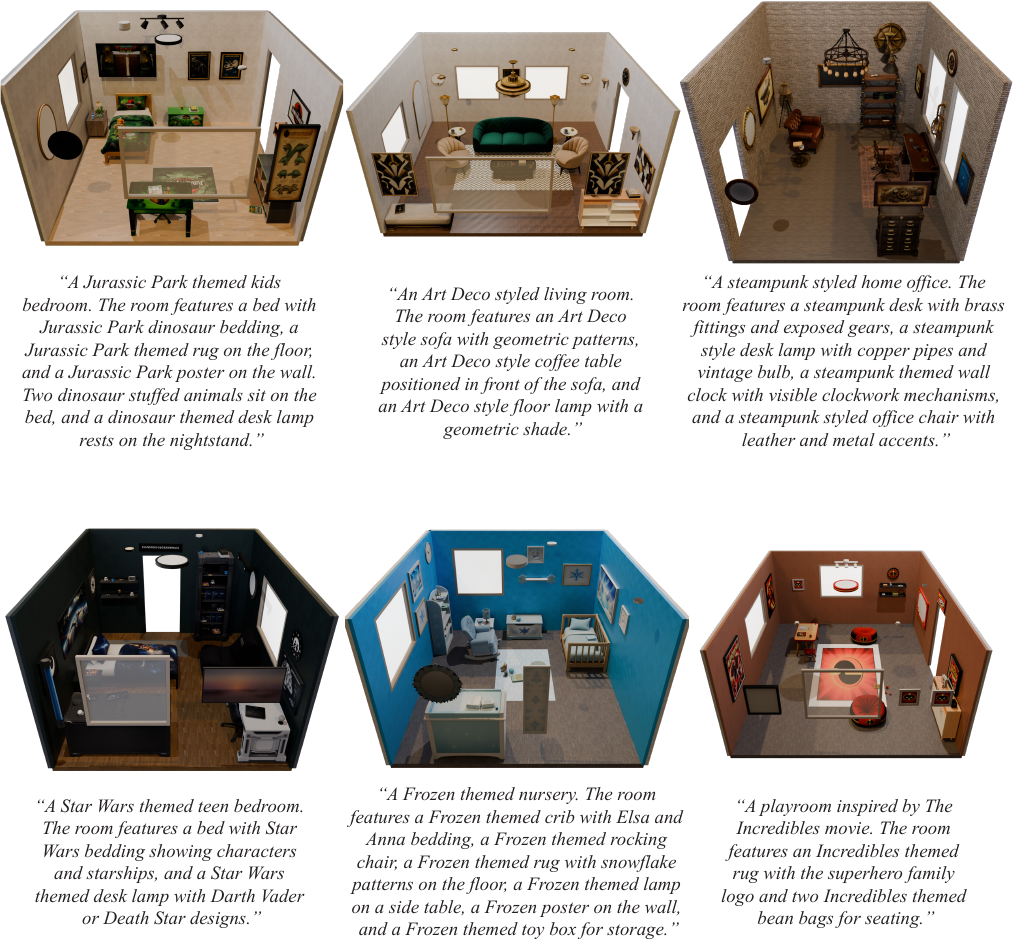}
\caption{\textbf{Themed room generation.} SceneSmith generates rooms with specific visual themes: Jurassic Park kids bedroom, Art Deco living room, steampunk home office, Star Wars teen bedroom, Frozen nursery, and Incredibles playroom. Generated assets and retrieved materials match the requested themes and styles.}
\label{fig:themed_prompts}
\end{figure}

Figure~\ref{fig:manipuland_examples} demonstrates the manipuland agent's ability to populate furniture with contextually appropriate objects. Figure~\ref{fig:themed_prompts} shows SceneSmith's capability to generate rooms matching highly specific visual themes.

\clearpage
\section{Limitations and Failure Analysis}
\label{app:limitations}

SceneSmith substantially improves over prior scene-generation baselines in scene density, prompt faithfulness, physical stability, and simulation readiness, but some limitations and failure cases remain. We summarize both representative failure modes observed in a review of the 210 generated evaluation scenes and broader limitations of the current pipeline. Across the automated evaluation and user study, we used all generated scenes as produced and did not discard or regenerate any outputs. The issues below are local imperfections observed in those scenes or scope limits of the current system, rather than catastrophic scene failures. Figure~\ref{fig:failure_cases} shows representative examples.

\subsection{Representative Failure Modes}

\begin{figure}[H]
    \centering
    \captionsetup[subfigure]{font=small,labelfont=bf,justification=centering}
    \begin{subfigure}[t]{0.315\textwidth}
        \centering
        \includegraphics[width=\linewidth]{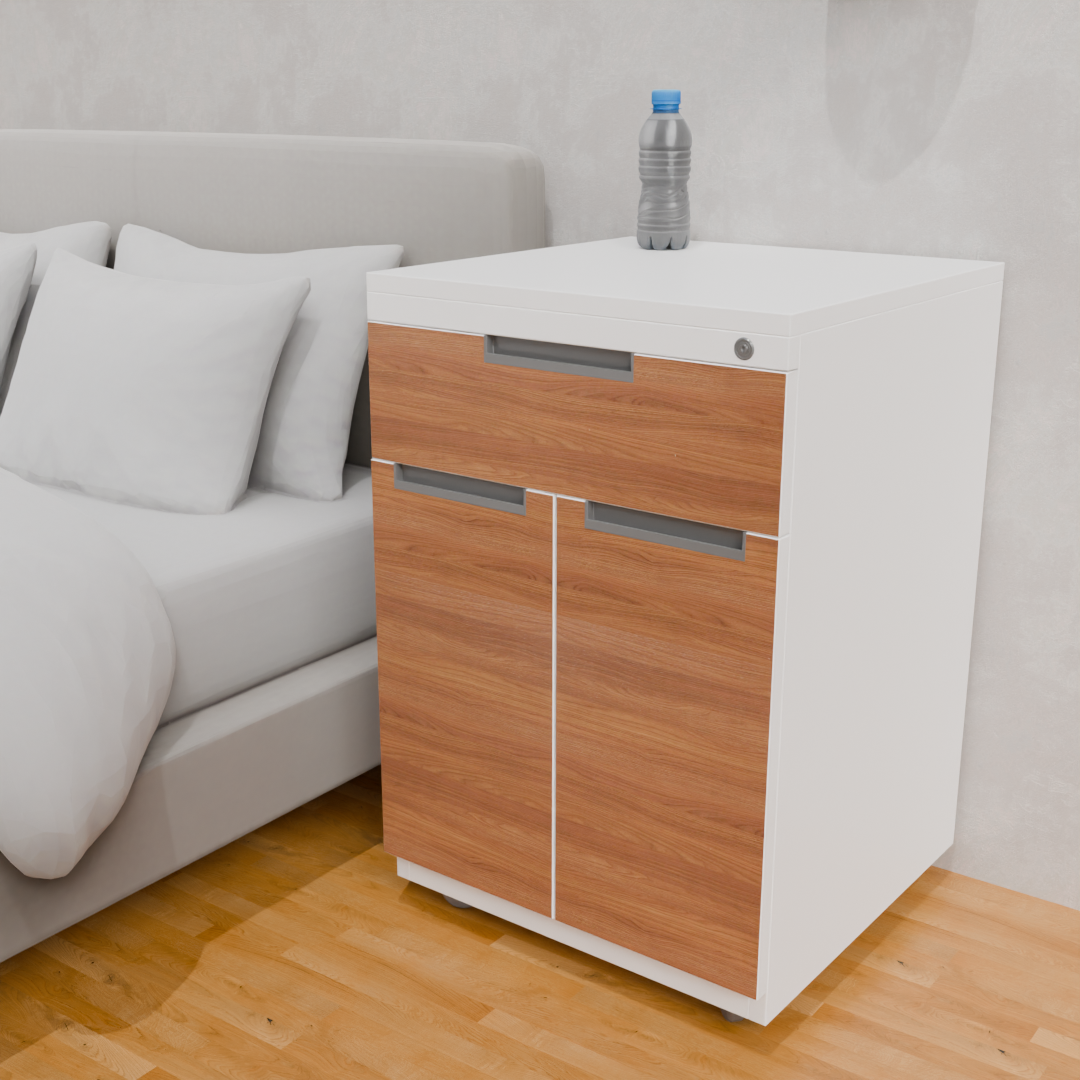}
        \caption{Wrong articulated retrieval}
        \label{fig:failure_retrieval}
    \end{subfigure}
    \hfill
    \begin{subfigure}[t]{0.315\textwidth}
        \centering
        \includegraphics[width=\linewidth]{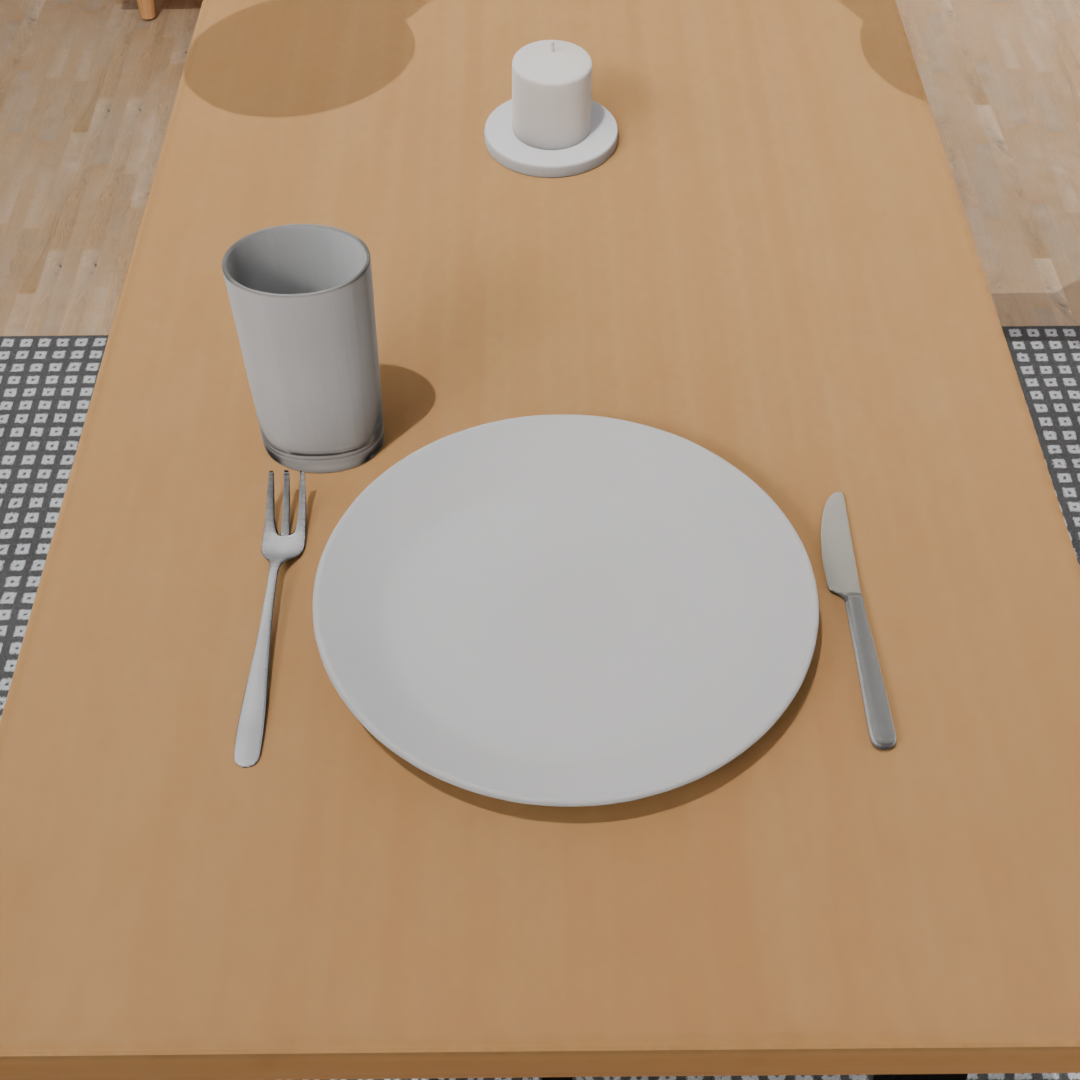}
        \caption{Missing PBR materials}
        \label{fig:failure_glass}
    \end{subfigure}
    \hfill
    \begin{subfigure}[t]{0.315\textwidth}
        \centering
        \includegraphics[width=\linewidth]{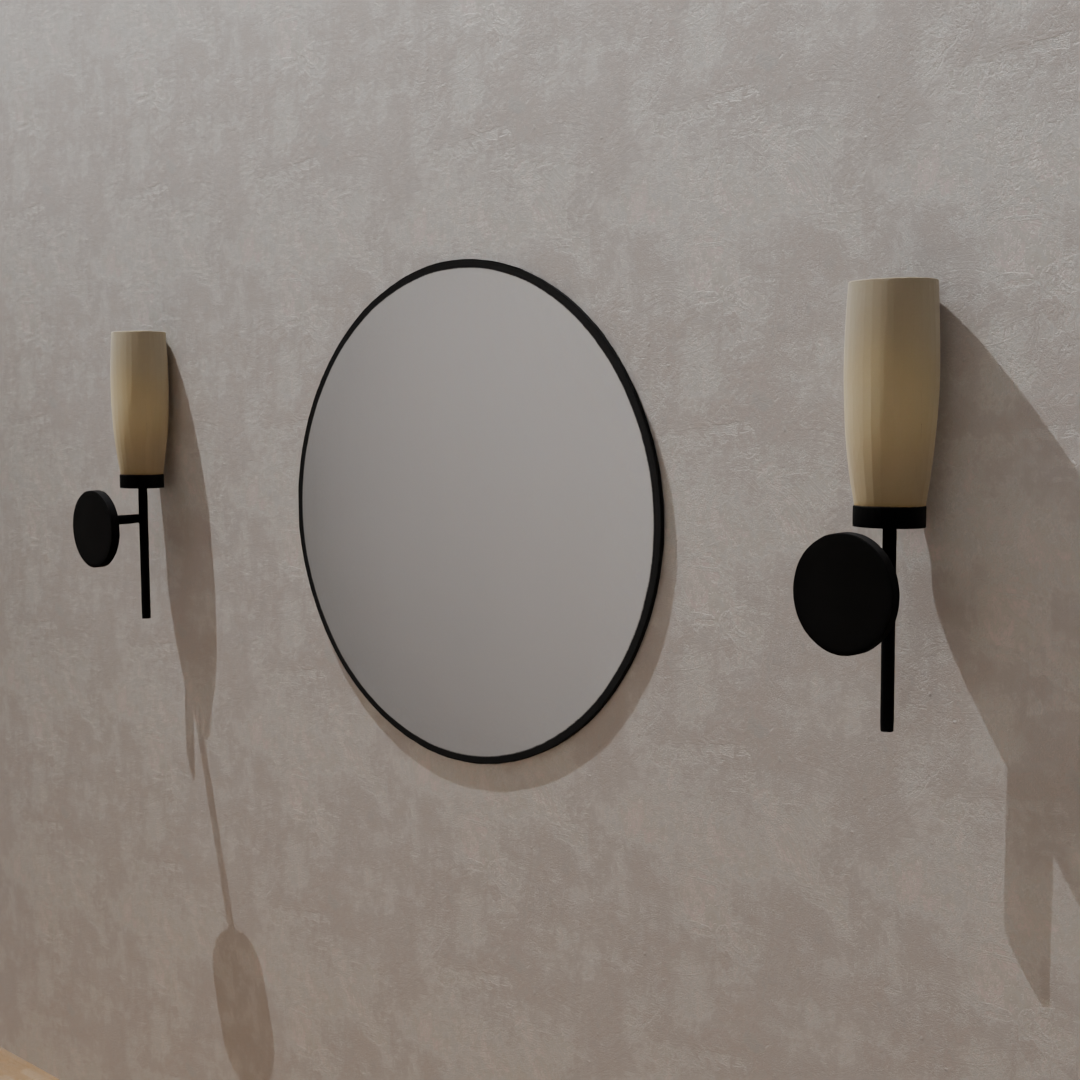}
        \caption{Lamp canonicalization}
        \label{fig:failure_lamp}
    \end{subfigure}
    \vspace{0.5em}

    \begin{subfigure}[t]{0.315\textwidth}
        \centering
        \includegraphics[width=\linewidth]{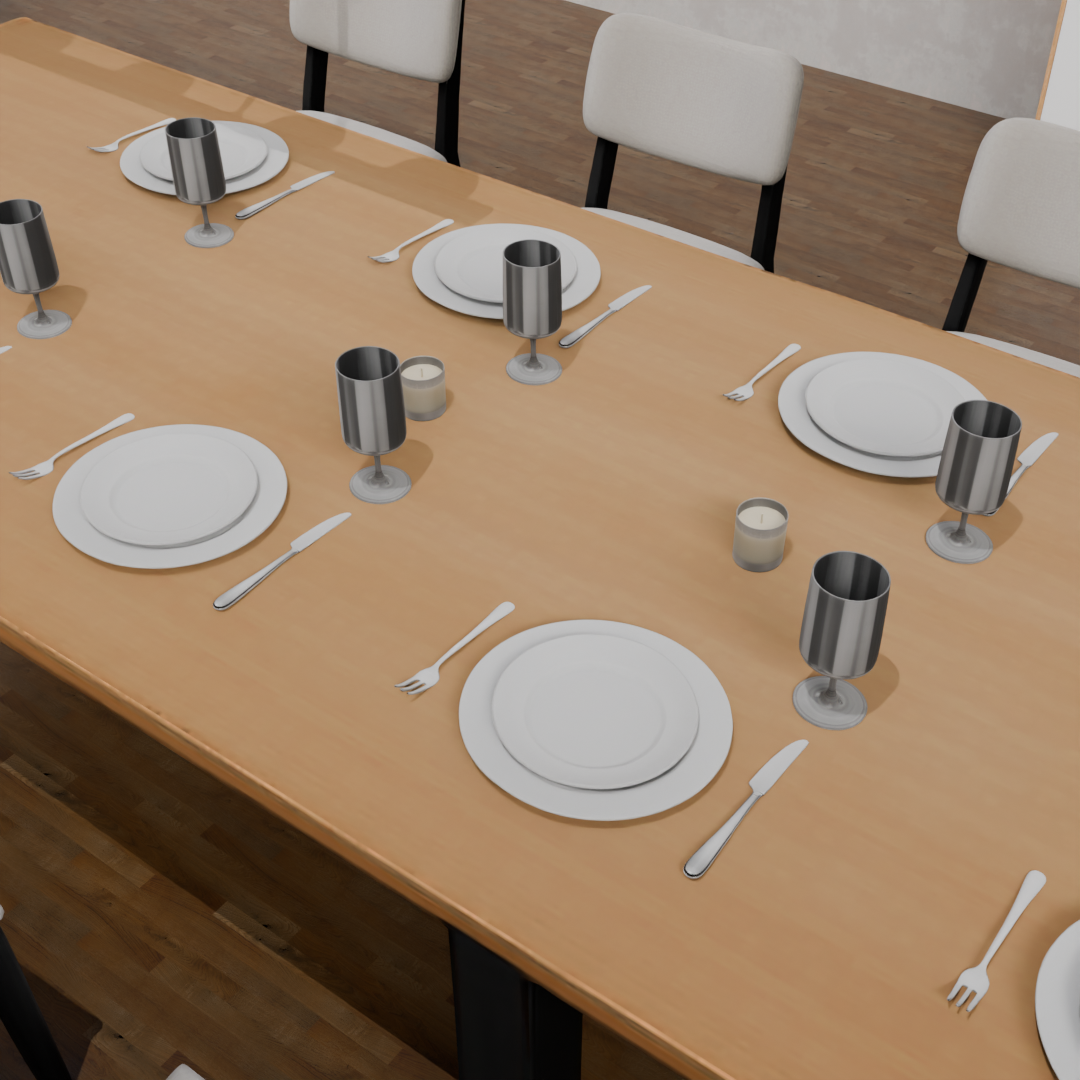}
        \caption{Utensil orientation}
        \label{fig:failure_utensils}
    \end{subfigure}
    \hfill
    \begin{subfigure}[t]{0.315\textwidth}
        \centering
        \includegraphics[width=\linewidth]{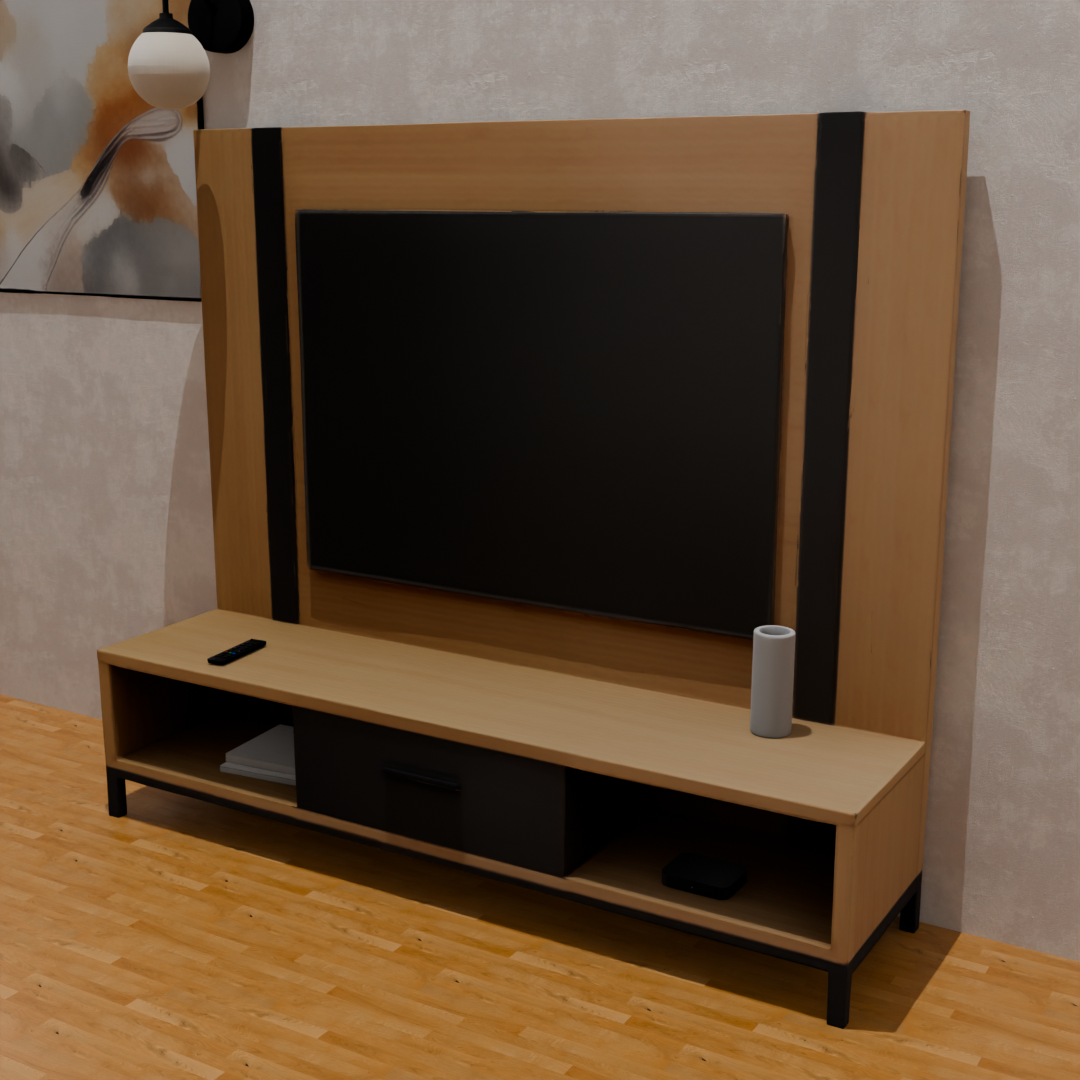}
        \caption{Stage-boundary error}
        \label{fig:failure_tv}
    \end{subfigure}
    \hfill
    \begin{subfigure}[t]{0.315\textwidth}
        \centering
        \includegraphics[width=\linewidth]{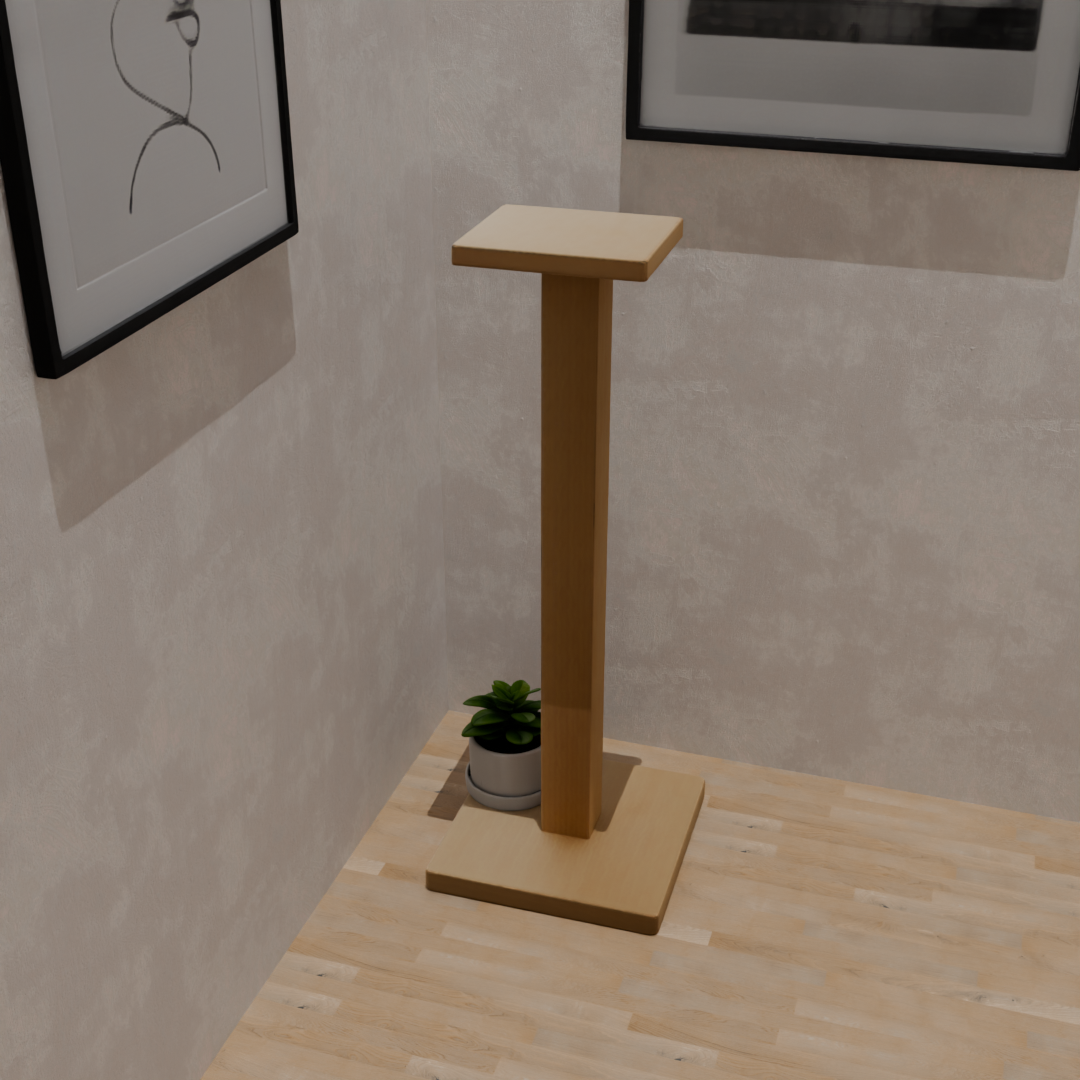}
        \caption{Missed support surface}
        \label{fig:failure_support}
    \end{subfigure}
    \caption{\textbf{Representative failure cases.} Local imperfections observed in generated evaluation scenes. \emph{Top row:} asset and asset-processing failures, including a storage cabinet retrieved for a requested mini fridge, glass reconstructed as opaque because PBR material channels are missing, and a wall lamp with incorrect canonical orientation. \emph{Bottom row:} reasoning and hierarchical-construction failures, including utensils that are not mirrored for diners on opposite sides of the table, a furniture-stage TV stand that bakes in the TV instead of leaving it to the manipuland stage, and a plant placed next to a stand after support-surface detection misses its small top surface.}
    \label{fig:failure_cases}
\end{figure}

\textbf{Asset-level failures.}
Retrieved articulated assets can occasionally be semantically close but incorrect; Figure~\ref{fig:failure_retrieval} shows a retrieved storage cabinet for a requested mini fridge. This occurs because articulated-object coverage is limited by the retrieval library: unlike generated assets, retrieved assets cannot be adjusted to exactly match the prompt. We therefore use a more tolerant validation prompt for retrieved articulated assets; otherwise, most articulated candidates would be rejected. This preserves articulated functionality but can occasionally accept a semantically close, incorrect asset. Separately, our current static-asset reconstruction uses SAM3D, which does not output full PBR material channels. As a result, transparent, metallic, or highly specular objects can have inaccurate material appearance: in Figure~\ref{fig:failure_glass}, the glass appears opaque rather than transparent and the cutlery lacks a metallic material response. Replacing SAM3D with a newer image-to-3D backbone that supports PBR materials, such as TRELLIS.2~\citep{xiang2025trellis2}, could address this limitation.

\textbf{VLM reasoning failures.}
Some errors arise from visual or spatial reasoning during asset processing and placement. In Figure~\ref{fig:failure_lamp}, canonicalization orients a wall-mounted lamp incorrectly relative to its mounting surface. In Figure~\ref{fig:failure_utensils}, the utensils are placed with the same orientation on both sides of the table instead of being mirrored for the diner on each side.

\textbf{Hierarchical construction failures.}
SceneSmith builds scenes sequentially from layout to furniture to manipulands. This decomposition makes generation scalable, but later stages cannot always revise earlier choices. The furniture stage can occasionally over-satisfy a prompt by incorporating a manipuland into a furniture asset. Figure~\ref{fig:failure_tv} shows a TV stand whose generated furniture asset already contains a TV, even though the TV should have been left to the manipuland stage. Support-surface detection can also miss small surfaces, and earlier-stage objects may not leave enough usable space for later placement. In Figure~\ref{fig:failure_support}, support-surface detection misses the small top surface of a plant stand, so the plant is placed on the floor next to it; similarly, a generated shelf may leave insufficient room for later bottle placement.

\subsection{System Limitations}

\textbf{Runtime and cost.}
The full system prioritizes high-fidelity, simulation-ready generation over single-scene latency. Cost is dominated by the Designer and refinement loop. The NoCritic ablation provides a lower-cost operating point, but it produces sparser scenes, reflecting a quality-cost trade-off.

\textbf{Articulated-object diversity.}
Articulated-object diversity is bounded by the retrieval library. If no suitable articulated asset is found, the router can fall back to text-to-3D generation, but the resulting object is static rather than articulated. This limits coverage of novel or out-of-distribution mechanisms. As image-to-articulated-object reconstruction and related generative methods mature, integrating them is a promising direction for expanding articulated coverage beyond fixed libraries.

\textbf{Estimated physical properties.}
Mass, friction, center of mass, and inertia are estimated automatically. To reduce hallucination risk, SceneSmith estimates material categories before mapping them to friction coefficients and represents mass as a range suitable for domain randomization. These estimates make broad simulation possible, but they are not a substitute for measured real-world physical parameters, especially for contact-rich tasks that depend on precise dynamics.

\textbf{Scope of simulated phenomena.}
SceneSmith focuses on rigid and articulated objects. Deformable objects, fluids, breakable objects, and detailed material behavior remain outside the current scope, both because they require specialized asset representations and because efficient, stable simulation remains challenging. A formal sim-to-real transfer study with SceneSmith-generated scenes is also future work.

\end{document}